\documentclass[10pt,journal,compsoc]{IEEEtran}

\usepackage{booktabs} 
\usepackage{graphicx}
\usepackage{amsmath}
\usepackage{amssymb}
\usepackage{color}
\usepackage{subfig}
\usepackage{amsmath}

\usepackage[pagebackref=true,breaklinks=true,letterpaper=true,colorlinks,bookmarks=false]{hyperref}
\usepackage{enumitem}

\usepackage{algorithm}
\usepackage{algpseudocode}
\usepackage{newfloat}
\usepackage{listings}
\usepackage{subfig}

\newcommand{\tvcg}[1]{\textcolor{black}{#1}}

\newcommand{\revision}[1]{\textcolor{black}{#1}}

\def\ie{i.e.}
\def\eg{{e.g.}}

\ifCLASSOPTIONcompsoc
  \usepackage[nocompress]{cite}
\else
  \usepackage{cite}
\fi

\begin{document}
%
\title{Zero-Shot Video Translation via Token Warping}
\author{Haiming Zhu,
        Yangyang~Xu,
        Jun Yu,
        and~Shengfeng~He,~\IEEEmembership{Senior Member,~IEEE}
\IEEEcompsocitemizethanks{\IEEEcompsocthanksitem Haiming Zhu is with the School of Computer Science and Engineering at South China University of Technology, China. E-mail: zhuhaimingzui@gmail.com.
\IEEEcompsocthanksitem Yangyang Xu and Jun Yu are with the School of Intelligence Science and Engineering, Harbin Institute of Technology (Shenzhen), China. E-mail: cnnlstm@gmail.com; yujun@hit.edu.cn.
\IEEEcompsocthanksitem Shengfeng He is with the School of Computing and Information Systems, Singapore Management University, Singapore. E-mail: shengfenghe@smu.edu.sg.
}
}

\markboth{IEEE Transactions on Visualization and Computer Graphics}%
{Zhu \MakeLowercase{\textit{et al.}}: Zero-Shot Video Translation via Token Warping}

\IEEEtitleabstractindextext{%
\begin{abstract}
With the revolution of generative AI, video-related tasks have been widely studied. However, current state-of-the-art video models still lag behind image models in visual quality and user control over generated content. In this paper, we introduce \emph{TokenWarping}, a novel framework for temporally coherent video translation. Existing diffusion-based video editing approaches rely solely on key and value patches in self-attention to ensure temporal consistency, often sacrificing the preservation of local and structural regions. Critically, these methods overlook the significance of the query patches in achieving accurate feature aggregation and temporal coherence. In contrast, \emph{TokenWarping} leverages complementary token priors by constructing temporal correlations across different frames. Our method begins by extracting optical flows from source videos. During the denoising process of the diffusion model, these optical flows are used to warp the previous frame's query, key, and value patches, aligning them with the current frame's patches. By directly warping the query patches, we enhance feature aggregation in self-attention, while warping the key and value patches ensures temporal consistency across frames. This token warping imposes explicit constraints on the self-attention layer outputs, effectively ensuring temporally coherent translation. Our framework does not require any additional training or fine-tuning and can be seamlessly integrated with existing text-to-image editing methods. We conduct extensive experiments on various video translation tasks, demonstrating that \emph{TokenWarping} surpasses state-of-the-art methods both qualitatively and quantitatively. Video demonstrations can be found on our project webpage: \url{https://alex-zhu1.github.io/TokenWarping/}. Code is available at: \url{https://github.com/Alex-Zhu1/TokenWarping}.
\end{abstract}

}

\maketitle


\section{Introduction}
\label{sec:intro}

Video translation has garnered significant attention and made substantial progress within the computer vision and graphics community. Prior works~\cite{wang2018video, li2019dense, ren2020deep, cui2021dressing} have leveraged Generative Adversarial Networks (GANs)~\cite{goodfellow2014generative} to various editing and translation applications~\cite{wu2023poce, 10816137,xiao2022appearance,jiang2023identity, Xu_2023_ICCV, chen2022sporthesia}. Despite their success, these translated videos merely mimic target frames and lack text-based editability~\cite{ma2023follow}. Recently, Text-to-Image (T2I) diffusion models~\cite{nichol2022glide, ramesh2021zero, saharia2022photorealistic} have made significant strides in static image synthesis, generating vivid images in various styles from text prompts. ControlNet~\cite{zhang2023adding} further enhances T2I's control capabilities by incorporating additional conditions beyond text prompts.

\begin{figure}[!t]
    \centering
        \captionsetup[subfloat]{justification=centering}
        \subfloat[DDIM Inv.]{
        \begin{minipage}{0.235\linewidth}
        \label{fig:com_attn-outputsa}
        \includegraphics[width=\linewidth]{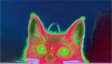}

        \vspace{0.5mm}

        \includegraphics[width=\linewidth]{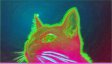}
        
        \vspace{0.5mm}

        \includegraphics[width=\linewidth]{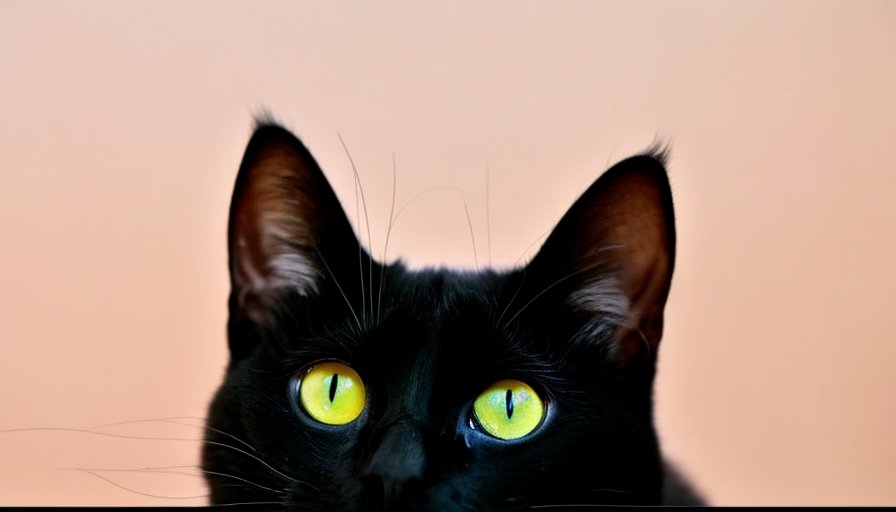}
        
        \vspace{0.5mm}

        \includegraphics[width=\linewidth]{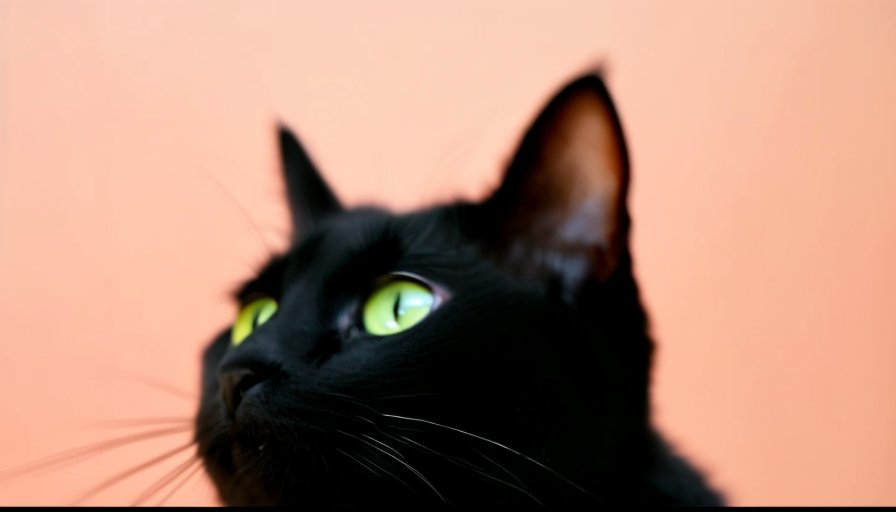}
        
        \vspace{0.5mm}

        \end{minipage}
        }
        \hspace{-2mm} 
        \subfloat[Warping \emph{KV}]{
        \begin{minipage}{0.235\linewidth}
        \label{fig:com_attn-outputsb}
        \includegraphics[width=\linewidth]{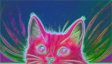}
                
        \vspace{0.5mm}

        \includegraphics[width=\linewidth]{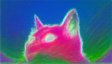}
                
        \vspace{0.5mm}

        \includegraphics[width=\linewidth]{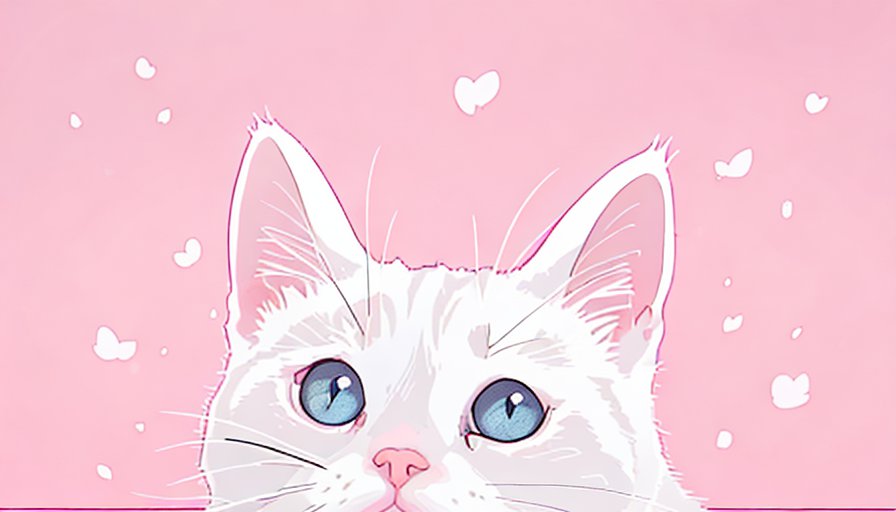}
                
        \vspace{0.5mm}

        \includegraphics[width=\linewidth]{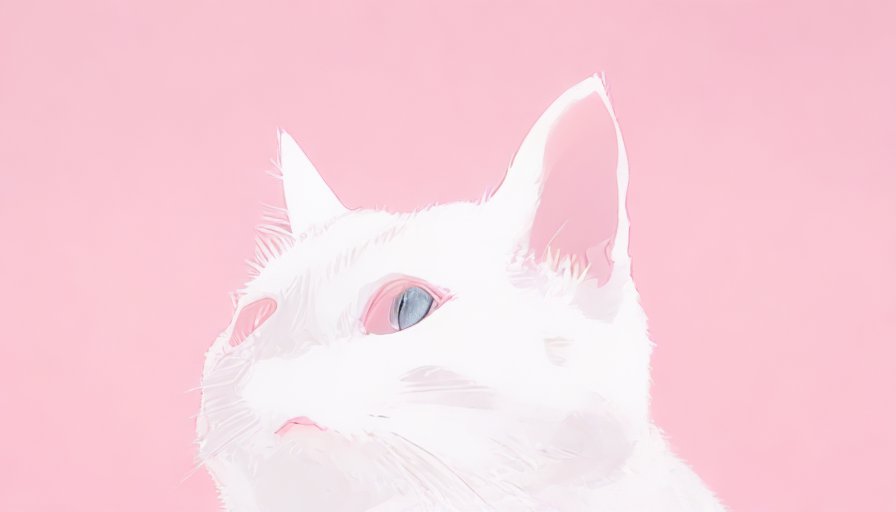}
                
        \vspace{0.5mm}

        \end{minipage}
        }
        \hspace{-2mm} 
        \subfloat[Warping \\ Attn-Out]{
        \begin{minipage}{0.235\linewidth}
        \label{fig:com_attn-outputsc}
        \includegraphics[width=\linewidth]{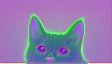}
                
        \vspace{0.5mm}

        \includegraphics[width=\linewidth]{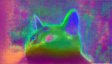}
                
        \vspace{0.5mm}

        \includegraphics[width=\linewidth]{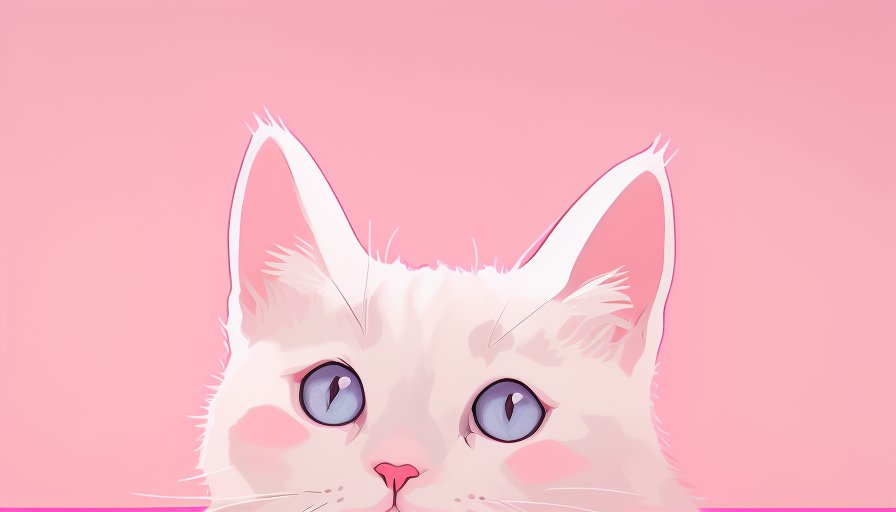}
                
        \vspace{0.5mm}

        \includegraphics[width=\linewidth]{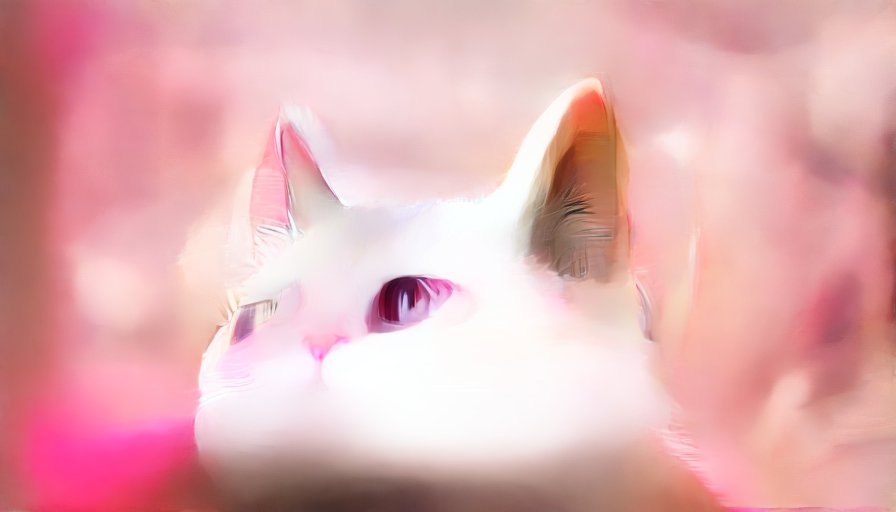}
                
        \vspace{0.5mm}

        \end{minipage}
        }
        \hspace{-2mm} 
        \subfloat[Ours]{
        \begin{minipage}{0.235\linewidth}
        \label{fig:com_attn-outputsd} 
        \includegraphics[width=\linewidth]{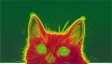}
                
        \vspace{0.5mm}

        \includegraphics[width=\linewidth]{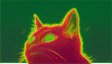}
                
        \vspace{0.5mm}

        \includegraphics[width=\linewidth]{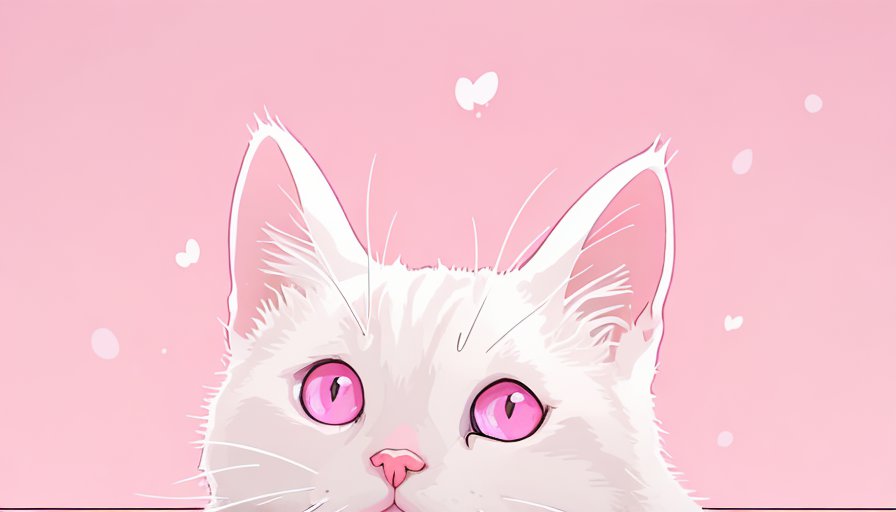}
                
        \vspace{0.5mm}

        \includegraphics[width=\linewidth]{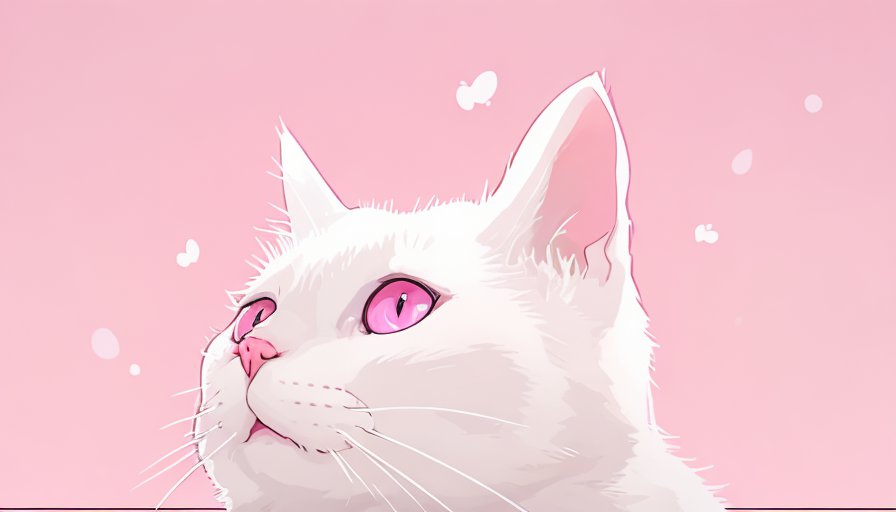}
                
        \vspace{0.5mm}

        \end{minipage}
        }
        \vspace{-2.5mm}
        \caption{Self-attention features visualization. The top two rows show attention features visualization after PCA, and the bottom two rows show the translated frames. \textit{Prompt: A white cat in pink background.}}\vspace{-5mm}
    \label{fig:com_attn-outputs}
    \end{figure}

However, maintaining structural consistency in transferred video motions remains a significant challenge. Existing works~\cite{wu2023tune, zhang2023controlvideo} focus on preserving temporal consistency by sharing \emph{key} and \emph{value} patches across frames, which can introduce irrelevant information and lead to misaligned token features. FLATTEN~\cite{cong2023flatten} addresses token alignment across frames using a flow-based attention mechanism applied to \emph{key} and \emph{value} patches. \revision{As shown in Fig.~\ref{fig:com_attn-outputsb}, warping the \emph{key} and \emph{value} patches leads to inaccurate feature aggregation, resulting in inconsistent and blurry translated frames.} Nevertheless, this approach can misalign spatial information within the current frame. 

Recently, TokenFlow~\cite{geyer2023tokenflow} utilizes nearest neighbor field to create the dense correspondence between frames, and uses it to warp the attention-output features directly. However, due to the misalignment of correspondence and output features, the warped output feature easily presenting the wrapped artifacts.
Here we visualize the feature map in self-attention using different flow integration methods. \revision{Our key observation is that consistent features tend to produce consistent appearance.  
Considering that DDIM inversion yields highly consistent features across frames (e.g., correspondence of the eyes), we adopt it as the ground truth to serve as a visual evaluation baseline, as illustrated in Fig.~\ref{fig:com_attn-outputsa}.  
The first approach, as in TokenFlow, directly warps the output features of the self-attention layer using optical flow, which introduces more severe warping artifacts (Fig.~\ref{fig:com_attn-outputsc}).  
The second approach, as in FLATTEN, warps the \emph{key} and \emph{value} patches, leading to feature degradation or disappearance and resulting in noticeable appearance inconsistency (Fig.~\ref{fig:com_attn-outputsb}).}

\begin{figure*}[t]
    \centering
    \begin{minipage}[c]{0.02\linewidth}
        \begin{minipage}[c][16mm][c]{\linewidth}
            \rotatebox{90}{ \small Source}
        \end{minipage}\\[0mm]
        \begin{minipage}[c][16mm][c]{\linewidth}
            \rotatebox{90}{ \small Rerender}
        \end{minipage}\\[0mm]
        \begin{minipage}[c][16mm][c]{\linewidth}
            \rotatebox{90}{ \small FRESCO}
        \end{minipage}\\[0mm]
        \begin{minipage}[c][16mm][c]{\linewidth}
            \rotatebox{90}{ \small Ours}
        \end{minipage}
    \end{minipage}%
    \begin{minipage}[c]{0.98\linewidth}
        \captionsetup[subfloat]{labelformat=empty,justification=centering}
        \begin{minipage}[c]{1\linewidth}
            \includegraphics[width=0.160\linewidth]{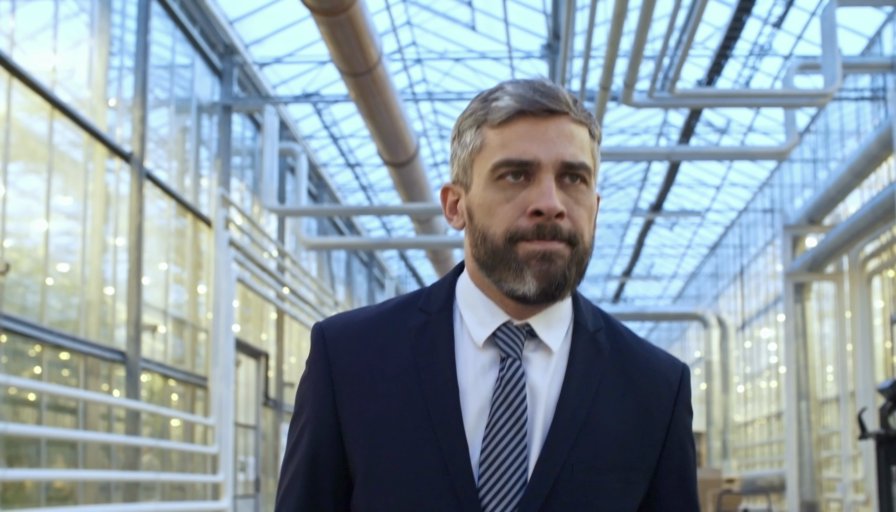}%
            \hspace{1mm}\includegraphics[width=0.160\linewidth]{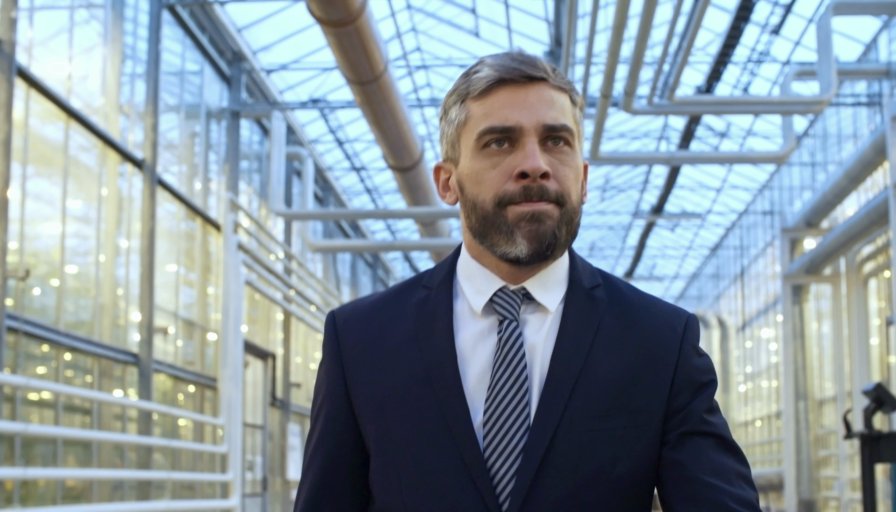}%
            \hspace{1mm}\includegraphics[width=0.160\linewidth]{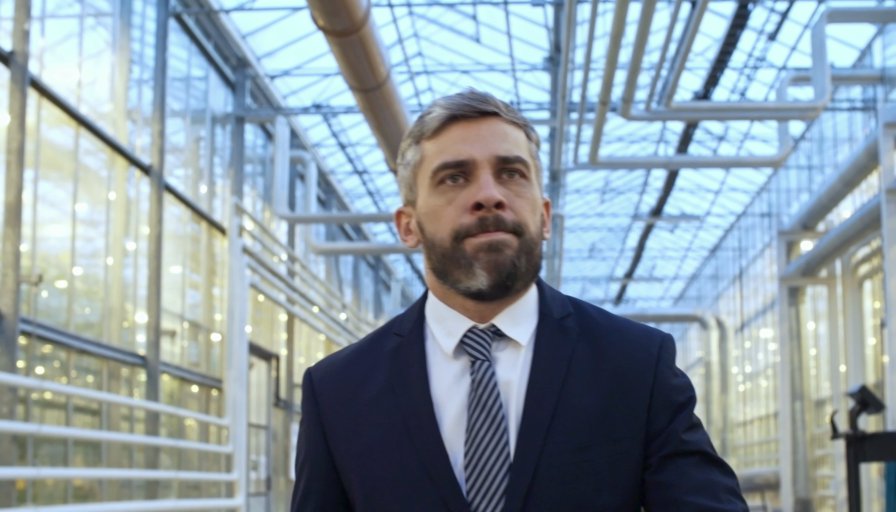}%
            \hspace{1mm}\includegraphics[width=0.160\linewidth]{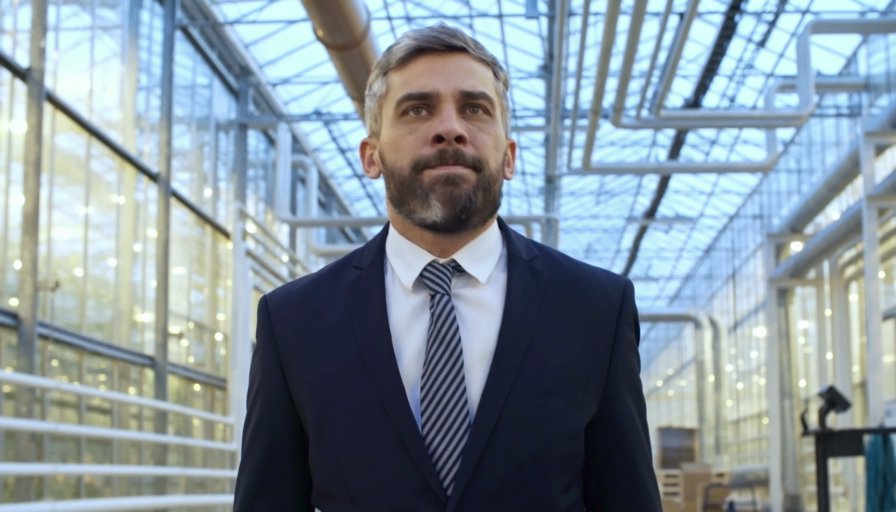}%
            \hspace{1mm}\includegraphics[width=0.160\linewidth]{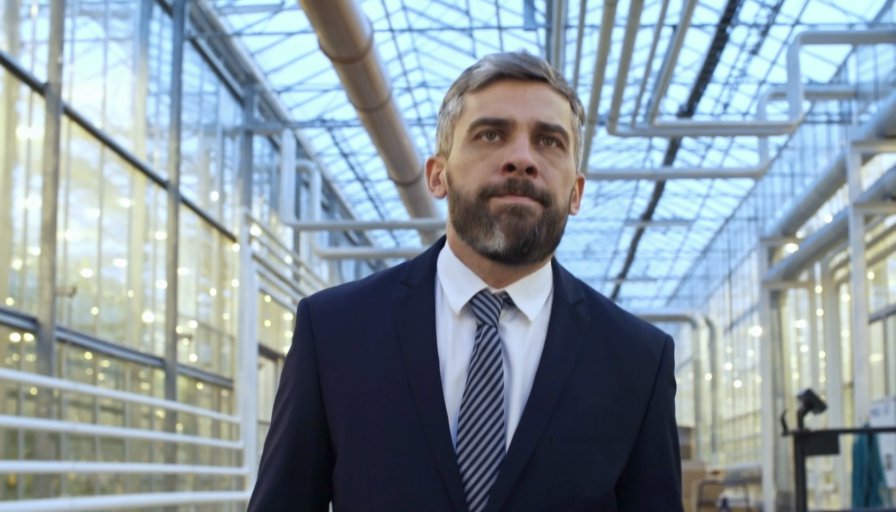}%
            \hspace{1mm}\includegraphics[width=0.160\linewidth]{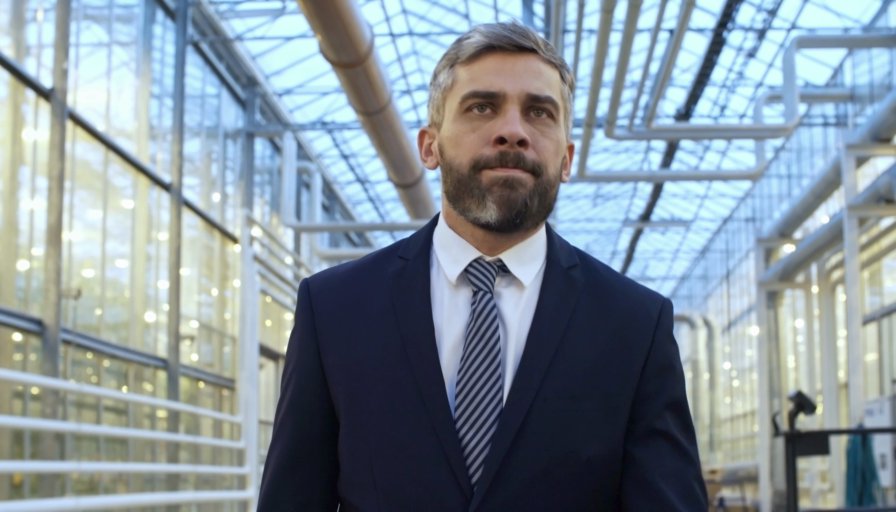}%
        \end{minipage}\vspace{0.5mm}
        \begin{minipage}[c]{1\linewidth}
            \includegraphics[width=0.160\linewidth]{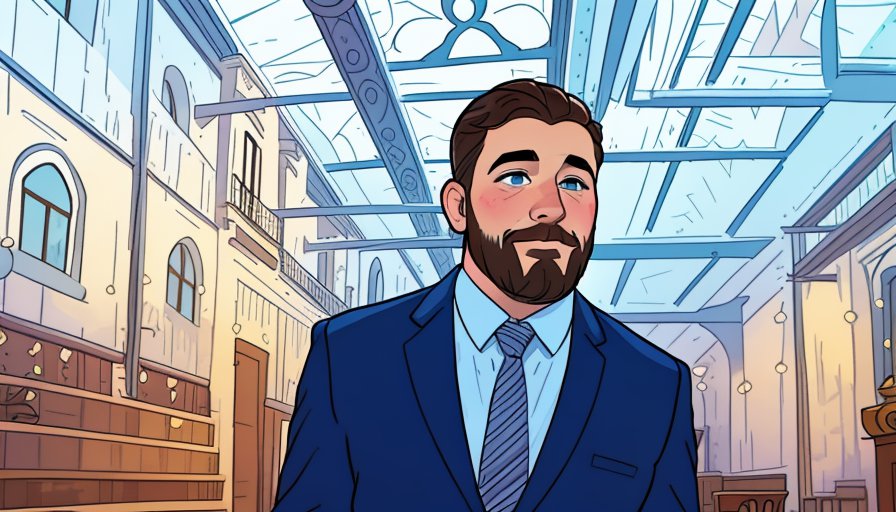}%
            \hspace{1mm}\includegraphics[width=0.160\linewidth]{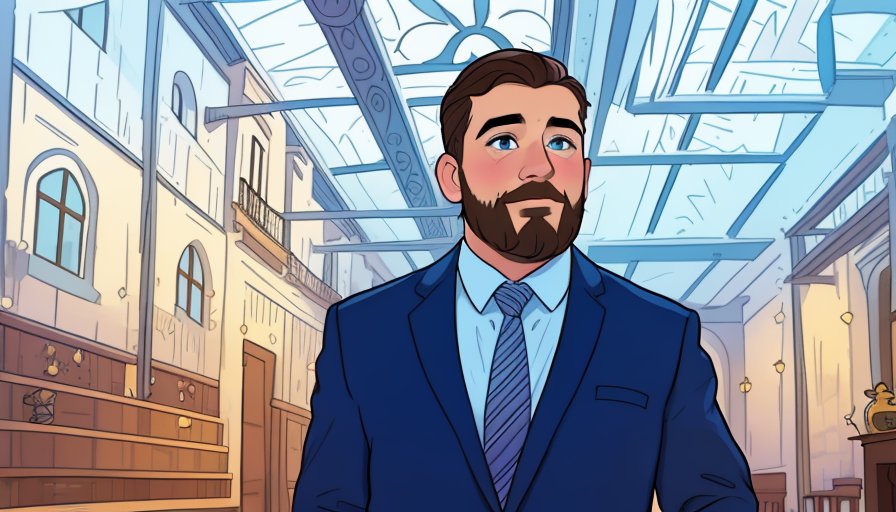}%
            \hspace{1mm}\includegraphics[width=0.160\linewidth]{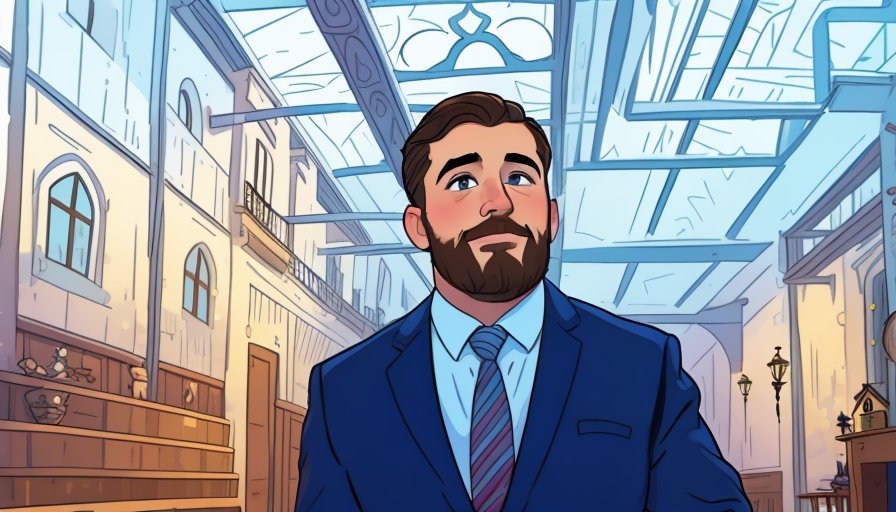}%
            \hspace{1mm}\includegraphics[width=0.160\linewidth]{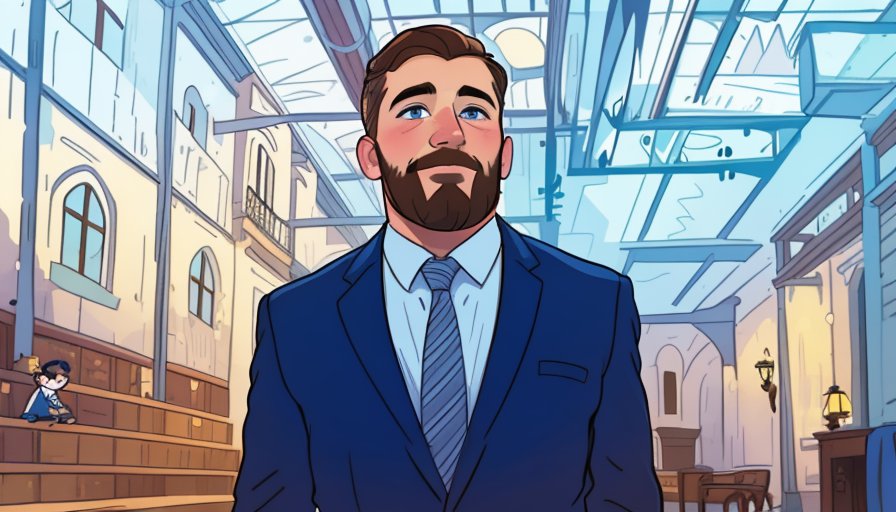}%
            \hspace{1mm}\includegraphics[width=0.160\linewidth]{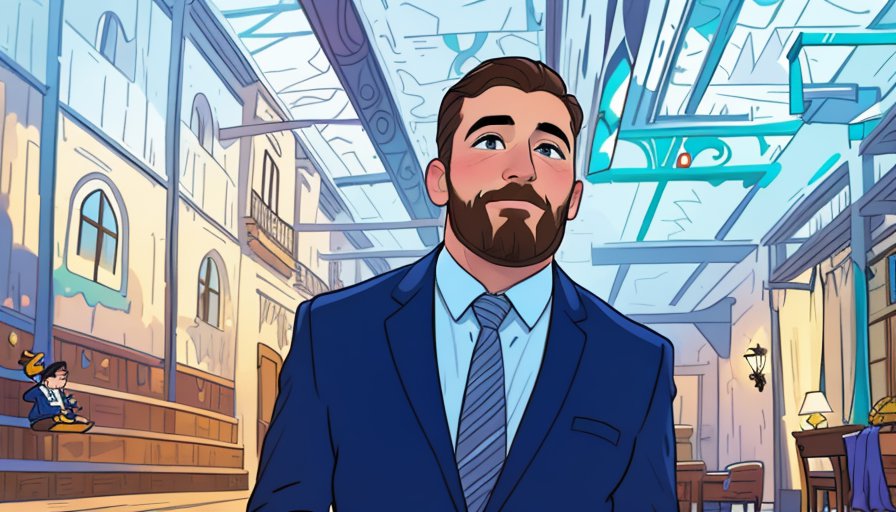}%
            \hspace{1mm}\includegraphics[width=0.160\linewidth]{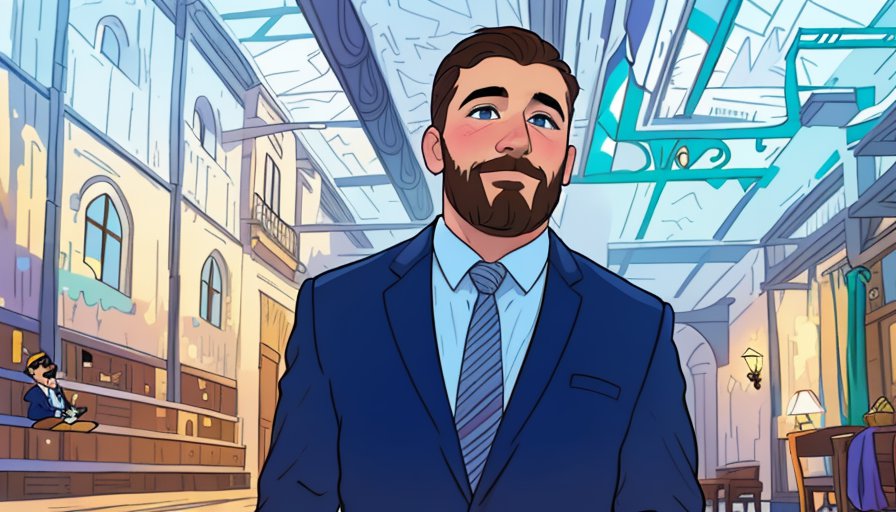}%
        \end{minipage}\vspace{0.5mm}
        \begin{minipage}[c]{1\linewidth}
            \includegraphics[width=0.160\linewidth]{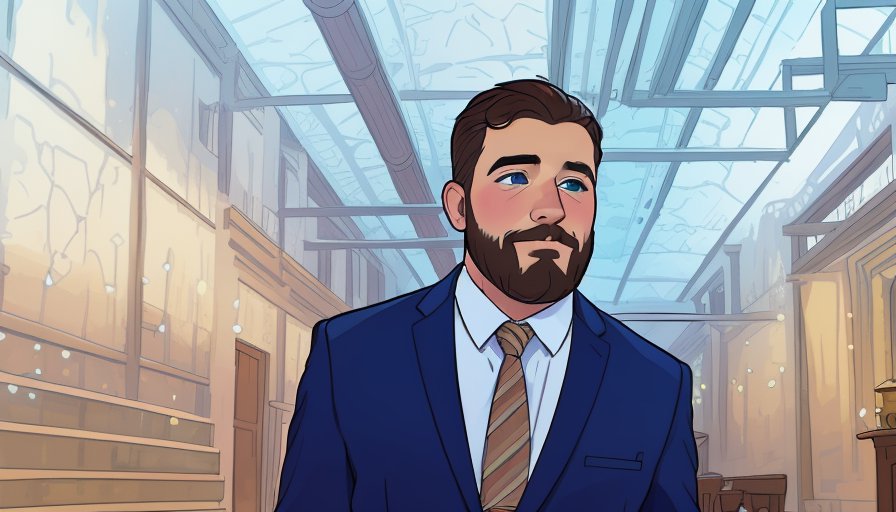}%
            \hspace{1mm}\includegraphics[width=0.160\linewidth]{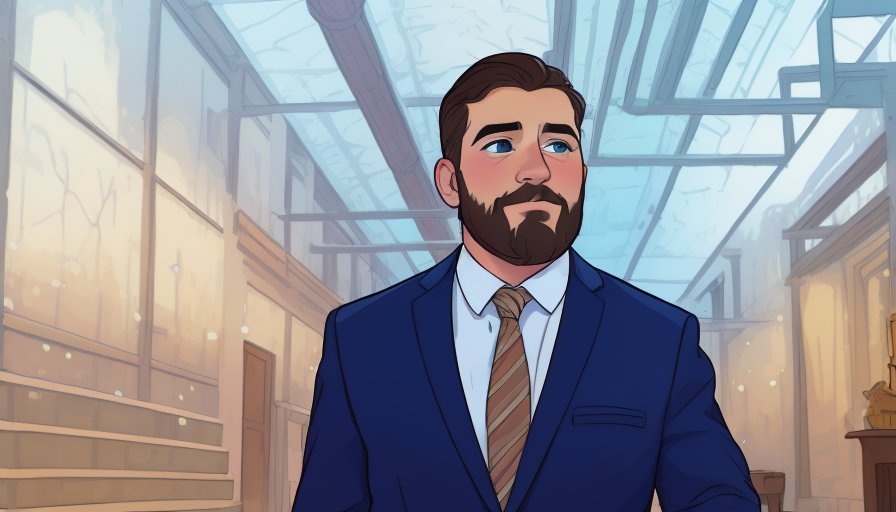}%
            \hspace{1mm}\includegraphics[width=0.160\linewidth]{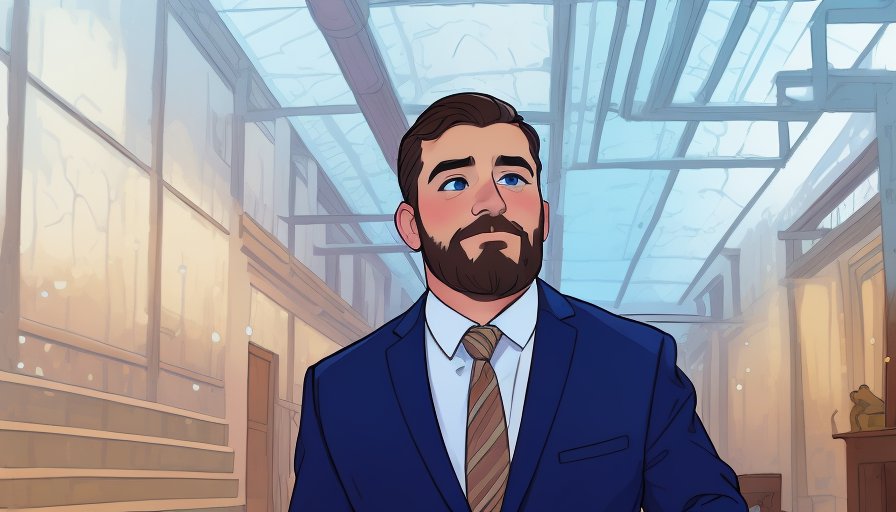}%
            \hspace{1mm}\includegraphics[width=0.160\linewidth]{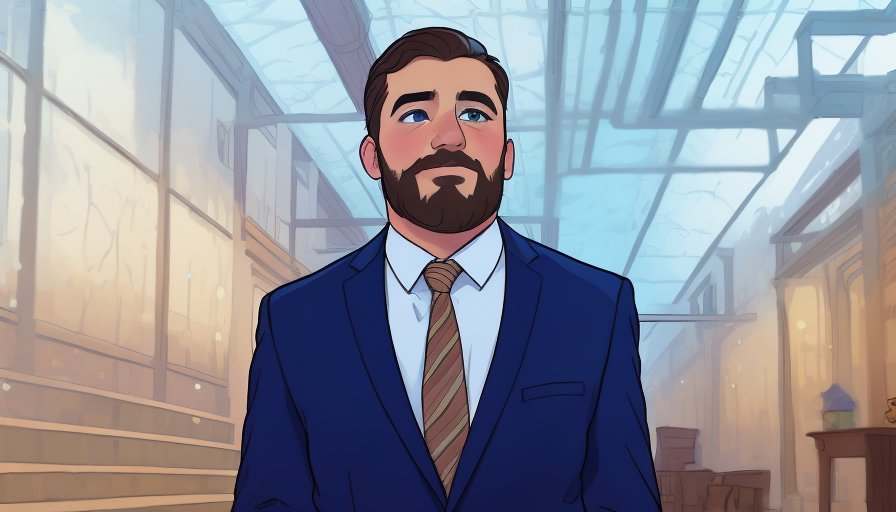}%
            \hspace{1mm}\includegraphics[width=0.160\linewidth]{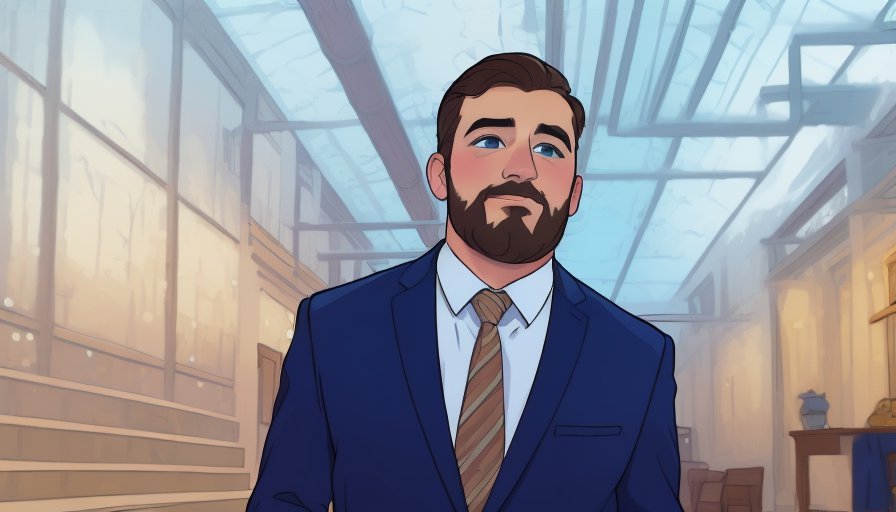}%
            \hspace{1mm}\includegraphics[width=0.160\linewidth]{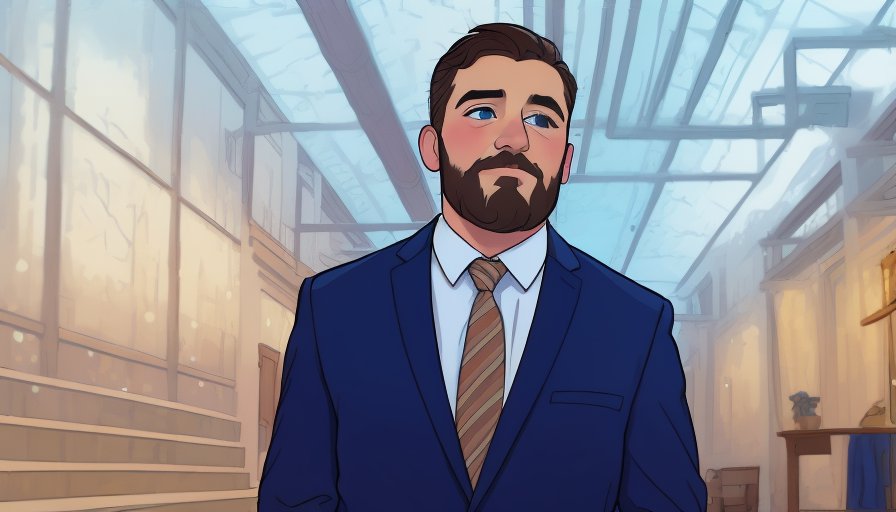}%
        \end{minipage}\vspace{0.5mm}
        \begin{minipage}[c]{1\linewidth}
            \includegraphics[width=0.160\linewidth]{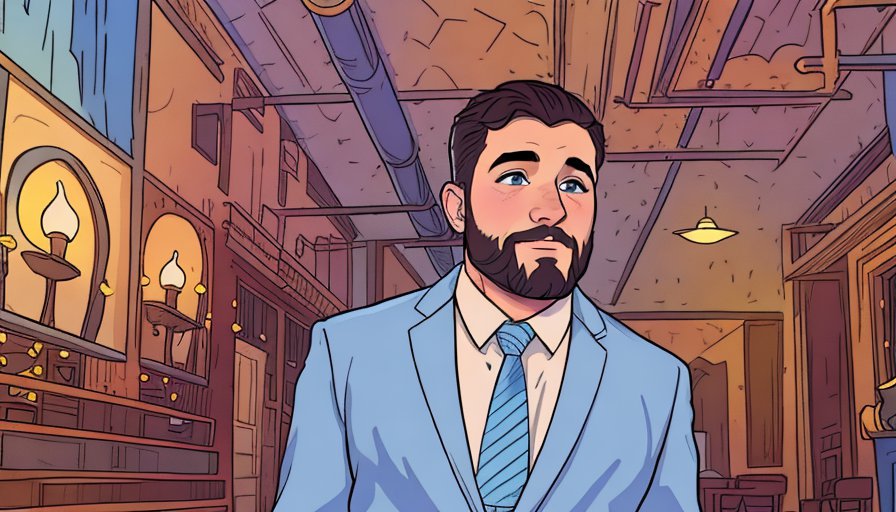}%
            \hspace{1mm}\includegraphics[width=0.160\linewidth]{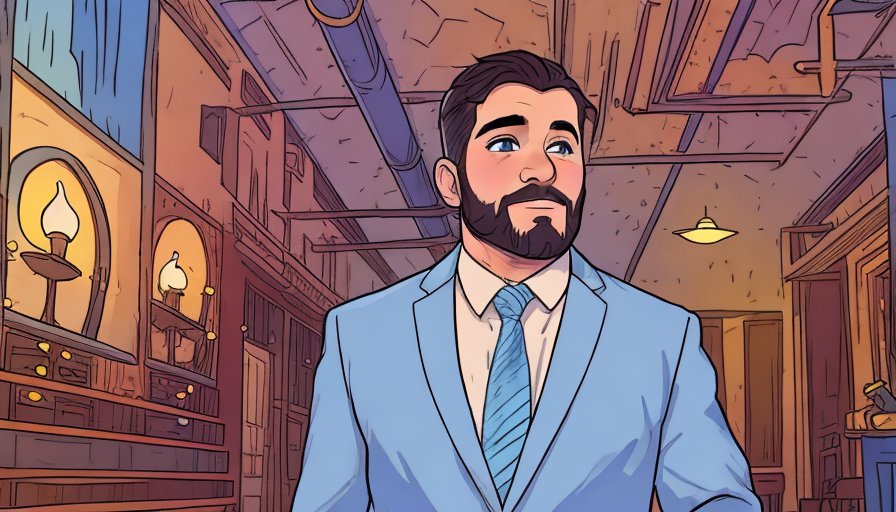}%
            \hspace{1mm}\includegraphics[width=0.160\linewidth]{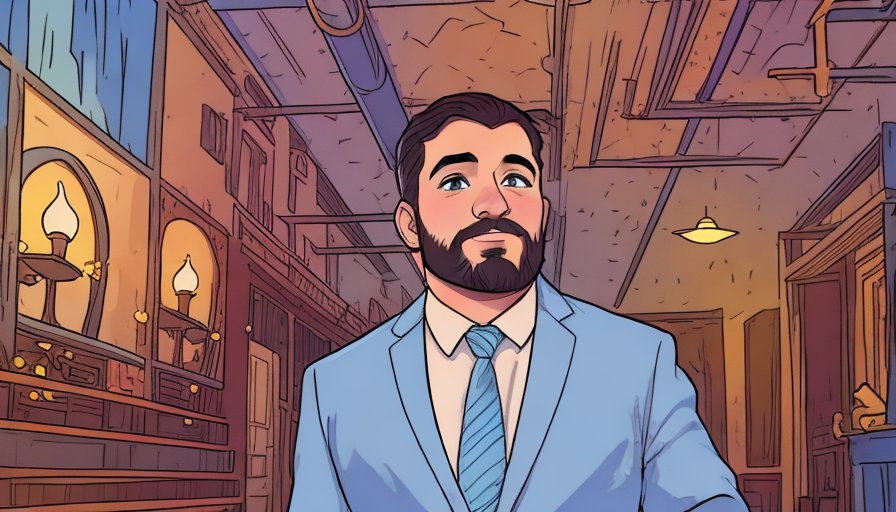}%
            \hspace{1mm}\includegraphics[width=0.160\linewidth]{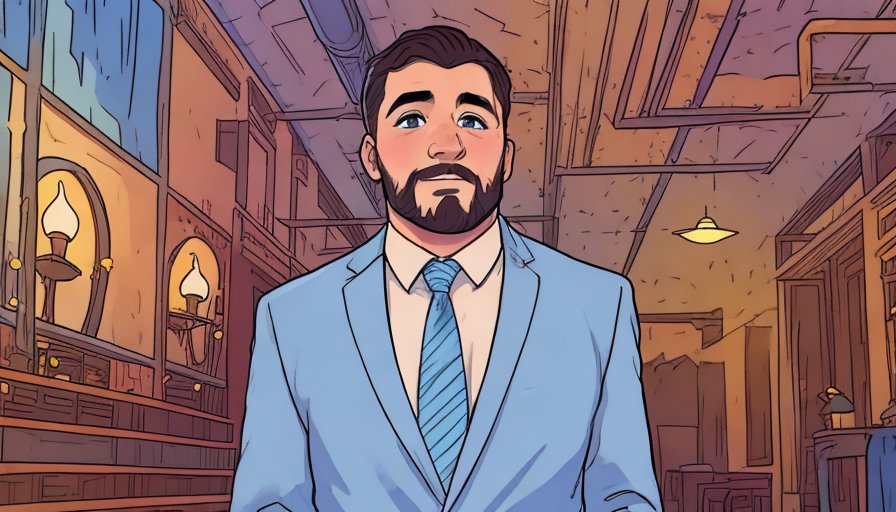}%
            \hspace{1mm}\includegraphics[width=0.160\linewidth]{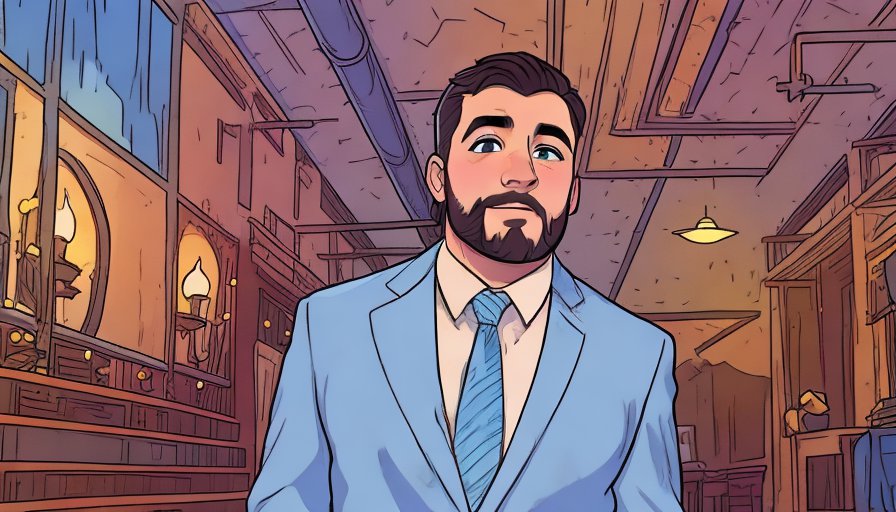}%
            \hspace{1mm}\includegraphics[width=0.160\linewidth]{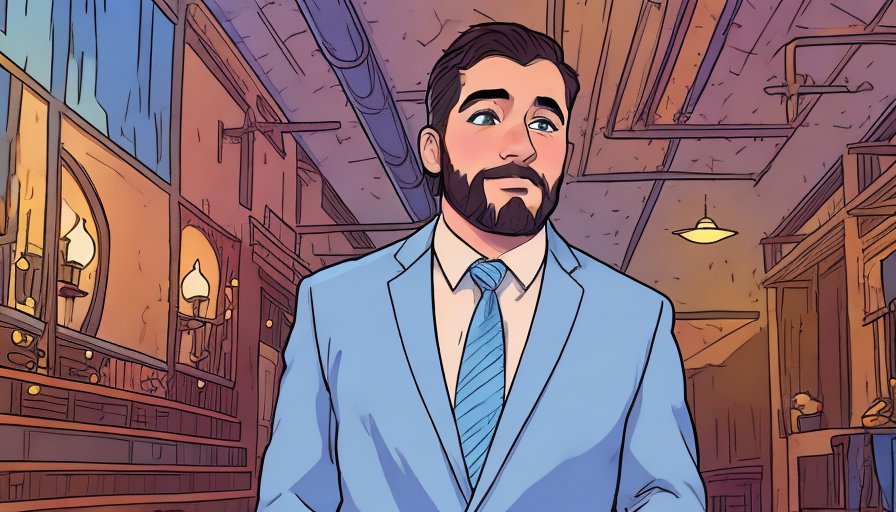}%
        \end{minipage}
    \end{minipage}
    \vspace{-1mm}
    \caption{We propose a novel zero-shot video translation method, \emph{TokenWarping}. Given the prompt \textit{``cartoon style, in the castle''}, \emph{TokenWarping} effectively transfers both the cartoon style and the background castle. In contrast, existing methods tend to overfit the source video, failing to edit the background.}\vspace{-5mm}
    \label{fig:com_1}
\end{figure*}

In this paper, we resolve these issues from an alternative perspective. We argue that the \emph{query} patches play a crucial role in addressing these challenges. Existing methods largely ignore the warping of the \emph{query} patches, focusing instead on the \emph{key} and \emph{value} patches. However, the \emph{query} patches are essential for accurate feature aggregation, and its misalignment can lead to significant temporal inconsistencies and visual artifacts. \revision{As shown in Fig.~\ref{fig:com_attn-outputsd}, warping the \emph{query}, \emph{key}, and \emph{value} patches using optical flow exhibits patch correspondences that closely resemble those of DDIM inversion, indicating improved temporal coherence and feature consistency.}

To this end, we introduce \emph{TokenWarping}, an efficient flow-guided attention mechanism that warps the \emph{query}, \emph{key}, and \emph{value} patches using optical flows \textbf{before} sending into the self-attention layer to achieve temporally coherent video translation. By directly warping the \emph{query} patches, we enhance feature aggregation in self-attention, while warping the \emph{key} and \emph{value} patches maintains temporal consistency across adjacent frames. To ensure long-term temporal consistency, we utilize anchor \emph{key} and \emph{value} patches for extended video translations. Initially, we align the \emph{key} and \emph{value} patches using flow, aiding in achieving inter-frame consistency. More importantly, by warping the token patches using flow and fusing occluded areas, our method reduces jitter and achieves smoother results, directly addressing the oversight of previous methods. As shown in Fig.~\ref{fig:com_attn-outputsd}, our warped feature are more similar with the Denoising Diffusion Implicit Models (DDIM) inversion's features~\cite{song2020denoising} (Fig.~\ref{fig:com_attn-outputsa}), indicating more consistent and temporally coherent results.
Our attention mechanism employs a single multi-head attention operation, eliminating the need for training or fine-tuning the diffusion model, and can be applied to both noisy and inverted latent codes.
We conducted extensive experiments on video translation tasks with various styles derived from text prompts. The results demonstrate the superiority of our method over state-of-the-art approaches in both quantitative and qualitative evaluations.

In summary, our contributions are three-fold:
\begin{itemize}[leftmargin=*]
    \item \revision{We provide a systematic analysis of different warping strategies in attention mechanism, revealing that warping only \emph{query}/\emph{key} or attention outputs leads to feature misalignment and temporal inconsistency.}
    \item \revision{We propose \emph{TokenWarping}, a novel framework for zeroshot video translation that warps the \emph{query}, \emph{key}, and \emph{value} patches using optical flow, ensuring local temporal coherence and reducing jitter.}
    \item Extensive experiments validate that our proposed \emph{TokenWarping} achieves state-of-the-art performance in video translation tasks.
\end{itemize}

\section{Related Work}

\subsection{Diffusion Models}

Diffusion models~\cite{austin2021structured,gu2022vector,kingma2021variational,rombach2022high,dhariwal2021diffusion} have gained significant attention for their generative capabilities. Starting with random noise, these models progressively denoise it to generate high-quality samples. Recently, diffusion-based T2I models~\cite{nichol2022glide,ramesh2021zero,saharia2022photorealistic} have set new benchmarks in image synthesis. The Latent Diffusion Model~\cite{rombach2022high}, in particular, performs the diffusion process in the latent space of a variational auto-encoder~\cite{kingma2014auto}, synthesizing high-quality images from text prompts. This generative ability has been leveraged by numerous works~\cite{hertz2022prompt,wu2023tune,zhang2023controlvideo,cao_2023_masactrl} for real image and video editing, \tvcg{sketch extraction~\cite{yang2024mixsa}, and anime customization~\cite{xu2024dreamanime}.}

\subsection{Diffusion-based Video Generation and Editing}

While diffusion models have demonstrated remarkable generative abilities, their application to video generation and editing is an emerging field. Video Diffusion Models~\cite{ho2022video} introduced a space-time U-Net to perform diffusion on pixels, while Imagen Video~\cite{ho2022imagen} achieved high-quality video generation using cascaded diffusion models and v-prediction parameterization. Make-A-Video~\cite{singer2022make} combined the appearance generation of T2I models with movement information from video data. Recent studies~\cite{esser2023structure, ge2023preserve, xing2024make} have explored re-training T2I models using video data to enable text-to-video functionality. \revision{ EVE~\cite{singer2024video} further proposed a Factorized Diffusion Distillation strategy to enable diverse video editing tasks. }

In the realm of video editing, maintaining temporal consistency is both essential and challenging. Tune-A-Video~\cite{wu2023tune} addresses this by inflating a 2D U-Net to 3D for modeling temporal information and introduces a temporal attention mechanism. VideoP2P~\cite{liu2023video} builds on this by utilizing Prompt-to-Prompt~\cite{hertz2022prompt} editing on the tuned model. \tvcg{Make-Your-Video~\cite{xing2024make} introduced a effective causal attention mask strategy to enable longer video synthesis.} Fatezero~\cite{QI_2023_ICCV} implements zero-shot text-driven video editing through a blending attention mechanism. \revision{RAVE~\cite{kara2024rave} introduces a noise shuffling strategy to ensure consistency across grids, whereas Slicedit~\cite{pmlr-v235-cohen24a} leverages spatiotemporal slices to preserve both structure and motion.}
By introducing ControlNet~\cite{zhang2023adding}, several works does not require the invert the source videos to the diffusion model at first~\cite{zhang2023controlvideo, yang2024fresco}. They follow Rerender~\cite{yang2023rerender} designed a inversion-free zero-shot video-to-video translation framework. ControlVideo~\cite{zhang2023controlvideo} introduces a fully attentive mechanism for video-to-video translation. {We also follow Rerender~\cite{yang2023rerender} and ControlVideo~\cite{zhang2023controlvideo} to perform zero-shot video translation.}

However, most of these works achieve temporal attention by sharing \emph{key} and \emph{value} patches across frames, neglecting the critical role of the \emph{query} patches in maintaining temporal consistency. 
\revision{Several works~\cite{cao_2023_masactrl, tewel2024training} have shown that \emph{query} patches encode structural layout information in T2I models, while MotionByQueries~\cite{atzmon2025motion} further demonstrates that \emph{query} patches influence not only the layout but also the subject's motion in T2V models.}
As evidenced in our experiment, the \emph{query} patches is essential for accurate feature aggregation, and its misalignment can lead to significant temporal inconsistencies and visual artifacts.
TokenFlow~\cite{geyer2023tokenflow} propagated attention output features from key frames to other frames based on correspondences in the source video features, these attention output features are not aligned with the video structure and cannot be directly applied using optical flow. {Furthermore, unlike inversion-free methods~\cite{yang2023rerender,zhang2023controlvideo}, inversion-based methods like those in \cite{geyer2023tokenflow,cong2023flatten} inherently retain more source-specific information and may remain unchanged in some cases, as demonstrated in our video demo.}

\subsection{Flow-guided Attention}


Recent advances in diffusion model editing~\cite{tumanyan2023plug, tang2023emergent} have highlighted the spatial correspondences inherent in the self-attention and decoder mechanisms of U-Net. To maintain temporal consistency in translated frames, some works~\cite{wu2023tune, zhang2023controlvideo} share \emph{key} and \emph{value} patches across frames in the self-attention mechanism, but ignores the essential \emph{query} patches, as the \emph{query} patches attends to all tokens and aggregates features. TokenFlow~\cite{geyer2023tokenflow} searches nearest neighbor field based on feature correspondences to create a flow-guided attention mechanism. However, TokenFlow~\cite{geyer2023tokenflow} decodes the warped attention-output features, leading to artifacts due to the insufficient reconstruction capability of the decoder. In contrast, our method directly warps the \emph{query}, \emph{key}, and \emph{value} patches before attention, which inherently integrates the reconstruction of warped tokens.

\revision{Some works have explored optical flow optimization of latent codes. Ground-A-Video~\cite{jeongground} employs optical flow to refine the inverted latents. Go-with-the-Flow~\cite{burgert2025go} leverages optical flow to warp noise, thereby enhancing motion control in T2V models.}
Within the attention mechanism, FLATTEN~\cite{cong2023flatten} designs flow-based sampling trajectories in self-attention to enhance fine-grained temporal consistency. However, FLATTEN focuses solely on \emph{key} and \emph{value} patches, overlooking the essential temporal consistency of \emph{query} patches.
FRESCO~\cite{yang2024fresco} introduces a FRESCO-guided attention mechanism and feature optimization to preserve intra-frame spatial correspondence, while effective, this method involves dual multi-head attention, leading to redundant computations and increased complexity. Moreover, the optimization process in FRESCO easily leading to the overfit of the source video's spatial structure (see in Fig.~\ref{fig:com_1}.)
Our flow-guided attention mechanism directly aligns the \emph{query}, \emph{key}, and \emph{value} patches using optical flow. This approach requires only one multi-head attention mechanism to process the aligned tokens, making it more efficient and effective in preserving temporal consistency.
\section{Approach}
\label{sec:approach}

\begin{figure*}
  \centering
  \includegraphics[width=.95\textwidth]{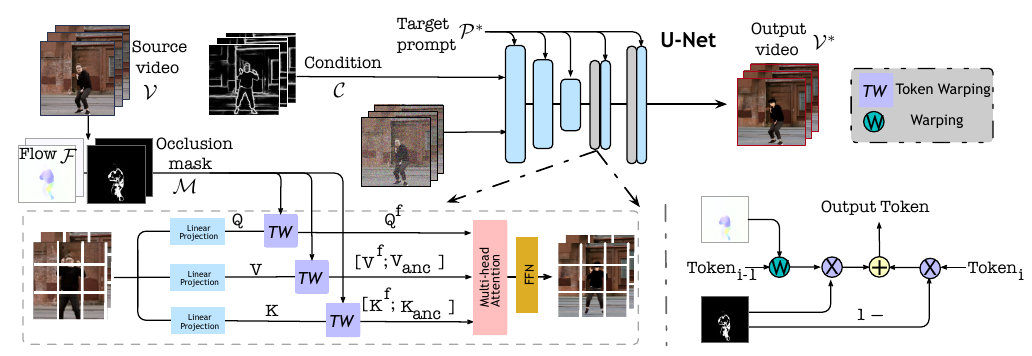}
  \vspace{-2mm}
  \caption{Pipeline of our \emph{TokenWarping}: Given a source video $\mathcal{V}$, we first predict the optical flow $\mathcal{F}$ and occlusion mask $\mathcal{M}$ using the method from~\cite{xu2022gmflow}. We then feed the sequence condition $\mathcal{C}$ and target prompt $\mathcal{P^*}$ to ControlNet, which controls the outputs of Stable Diffusion. During each denoising step, we warp the \emph{query}, \emph{key}, and \emph{value} tokens in the U-Net decoder's self-attention layers using optical flow. At each timestep $i$, we sample the anchor area ($1_{st}$ frame) of the \emph{key} and \emph{value} patches and concatenate it with the warped patches along the feature axis. Additionally, we use optical flow to warp the \emph{query} patches, enhancing local temporal consistency. The detailed illustration of warping and fusion are shown in bottom-right.}
  \label{fig:framework}
  \vspace{-5mm}
  \end{figure*}

\subsection{Preliminaries}
\label{subsec:pre}

\textbf{Latent Diffusion Models (LDMs)} is a text-to-image model that conducts a diffusion process in the latent space of an autoencoder. It has an autoencoder and a Denoising Diffusion Probabilistic Model (DDPM)~\cite{ho2020denoising}. Given an image $x$, \revision{it is} encoded to a latent code $z$ through an autoencoder $\mathcal{E}$, \ie, $z = \mathcal{E}(x)$. In the forward diffusion process, the DDPM adds Gaussian noise to the latent code $z$ iteratively:
\begin{equation}
  q(z_{t}|z_{t-1})=\mathcal{N}(z_{t};\sqrt{1-\beta_{t}}z_{t-1},\beta_{t}I), 
\end{equation}
where $q(z_{t}|z_{t-1})$ is the conditional density of $z_{t}$ given $z_{t-1}$, $\{\beta_{t}\}^T_{t=0}$ are the scale of noises, and $T$ is the total timestep of the diffusion process.

The backward denoising process is represented as:
\begin{equation}
  p_\theta(z_{t-1}|z_t) = \mathcal{N}(z_{t-1}; {\mu}_\theta(z_t, t), {\Sigma}_\theta(z_t, t)),
\end{equation}
where the ${\mu}_\theta$ and ${\Sigma}_\theta$ are implemented with denoised model~$\epsilon_\theta$, it is trained using following objective:
\begin{equation}
  \mathcal{E}_{{z, \epsilon} \sim {\mathcal{N}(0,1), t}} [\|\epsilon - \epsilon_\theta(z_t, t, c_\mathcal{P})\|_2^2],
  \label{train}
  \end{equation}
where $c_\mathcal{P}$ is the text prompt.\\
\textbf{DDIM Sampling and Inversion} During inference, deterministic DDIM sampling~\cite{song2020denoising} is employed to progressively convert a random Gaussian noise $z_T$ to a clean latent code $z_0$ with following equation:
\begin{equation}
  z_{t-1} = \sqrt{\alpha_{t-1}} \frac{z_t - \sqrt{1 - \alpha_t}\epsilon_\theta}{\sqrt{\alpha_t}} + \sqrt{1 - \alpha_{t-1}}\epsilon_\theta,
\end{equation}
where $t$ is denoising step $t:T \rightarrow 1 $ and $\alpha_{t}$ is a parameter for noise scheduling ~\cite{ho2020denoising, song2020denoising}.

To reconstruct real images and perform editing ~\cite{hertz2022prompt, mokady2023null}, DDIM inversion is employed to encode a real image latent code $z_0$ to related inversion noise by reversing the above process in revered steps $t:1\rightarrow T$.\\
\textbf{ControlNet} is a conditional text-to-image generative model, capable of handling various conditions $\mathcal{C}$, \eg, depth maps, poses, edges. \revision{By constructing the noise prediction network~$\epsilon_\theta(z_t, t, \mathcal{P}, \mathcal{C})$}, ControlNet~\cite{zhang2023adding} adds a trainable copy encoding model for the conditional input $\mathcal{C}$. It then utilizes zero-convolutions connected with the prompt input $\mathcal{P}$ for task-specific conditional image generation.

\subsection{Procedure}

Given a source video $\mathcal{V}=\{v_i \mid i \in[1, N]\}$ with $N$ frames, our goal is to translate it to a target temporal coherent video $\mathcal{V^*}$ under the structure condition $\mathcal{C}=\{c_i \mid i \in[1, N]\}$, meanwhile retains the sequential motions from the source video.
The translated video appearance is controlled by the given target prompt $\mathcal{P^*}$ and the structure guidance $\mathcal{C}$. {The pipeline of our~\emph{TokenWarping} is shown in the top part of Fig.~\ref{fig:framework}, we build the framework based on Stable Diffusion~\cite{rombach2022high} and ControlNet~\cite{zhang2023adding}. We first follow Tune-A-Video~\cite{wu2023tune}, which inflates the 2D U-Net~\cite{ronneberger2015u} of the T2I model to a pseudo-3D U-Net by converting the 3×3 convolution kernels in the convolutional residual blocks to 1×3×3 kernels through the addition of a pseudo-temporal channel, requiring no additional parameters or layers.
And we further reprogram the self-attention layer with the optical flows into a flow-guided attention for preserving the temporal consistency of translated videos.}

\subsubsection{Self-Attention Mechanism}

Before introducing our method, we would like to introduce the attention mechanism of the original T2I model at first. In specific, given the latent representation $z_i$ of frame $i$, the original self-attention mechanism first projects it to~\emph{query},~\emph{key}, and~\emph{value} patches ($Q_i$, $K_i$, and $V_i$) respectively. Then the self-attention mechanism is presented as:
\begin{equation}
Q_i = W^Qz_i, K_i = W^Kz_i, V_i=W^Vz_i,
\end{equation}
\begin{equation}\operatorname{Attn}(Q_i,K_i,V_i)=\operatorname{SoftMax}(\frac{Q_iK_i^T}{\sqrt{d}})\cdot V_i,
\end{equation}
where $W^Q$, $W^K$, and $W^V$ project $z_i$ into~\emph{query},~\emph{key},~\emph{value} respectively, and $d$ is the output dimension of~\emph{key} and~\emph{query} patches.

The self-attention mechanism handles each frame individually, which cannot guarantee the temporal consistency of frames. To eliminate the content inconsistency, existing T2I-based video editing works~\cite{wu2023tune,zhao2023controlvideo} select a key frame and propagate its content to other frames. Particularly, they replace the \emph{key} and \emph{value} patches of different frames using an anchor frame's token, that is, they extend the self-attention to cross-frame attention using the shared anchor \emph{key} and \emph{value} patches. Specifically, on frame $i$, the cross-frame attention can be presented as:
\begin{equation}
\operatorname{CFAttn}_i=\operatorname{SoftMax}(\frac{Q_iK_{anc}^T}{\sqrt{d}})\cdot V_{anc},
\label{eq:1}
\end{equation}
where $K_{anc}$ and $V_{anc}$ denotes the selected anchor \emph{key} and \emph{value} patches.
However, the temporal inconsistency cannot be eliminated due to the \emph{query} patches $Q_i$ is adopted from the current frame, which is unaligned with the shared \emph{value} and \emph{key} patches from key frame.

\subsubsection{{Flow-guided Attention}}
\label{fga}

For reducing the temporal inconsistency of translated videos, we reprogram the cross-frame attention mechanism into a flow-guided attention mechanism that builds the temporal correlations on the aligning tokens.


Existing flow-based attention~\cite{cong2023flatten, yang2024fresco} uses a flow-based sampling operation where the~\emph{key} and~\emph{value} patches are sampling from previous frame.
They designed a flow-based sampling trajectories to ensure consistent patches in the \emph{key} and \emph{value} patches.
This can be seen as using the resized flow $f_{i\Rightarrow i-1}$~\footnote{Here we use $f_{i\Rightarrow i-1}$ to denote original and resized flows for simplify.} to warp~\emph{key} and \emph{value} patches:
\begin{equation}
K^{\prime}_{i-1} = \texttt{W}(K_{i-1}, f_{i\Rightarrow i-1}), V^{\prime}_{i-1} = \texttt{W}(V_{i-1}, f_{i\Rightarrow i-1})
\end{equation}
where $\texttt{W}(,\cdot,)$ is the backward warping operation, $K_{i-1}$ and $V_{i-1}$ is the previous tokens, $K^{\prime}_{i-1}$ and $V^{\prime}_{i-1}$ is the warped results. In FLATTEN~\cite{cong2023flatten}, only the embeddings of the patches along the flow trajectory are gathered, and the unaligned patches (i.e., occlusion regions) are not considered.

As shown in the bottom-right of Fig.~\ref{fig:framework}, we introduce the occlusion map $m_{i\Rightarrow i-1}$ to handle the occlusion region. The occlusion map is used to control the fusion of the warped tokens and the original tokens:
\begin{equation}
K^{f}_{i} = m_{i\Rightarrow i-1} \cdot K^{\prime}_{i-1} + (1-m_{i\Rightarrow i-1}) \cdot K_{i}  ,
 \label{eq:k}
\end{equation}
\begin{equation}
V^{f}_{i} = m_{i\Rightarrow i-1} \cdot V^{\prime}_{i-1} + (1-m_{i\Rightarrow i-1}) \cdot V_{i},
 \label{eq:v}
\end{equation}
where $K_{i}$, $V_{i}$ is the current key and value patches, and $K^{f}_{i}$, $V^{f}_{i}$ is the fusion results.

Previous works directly utilize current \emph{query} patches to aggregate previous patches, in contrast our flow-based attention, patches in the flow-based sampling trajectories will aggregate each other based on the \emph{similarities} of previous patches rather than previous \emph{key} patches. Specifically, we further warp the \emph{query} patches before attention calculation:

\begin{equation}
  Q^{\prime}_{i-1} = \texttt{W}(Q_{i-1}, f_{i\Rightarrow i-1}),
  \end{equation}
  \begin{equation}
  Q^{f}_{i} = m_{i\Rightarrow i-1} \cdot Q^{\prime}_{i-1} + (1-m_{i\Rightarrow i-1}) \cdot Q_{i},
  \label{eq:q}
\end{equation}
where $Q^{\prime}_{i-1}$ is the warped results of \emph{query} patches $Q_{i-1}$, and $Q^{f}_{i}$ is the fused \emph{query} patches.

The warped \emph{query} patches will align with the warped \emph{key} and \emph{value} patches in flow-based sampling trajectories, and allowing the occlusion region to aggregate feature from the current frame rather than only sampling from the previous frame.
While the warped tokens align with the source video's motion, translating long-term videos remains challenging. To regularize global style consistency, we introduce anchor ($1_{st}$ frame) patches $K_{anc}$ and $V_{anc}$ to preserve the global appearance. The warped \emph{key} and \emph{value} patches are then concatenated with the anchor patches along the feature axis. The flow-guided attention can be expressed as follows:
\begin{equation}
\operatorname{FGAttn}_i=\operatorname{SoftMax}(\frac{Q^{f}_i [K_{anc}, K^{f}_{i}]^{T} }{\sqrt{d}})\cdot [V_{anc}, V^{f}_{i}]^{T}.
\label{eq:fga_}
\end{equation}

We apply flow-guided attention on SD's U-Net decoder~\cite{cao_2023_masactrl,zhang2023adding} which retains much of the layout and spatial information, in our flow-guided attention, the flow controls the feature tokens across different frames, and together with sharing \emph{key} and \emph{value} patches across different frames, the temporal coherence can be preserved effectively.


Finally, we summarize the algorithm of~\emph{TokenWarping} in Alg.~\ref{alg:TokenWarping}, and demonstrate how to conduct our flow-guided attention during the denoising process. The {Token Warping operator} is defined as the process of warping token features using optical flow and an occlusion map.

\begin{algorithm}[th]
  \caption{\emph{TokenWarping} zero-shot video-to-video translation}
  \label{alg:TokenWarping}
  \begin{algorithmic}[1]
      \State \textbf{Input:}
      \begin{itemize}
          \item $\mathcal{V}$: Input video
          \item $\mathcal{C}$: Control condition
          \item $\mathcal{P^*}$: Target text prompt
      \end{itemize}
      \State \textbf{Output:} Translated video $\mathcal{V^*}$
      \State Estimate optical flow $\mathcal{F}$ and occlusion map $\mathcal{M}$
      \State \revision{Use random Gaussian noise code}
      \State Translate the first frame and store $Q_{1}$, $K_{1}$, $V_{1}$ patches
      \For {$i = 2, 3, \ldots, N$}
          \For {$t = T, T-1, \ldots, 1$}
              \State Get previous $Q_{i-1}, K_{i-1}, V_{i-1}$ patches and first $K_{1}$, $V_{1}$ patches
              \State Update $Q_{i}$ patches using \textbf{Token Warping operator}:
              \[  
                  Q^{\prime}_{i-1} = \texttt{W}(Q_{i-1}, f_{i\Rightarrow i-1}), 
              \]
              \[
                  Q^{f}_{i} = m_{i\Rightarrow i-1} \cdot Q^{\prime}_{i-1} + (1-m_{i\Rightarrow i-1}) \cdot Q_{i},
              \]
              \State \revision{Update $K_{i}, V_{i}$ patches:}
              \[  
                \revision{K^{\prime}_{i-1} = \texttt{W}(K_{i-1}, f_{i\Rightarrow i-1}),}
                \revision{V^{\prime}_{i-1} = \texttt{W}(V_{i-1}, f_{i\Rightarrow i-1}),} 
              \]
              \[
                \revision{K^{f}_{i} = m_{i\Rightarrow i-1} \cdot K^{\prime}_{i-1} + (1-m_{i\Rightarrow i-1}) \cdot K_{i},}
              \]
              \[
                \revision{V^{f}_{i} = m_{i\Rightarrow i-1} \cdot V^{\prime}_{i-1} + (1-m_{i\Rightarrow i-1}) \cdot V_{i},}
              \]
              \State Concatenate with anchor patches $K_{anc}$ and $V_{anc}$
              \State Compute the self-attention output $\operatorname{FGAttn}_i$ using
              \[
                  \operatorname{FGAttn}_i=\operatorname{SoftMax}(\frac{Q^{f}_i [K_{anc}, K^{f}_{i}]^{T} }{\sqrt{d}})\cdot [V_{anc}, V^{f}_{i}]^{T}.
              \]
          \EndFor
          \State Decode latent $\hat{Z}_{i}$ to get the $i$-th translated frame $\mathcal{\hat{V}}_{i}$
      \EndFor
  \end{algorithmic}
\end{algorithm}

\section{Experiments}
\label{sec:experiments}

\begin{figure*}[ht]
\centering
    \captionsetup[subfloat]{labelformat=empty,justification=centering}
    
    \begin{minipage}[c]{0.02\linewidth}
        \rotatebox{90}{\small Source}
    \end{minipage}%
    \begin{minipage}[c]{0.98\linewidth}
        \includegraphics[width=0.105\linewidth]{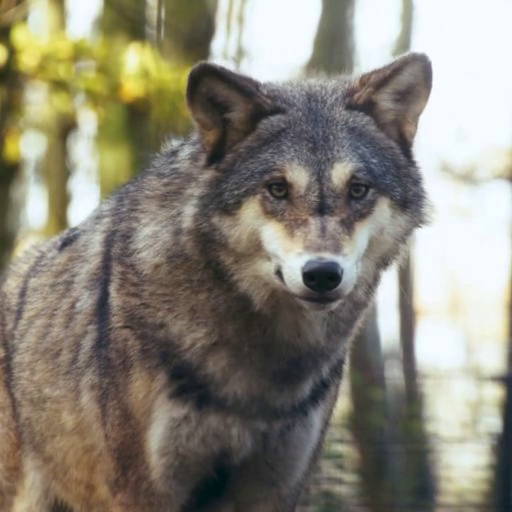}%
        \hspace{0.01mm}
        \includegraphics[width=0.105\linewidth]{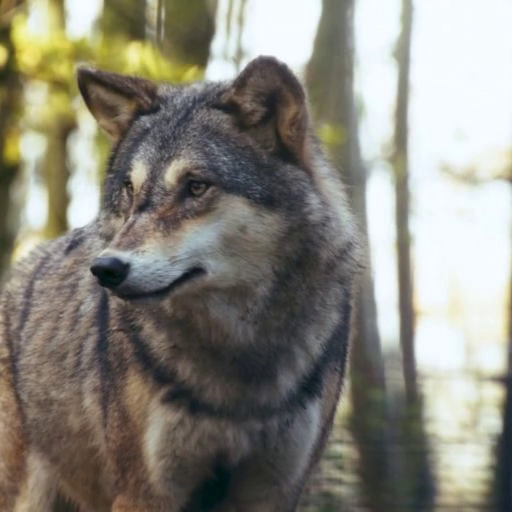}%
        \hspace{0.01mm}
        \includegraphics[width=0.105\linewidth]{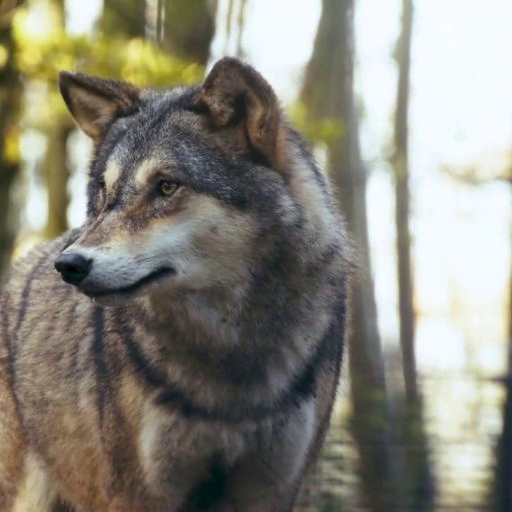}
        \hspace{0.01\linewidth} 
        \includegraphics[width=0.105\linewidth]{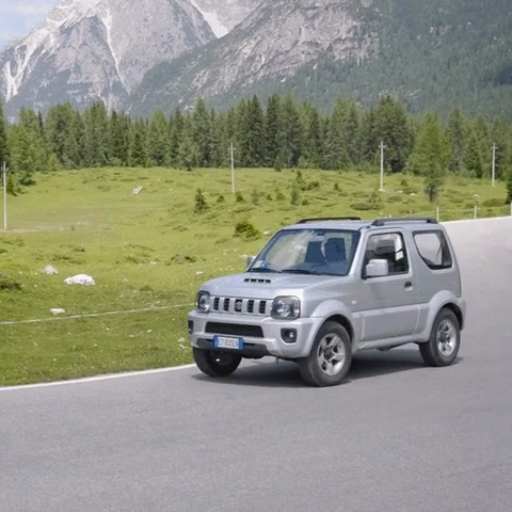}%
        \hspace{0.01mm}
        \includegraphics[width=0.105\linewidth]{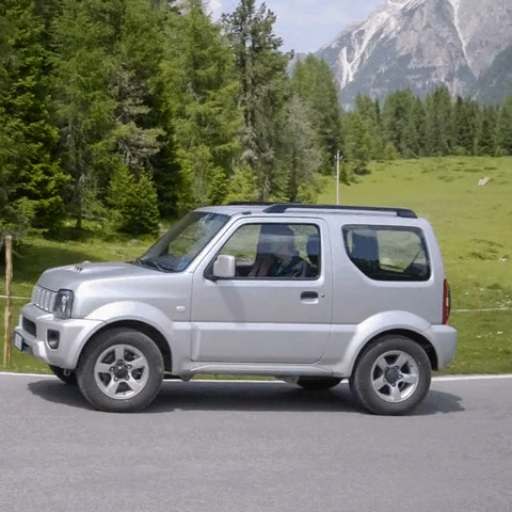}%
        \hspace{0.01mm}
        \includegraphics[width=0.105\linewidth]{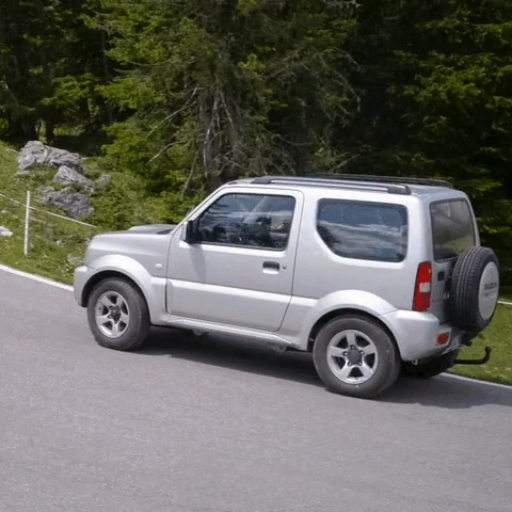}
        \hspace{0.01\linewidth} 
        \includegraphics[width=0.105\linewidth]{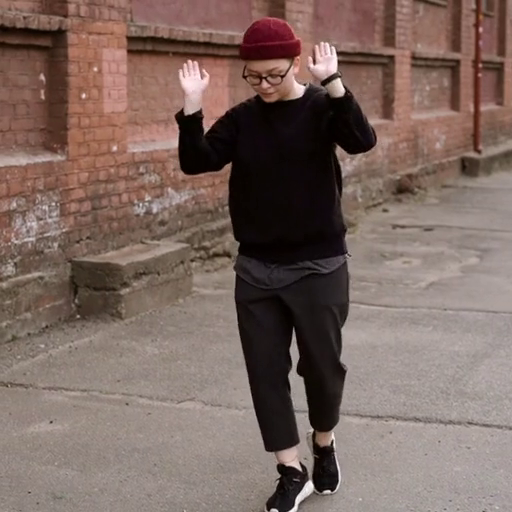}%
        \hspace{0.01mm}
        \includegraphics[width=0.105\linewidth]{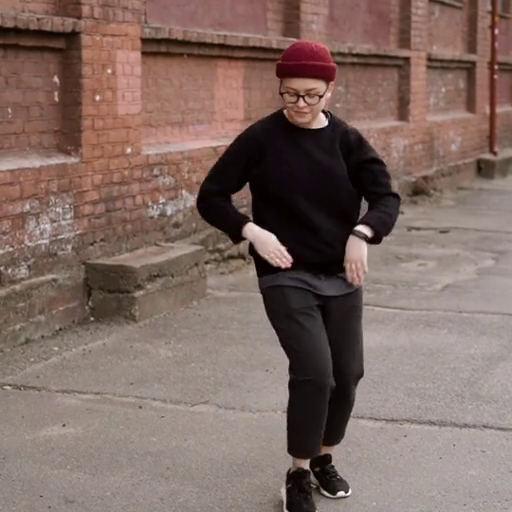}%
        \hspace{0.01mm}
        \includegraphics[width=0.105\linewidth]{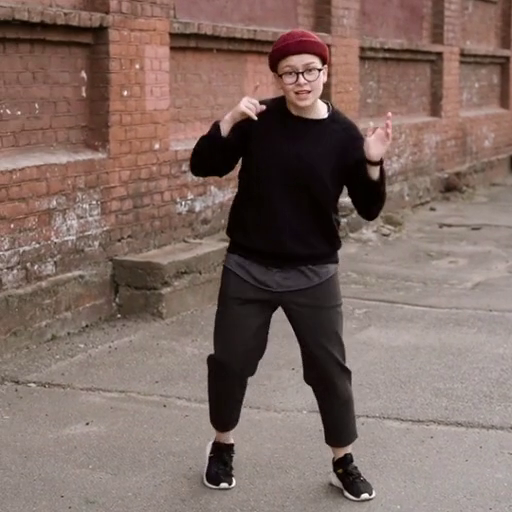}
    \end{minipage}

    \vspace{1mm} 

    \begin{minipage}[c]{0.02\linewidth}
        \rotatebox{90}{\small T2V-Zero}
    \end{minipage}%
    \begin{minipage}[c]{0.98\linewidth}
        \includegraphics[width=0.105\linewidth]{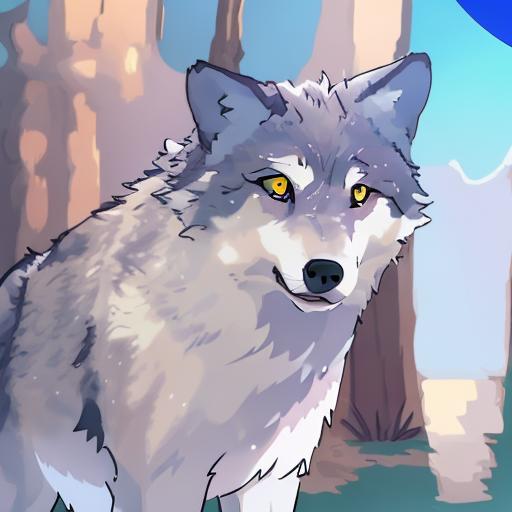}%
        \hspace{0.01mm}
        \includegraphics[width=0.105\linewidth]{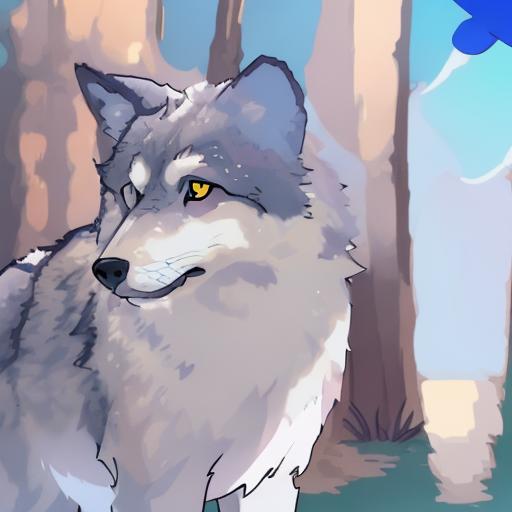}%
        \hspace{0.01mm}
        \includegraphics[width=0.105\linewidth]{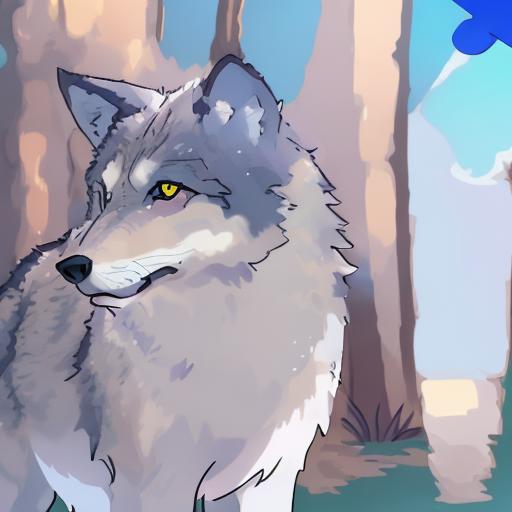}
        \hspace{0.01\linewidth} 
        \includegraphics[width=0.105\linewidth]{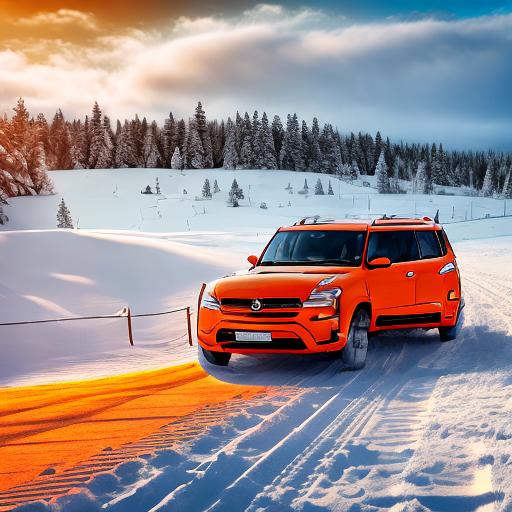}%
        \hspace{0.01mm}
        \includegraphics[width=0.105\linewidth]{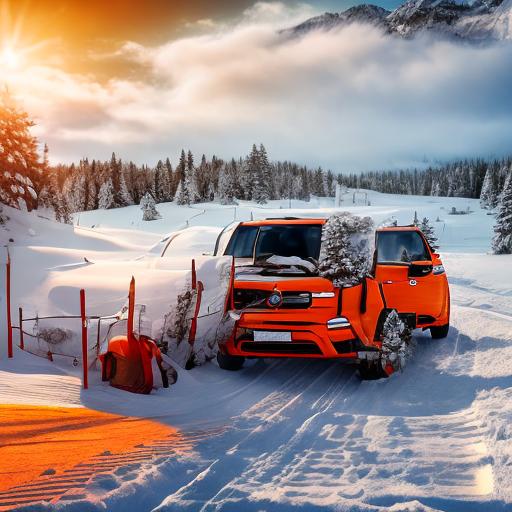}%
        \hspace{0.01mm}
        \includegraphics[width=0.105\linewidth]{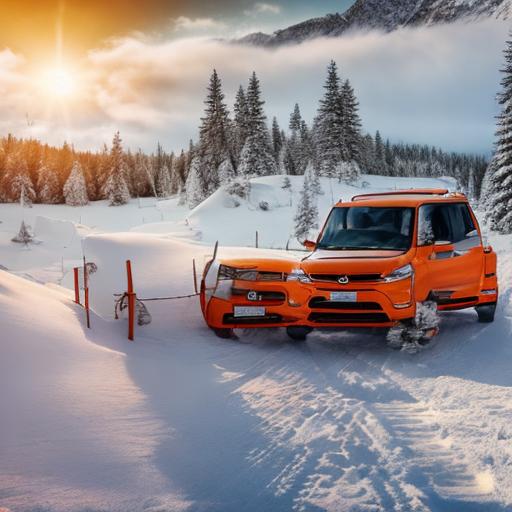}
        \hspace{0.01\linewidth} 
        \includegraphics[width=0.105\linewidth]{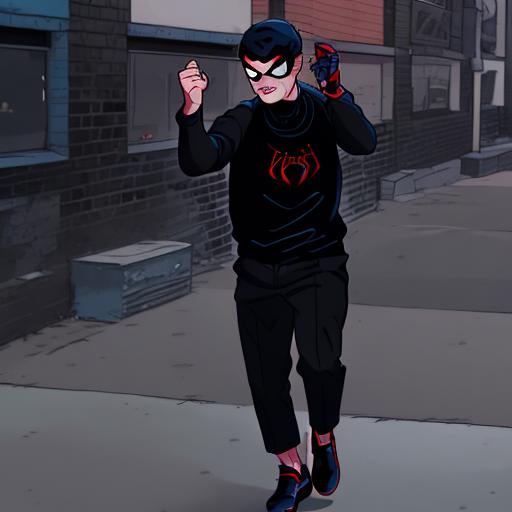}%
        \hspace{0.01mm}
        \includegraphics[width=0.105\linewidth]{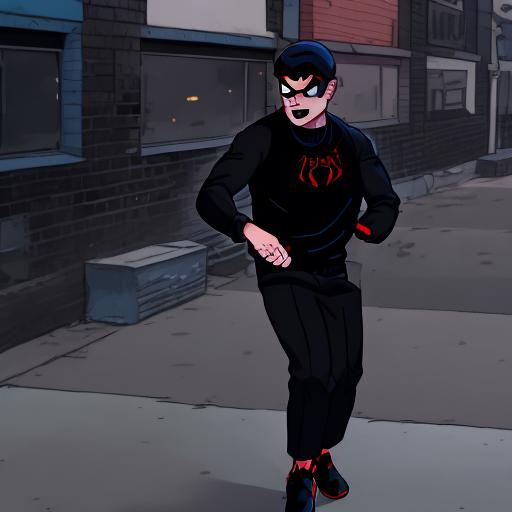}%
        \hspace{0.01mm}
        \includegraphics[width=0.105\linewidth]{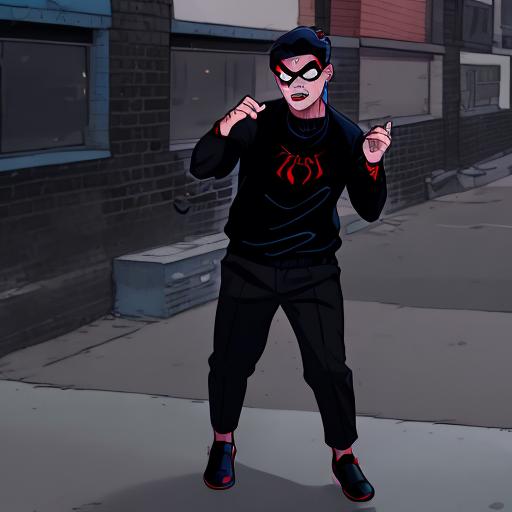}
    \end{minipage}

    \vspace{1mm} 

    \begin{minipage}[c]{0.02\linewidth}
        \rotatebox{90}{\small ControlVideo}
    \end{minipage}%
    \begin{minipage}[c]{0.98\linewidth}
        \includegraphics[width=0.105\linewidth]{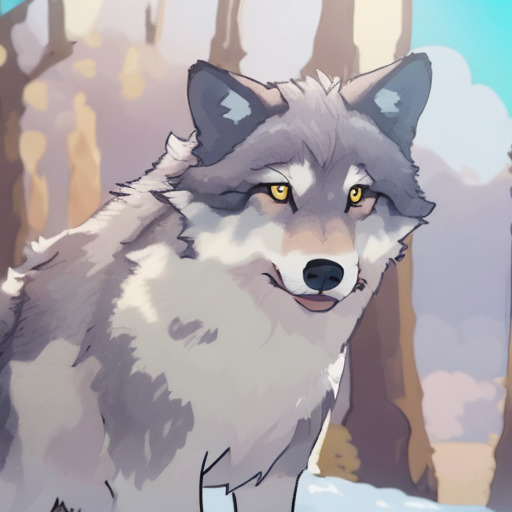}%
        \hspace{0.01mm}
        \includegraphics[width=0.105\linewidth]{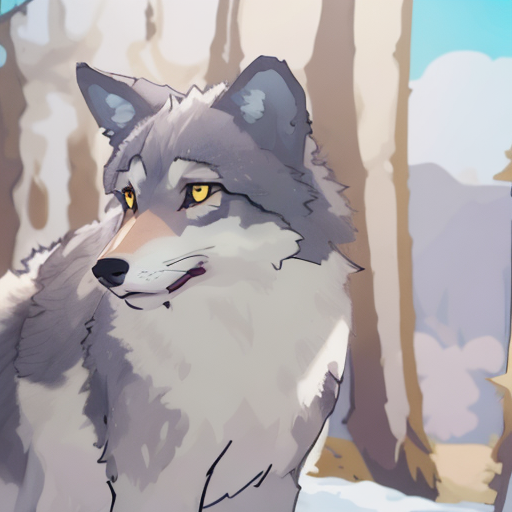}%
        \hspace{0.01mm}
        \includegraphics[width=0.105\linewidth]{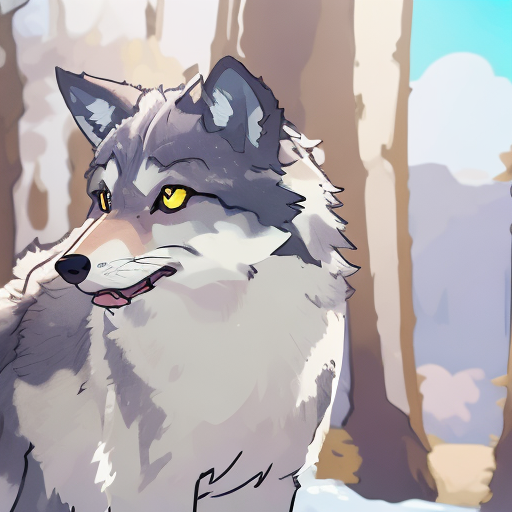}
        \hspace{0.01\linewidth} 
        \includegraphics[width=0.105\linewidth]{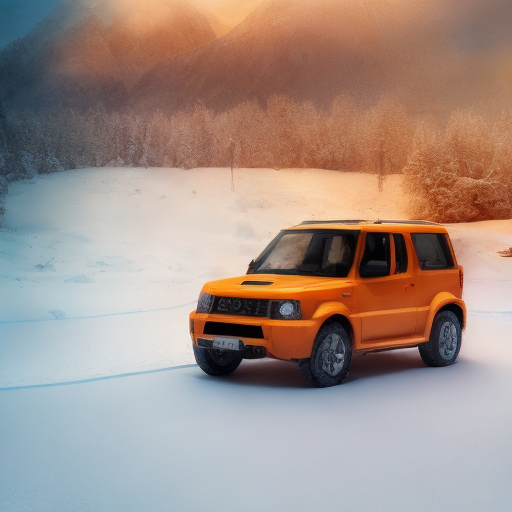}%
        \hspace{0.01mm}
        \includegraphics[width=0.105\linewidth]{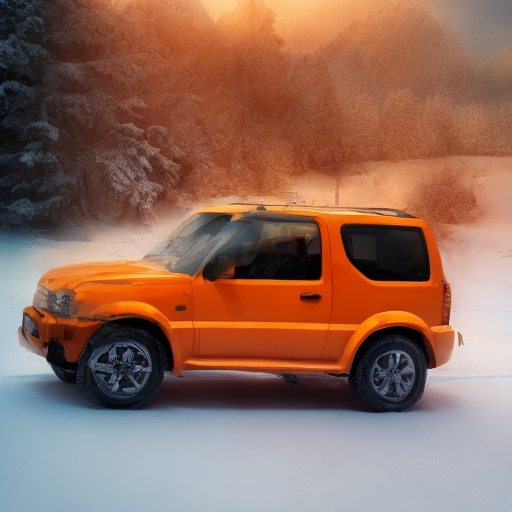}%
        \hspace{0.01mm}
        \includegraphics[width=0.105\linewidth]{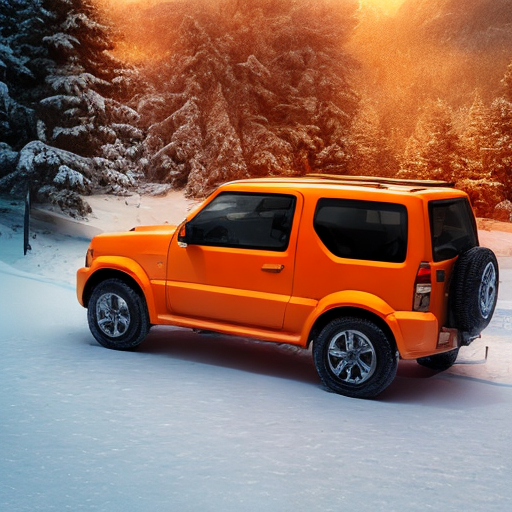}
        \hspace{0.01\linewidth} 
        \includegraphics[width=0.105\linewidth]{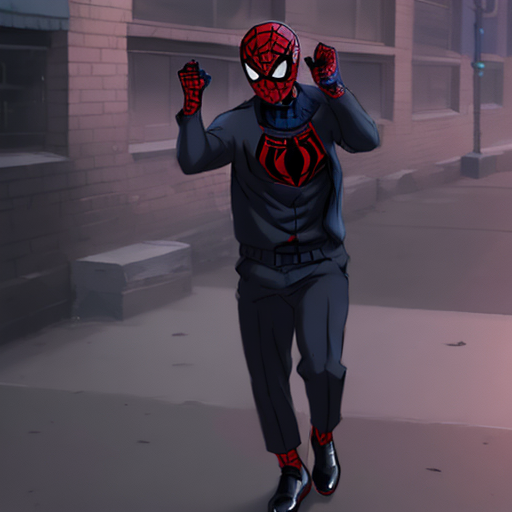}%
        \hspace{0.01mm}
        \includegraphics[width=0.105\linewidth]{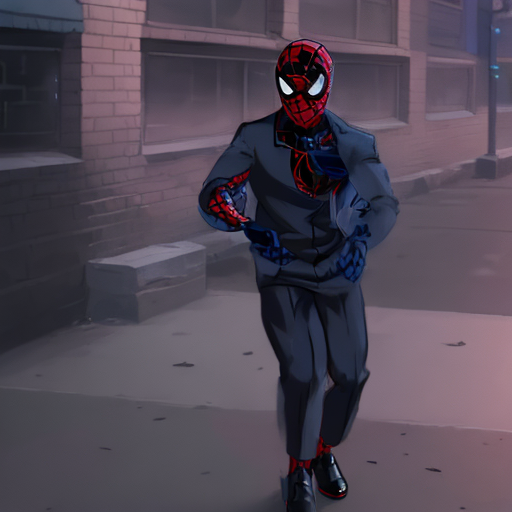}%
        \hspace{0.01mm}
        \includegraphics[width=0.105\linewidth]{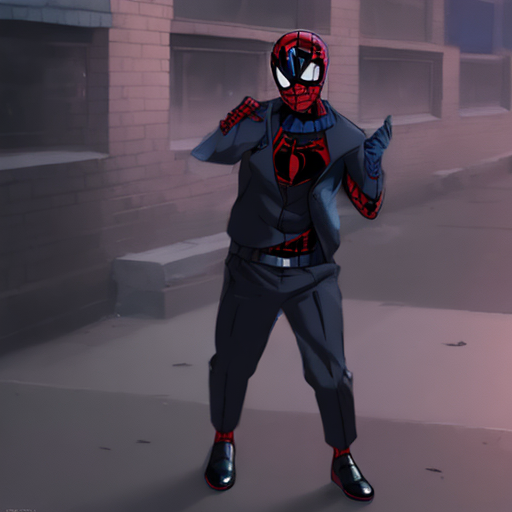}
    \end{minipage}

    \vspace{1mm} 

    \begin{minipage}[c]{0.02\linewidth}
        \rotatebox{90}{\small Rerender}
    \end{minipage}%
    \begin{minipage}[c]{0.98\linewidth}
        \includegraphics[width=0.105\linewidth]{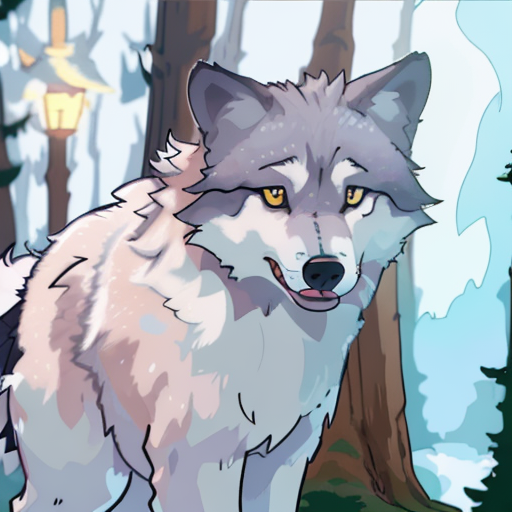}%
        \hspace{0.01mm}
        \includegraphics[width=0.105\linewidth]{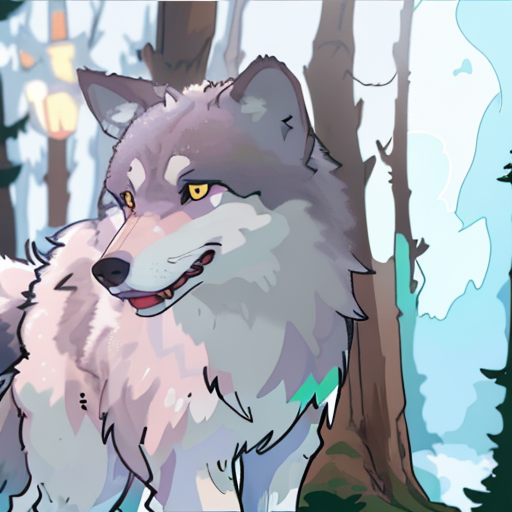}%
        \hspace{0.01mm}
        \includegraphics[width=0.105\linewidth]{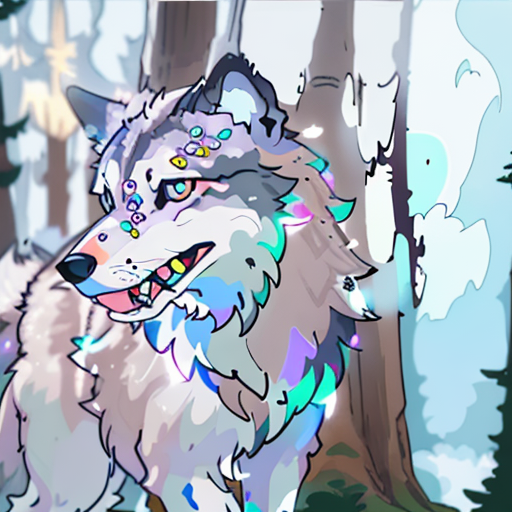}
        \hspace{0.01\linewidth} 
        \includegraphics[width=0.105\linewidth]{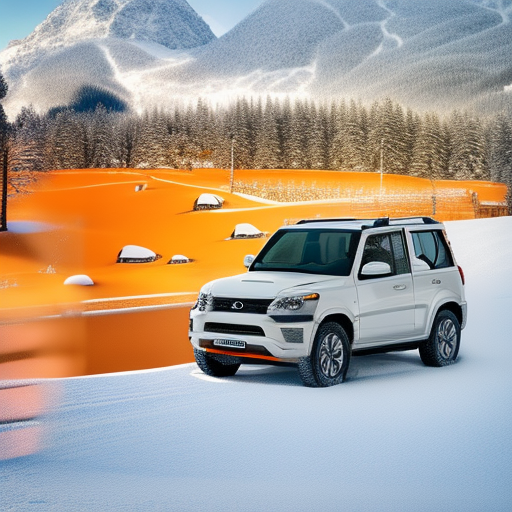}%
        \hspace{0.01mm}
        \includegraphics[width=0.105\linewidth]{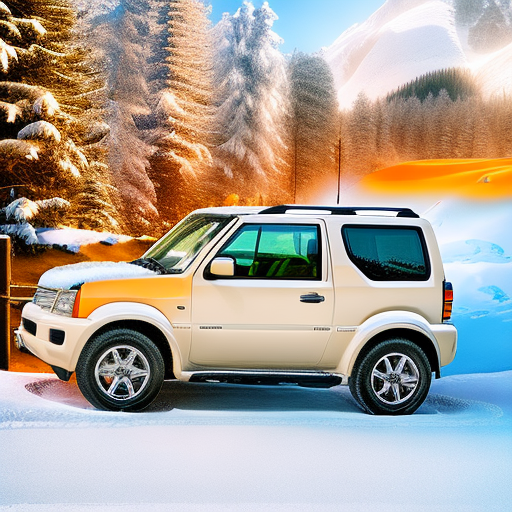}%
        \hspace{0.01mm}
        \includegraphics[width=0.105\linewidth]{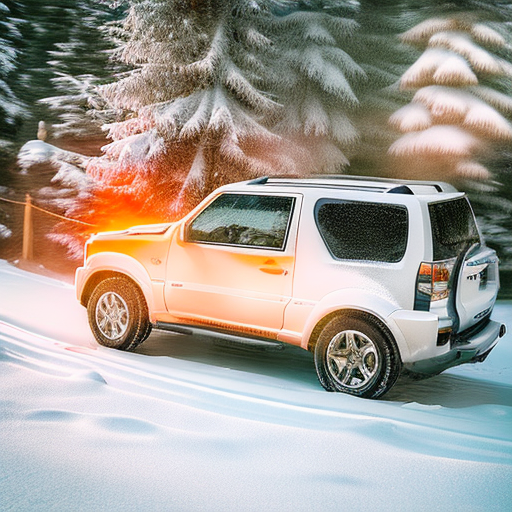}
        \hspace{0.01\linewidth} 
        \includegraphics[width=0.105\linewidth]{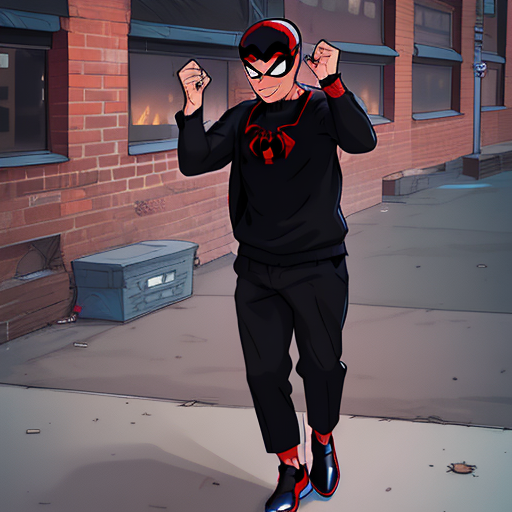}%
        \hspace{0.01mm}
        \includegraphics[width=0.105\linewidth]{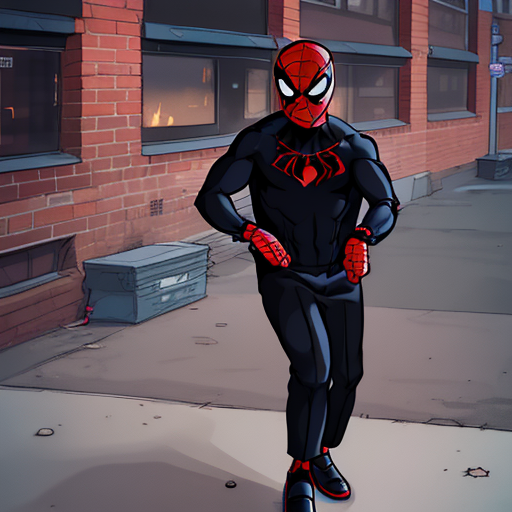}%
        \hspace{0.01mm}
        \includegraphics[width=0.105\linewidth]{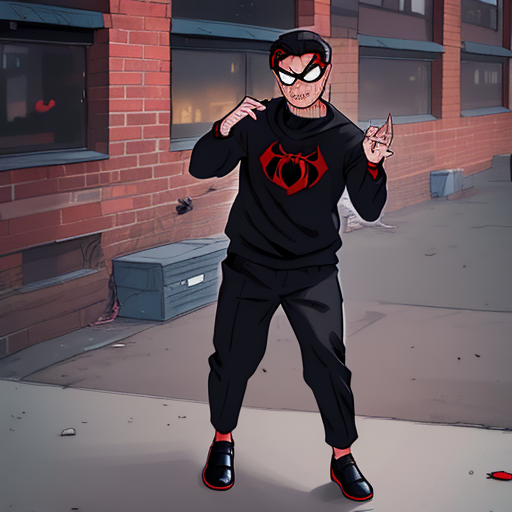}
    \end{minipage}

    \vspace{1mm} 

    \begin{minipage}[c]{0.02\linewidth}
        \rotatebox{90}{\small FRESCO}
    \end{minipage}%
    \begin{minipage}[c]{0.98\linewidth}
        \includegraphics[width=0.105\linewidth]{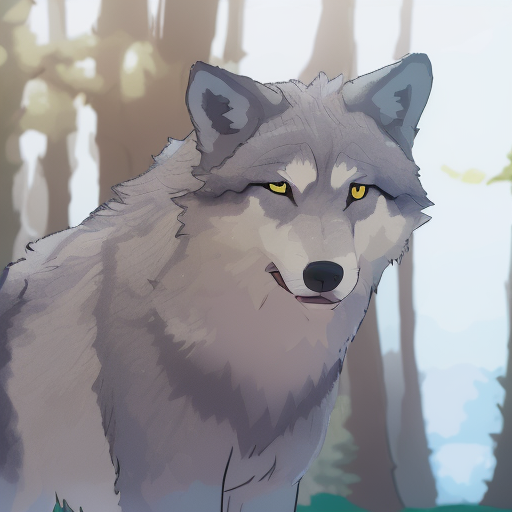}%
        \hspace{0.01mm}
        \includegraphics[width=0.105\linewidth]{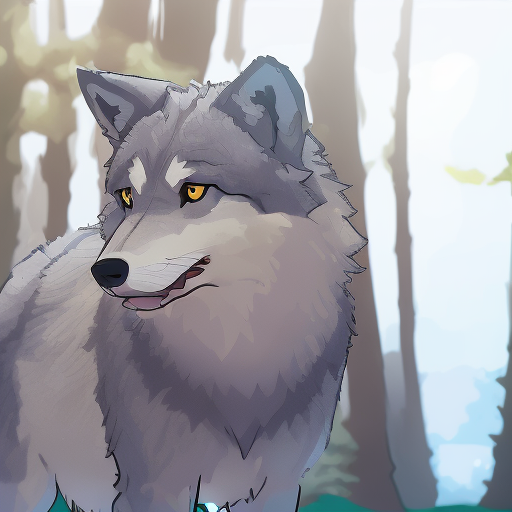}%
        \hspace{0.01mm}
        \includegraphics[width=0.105\linewidth]{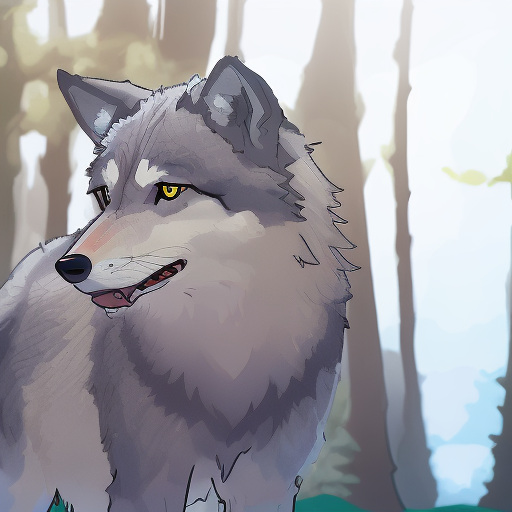}
        \hspace{0.01\linewidth} 
        \includegraphics[width=0.105\linewidth]{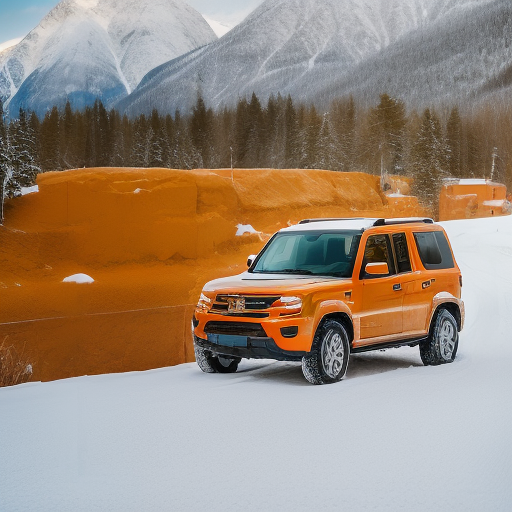}%
        \hspace{0.01mm}
        \includegraphics[width=0.105\linewidth]{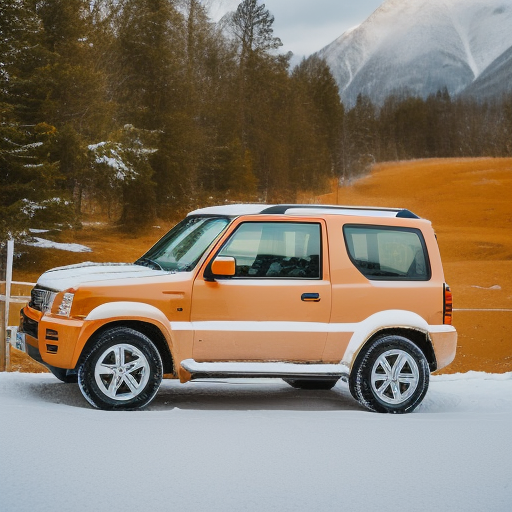}%
        \hspace{0.01mm}
        \includegraphics[width=0.105\linewidth]{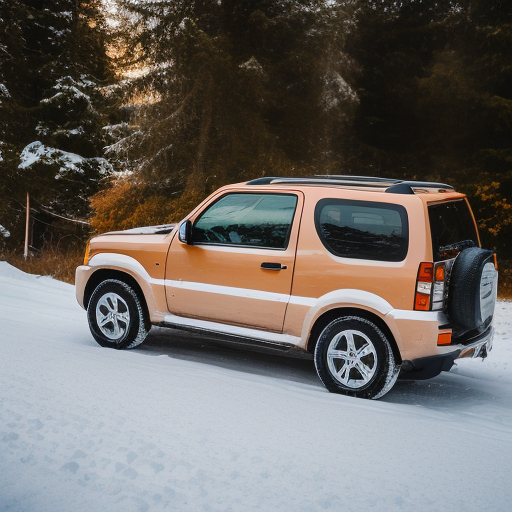}
        \hspace{0.01\linewidth} 
        \includegraphics[width=0.105\linewidth]{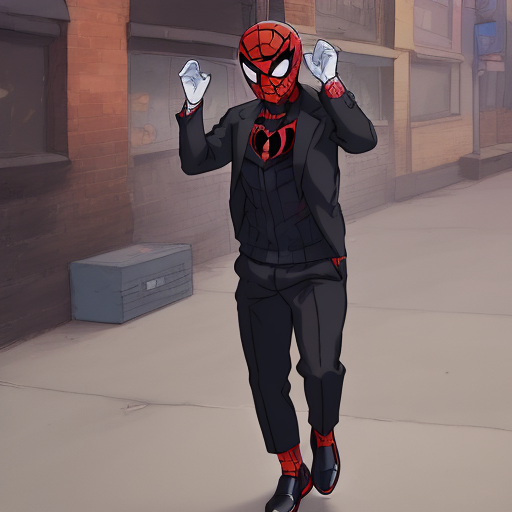}%
        \hspace{0.01mm}
        \includegraphics[width=0.105\linewidth]{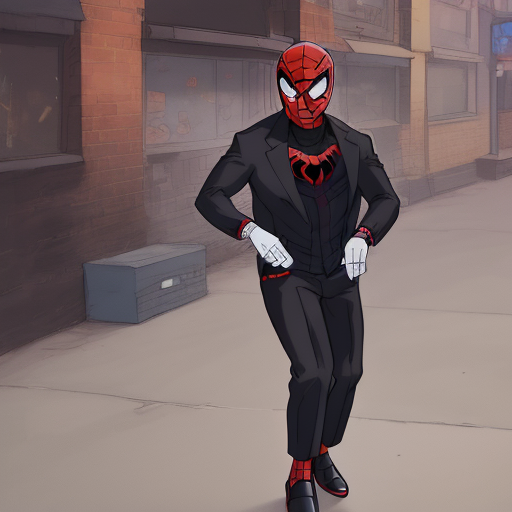}%
        \hspace{0.01mm}
        \includegraphics[width=0.105\linewidth]{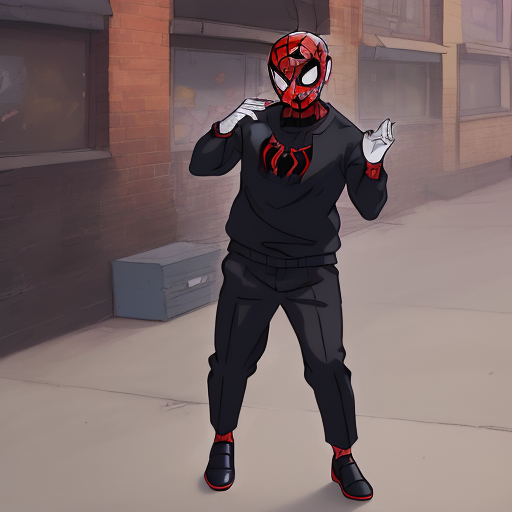}
    \end{minipage}

    \vspace{1mm} 

    \begin{minipage}[c]{0.02\linewidth}
        \rotatebox{90}{\small Ours}
    \end{minipage}%
    \begin{minipage}[c]{0.98\linewidth}
        \includegraphics[width=0.105\linewidth]{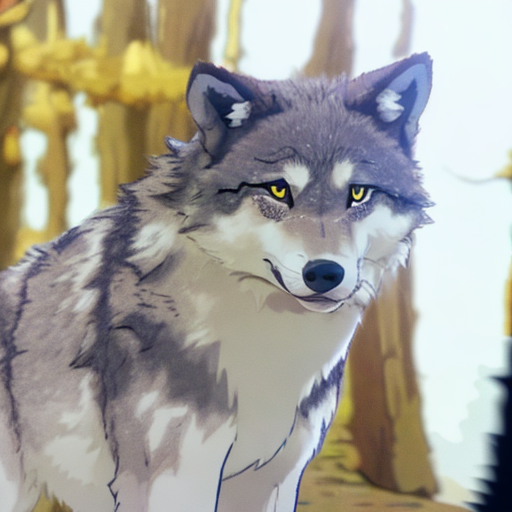}%
        \hspace{0.01mm}
        \includegraphics[width=0.105\linewidth]{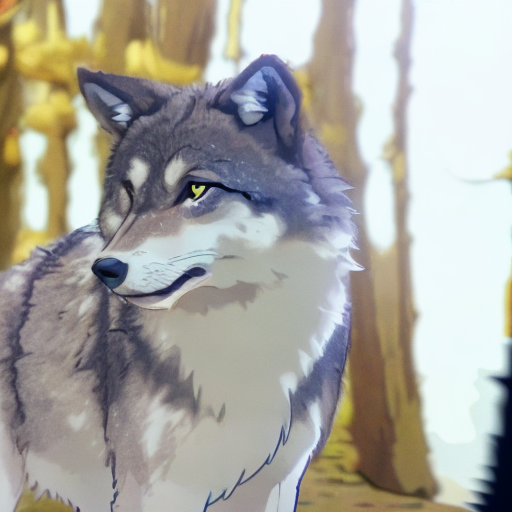}%
        \hspace{0.01mm}
        \includegraphics[width=0.105\linewidth]{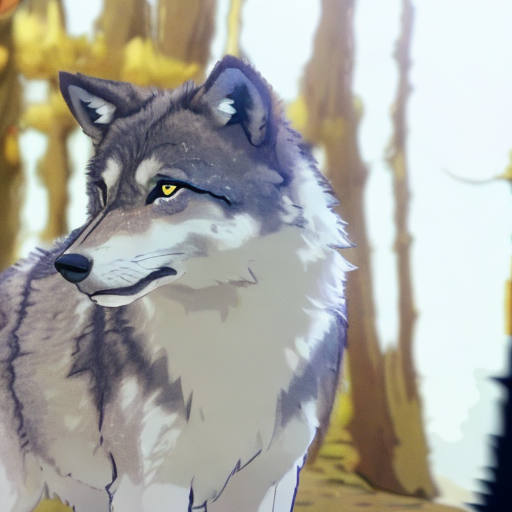}
        \hspace{0.01\linewidth} 
        \includegraphics[width=0.105\linewidth]{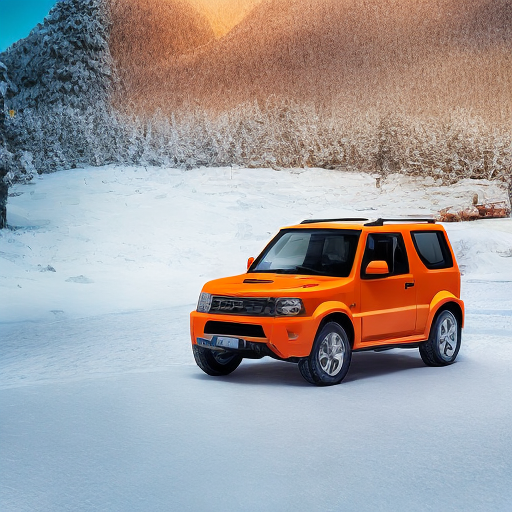}%
        \hspace{0.01mm}
        \includegraphics[width=0.105\linewidth]{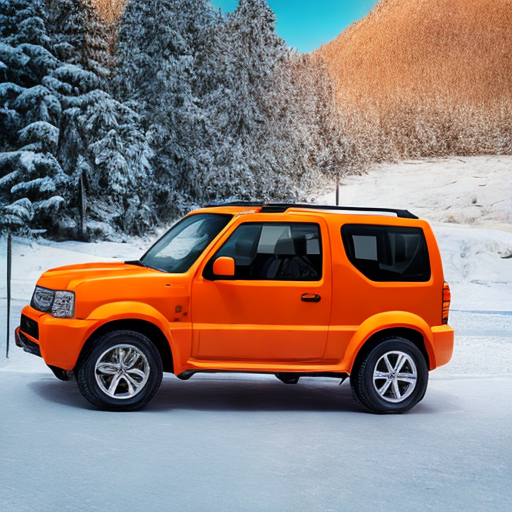}%
        \hspace{0.01mm}
        \includegraphics[width=0.105\linewidth]{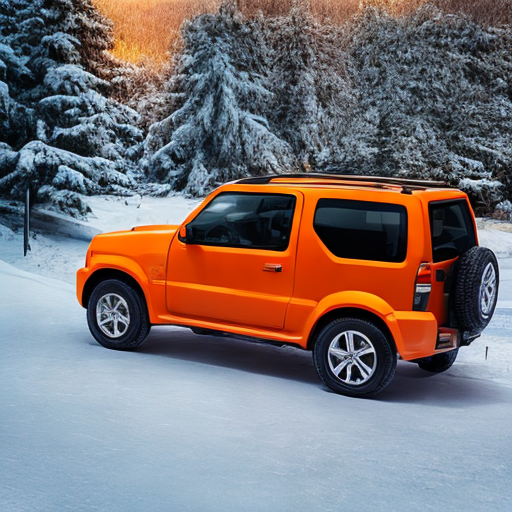}
        \hspace{0.01\linewidth} 
        \includegraphics[width=0.105\linewidth]{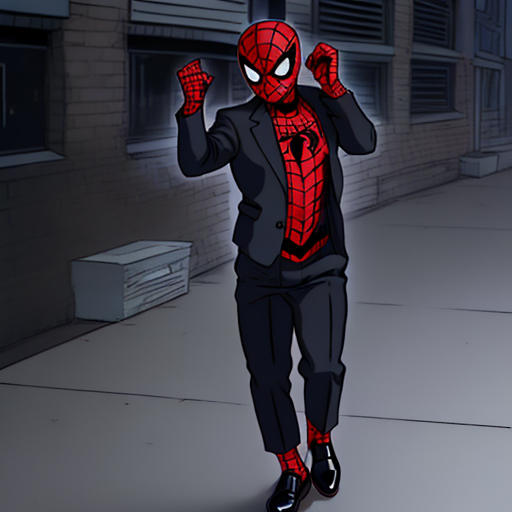}%
        \hspace{0.01mm}
        \includegraphics[width=0.105\linewidth]{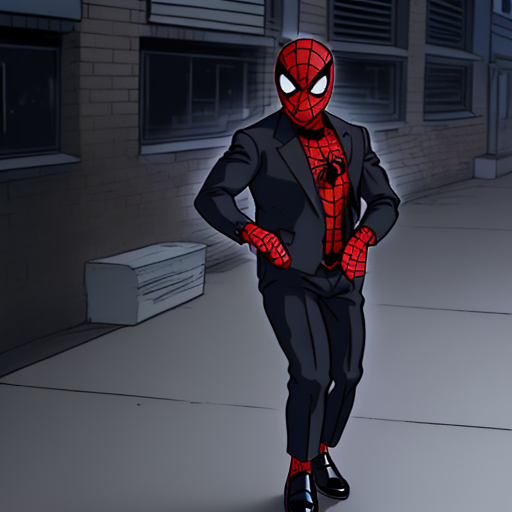}%
        \hspace{0.01mm}
        \includegraphics[width=0.105\linewidth]{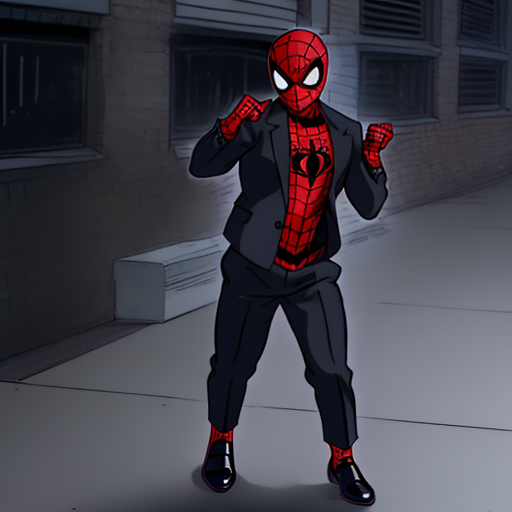}
    \end{minipage}

    \vspace{1mm} 
    \hspace{0.01\linewidth}
    \begin{minipage}[c]{0.30\linewidth}
        \centering
        \fontsize{8.6pt}{11pt}\selectfont
        \textit{Prompt: A hand-drawn animation of a wolf}
    \end{minipage}%
    \hspace{0.03\linewidth}
    \begin{minipage}[c]{0.30\linewidth}
        \centering
        \fontsize{8.6pt}{11pt}\selectfont
        \textit{Prompt: Orange SUV in sunny snow winter}
    \end{minipage}%
    \hspace{0.03\linewidth}
    \begin{minipage}[c]{0.30\linewidth}
        \centering
        \fontsize{8.6pt}{11pt}\selectfont
        \textit{Prompt: A cartoon spiderman in black suit}
    \end{minipage}

    \vspace{-1mm}
    \caption{Qualitative comparisons with zero-shot video methods. \emph{TokenWarping} aligns with the video structure and target prompt.}
    \vspace{-3mm}
    \label{fig:com_teaser}
\end{figure*}

\subsection{Implementation Details}
\label{subsec:Implementation Details}

We collect 40 videos from the Internet and previous works~\cite{yang2023rerender,yang2024fresco,zhang2023controlvideo}, which consist of human motions videos and slow motion videos. Then we manually add the caption to form text-video pairs. The Stable Diffusion 1.5~\cite{rombach2022high} and ControlNet 1.0~\cite{zhang2023adding} are adopted in our framework. Following previous work, we sample uniformly from the video. During sampling, a DDIM sampler with 50 steps and 7.5 classifier-free guidance is used for inference.

We set the anchor frame to the first frame by default, and use bilinear interpolation for backward warping. For the conditional input, we use edge maps, depth, and canny edge maps as conditional input. More results and long-term translated videos are available in the supplementary material. All the source code and datasets will be released.

\subsection{Metrics} For quantitative evaluation, adhering to standard practices~\cite{cong2023flatten, QI_2023_ICCV, esser2023structure}, three metrics are utilized to assess text alignment, temporal consistency, and pixel-alignment, which includes: \romannumeral1) Editing Accuracy (\textbf{Edit-Acc}): This metric measures the frame-wise editing accuracy, representing the percentage of frames where the edited image has a higher CLIP similarity to the target prompt than to the source prompt. A successful editing is indicated if the target similarity exceeds the source similarity.
\romannumeral2) Temporal Consistency (\textbf{Tem-Con}): This metric computes CLIP image embeddings on all frames of output videos and report the average cosine similarity between all pairs of consecutive frames.
\romannumeral3) Warp Error (\textbf{Warp-Err}): This metric calculates the average mean-squared pixel-level difference between edited consecutive frames. Specifically, the optical flow between source consecutive frames is computed, and each frame in the edited video is warped to the next using the flow. The average mean-squared pixel error is then calculated between each warped frame and its corresponding target frame.  

\vspace{1em}

\noindent \textbf{Competitors.} We compare our \emph{TokenWarping} method with several zero-shot related works. {Notably, one-shot tuning methods like TAV~\cite{wu2023tune} are not included in the comparison list.} The zero-shot methods include Rerender~\cite{yang2023rerender}, ControlVideo~\cite{zhang2023controlvideo}, Text2Video-Zero~\cite{text2video-zero} and TokenFlow~\cite{geyer2023tokenflow}. We also compare with Flow based methods, including FRESCO~\cite{yang2024fresco}, FLATTEN~\cite{cong2023flatten}. They all use optical flow to warp the \emph{key} and \emph{value} patches in the self-attention.

\begin{table}[t]
    \caption{Quantitative Comparison with Various Methods\label{tab:com_zero_shot}}
    \centering
    \small
    \setlength{\tabcolsep}{1mm}
    \begin{tabular}{lcccc}
    \toprule
    Methods          &Tem-Con $\uparrow$    &Edit-Acc $\uparrow$      &Warp-Err $\downarrow$  & User $\uparrow$\\
    \midrule
    T2V-Zero         &0.9609                &0.9164                   &0.0235                 & 1.9\%  \\
    ControlVideo     &0.9765                &0.8685                   &0.0102                 & 5.8\%   \\
    Rerender         &0.9799                &0.8783                   &0.0132                 & 15.6\%   \\
    FRESCO           &0.9843                &0.8669                   &0.0078                 & 29.4\%   \\
    \cmidrule(lr){1-5} 
    TokenFlow        &0.9726                &0.7925                   &\textbf{0.0039}        & 3.9\%   \\
    FLATTEN          &0.9648                &0.8259                   &0.0150                 & 7.8\%   \\
    \cmidrule(lr){1-5} 
    Ours             &\textbf{0.9868}       &\textbf{0.9488}          &0.0089            & \textbf{35.2}\%  \\
    \bottomrule
    \end{tabular}
    \vspace{-3mm}
    \end{table}

\subsection{Qualitative Comparison}
We first present the qualitative comparison with zero-shot methods in Fig.~\ref{fig:com_teaser}, Text2Video-Zero~\cite{text2video-zero} does not support long-term motion and produces poor editing results in the ``car'' sequences. Other methods successfully translate videos according to the provided text prompts. However, we can observe that ControlVideo~\cite{zhang2023controlvideo} captures the structure of the source sequence but tends to produce blurry results. Rerender~\cite{yang2023rerender} fails to achieve consistent and robust results. FRESCO~\cite{yang2024fresco} achieves excellent spatial correspondences but is unable to maintain the correct color for subsequent frames in ``car'' and ``spiderman''. In contrast, our method generates consistent videos and achieves better editing based on warping tokens.

\begin{figure}[t]
\centering
    \captionsetup[subfloat]{labelformat=empty,justification=centering}

    \begin{minipage}[c]{0.04\linewidth}
        \rotatebox{90}{\small Source}
    \end{minipage}%
    \begin{minipage}[c]{0.96\linewidth}
        \includegraphics[width=0.24\linewidth]{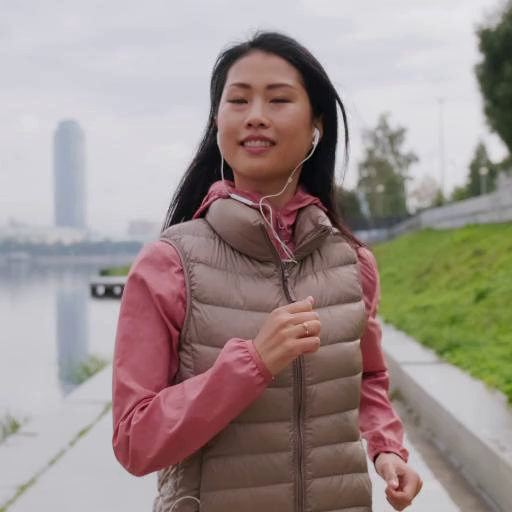}%
        \hspace{0.01mm}
        \includegraphics[width=0.24\linewidth]{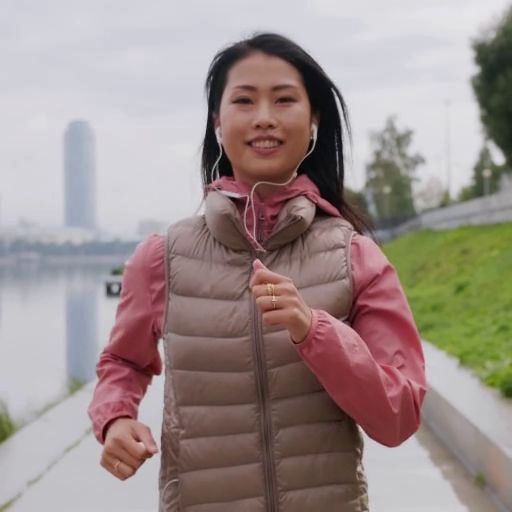}%
        \hspace{0.01mm}
        \includegraphics[width=0.24\linewidth]{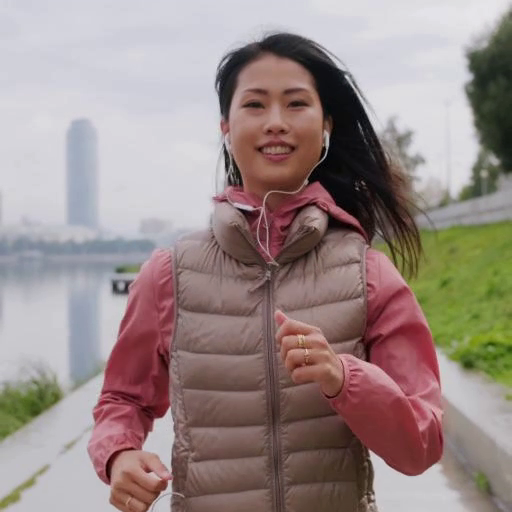}
        \hspace{0.01mm}
        \includegraphics[width=0.24\linewidth]{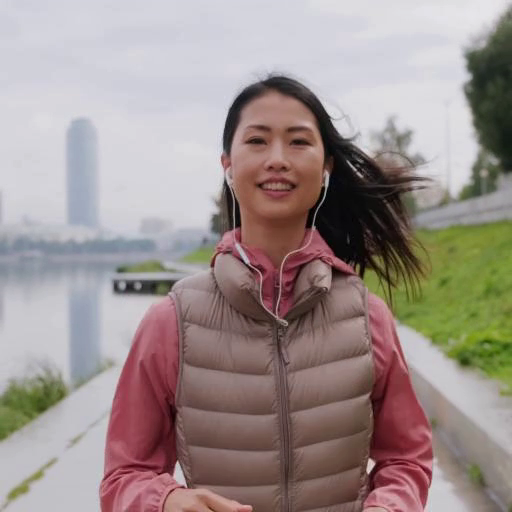}
    \end{minipage}\vspace{0.5mm}





    \begin{minipage}[c]{0.04\linewidth}
        \rotatebox{90}{\small TokenFlow}
    \end{minipage}%
    \begin{minipage}[c]{0.96\linewidth}
        \includegraphics[width=0.24\linewidth]{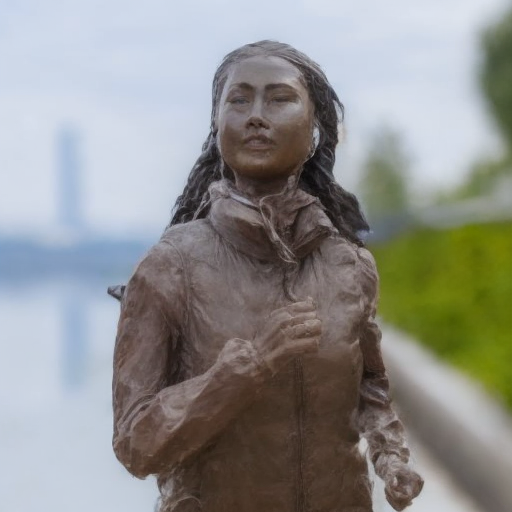}%
        \hspace{0.01mm}
        \includegraphics[width=0.24\linewidth]{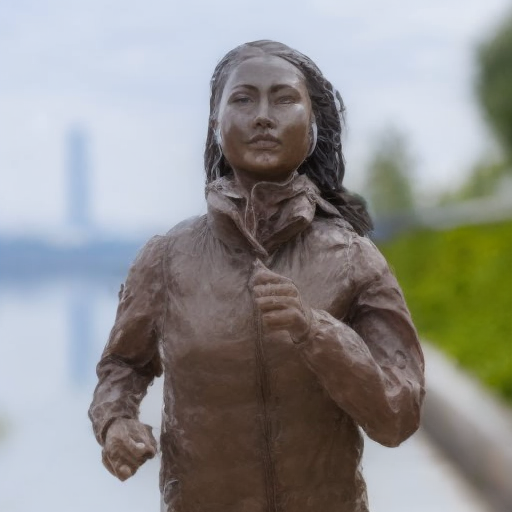}%
        \hspace{0.01mm}
        \includegraphics[width=0.24\linewidth]{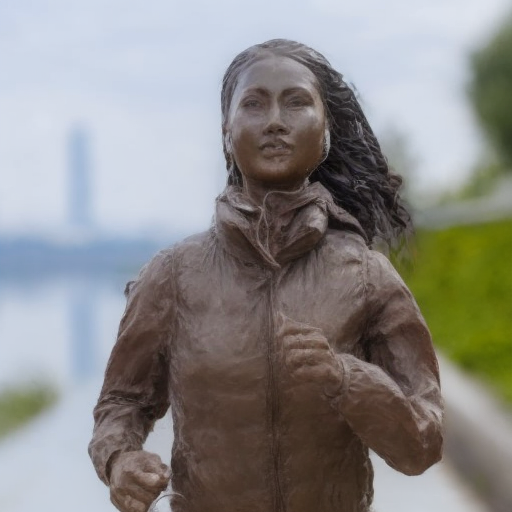}
        \hspace{0.01mm}
        \includegraphics[width=0.24\linewidth]{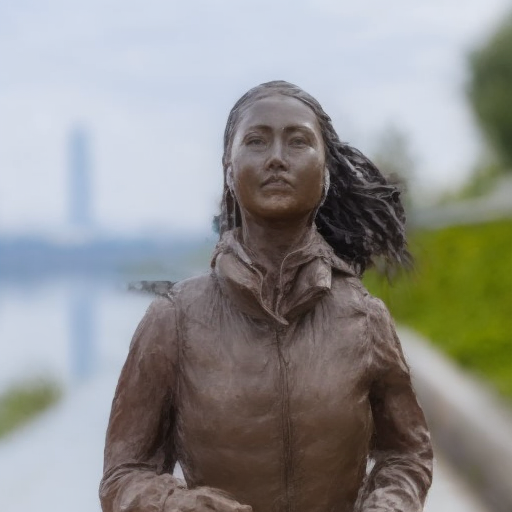}
    \end{minipage}\vspace{0.5mm}

    \begin{minipage}[c]{0.04\linewidth}
        \rotatebox{90}{\small FLATTEN}
    \end{minipage}%
    \begin{minipage}[c]{0.96\linewidth}
        \includegraphics[width=0.24\linewidth]{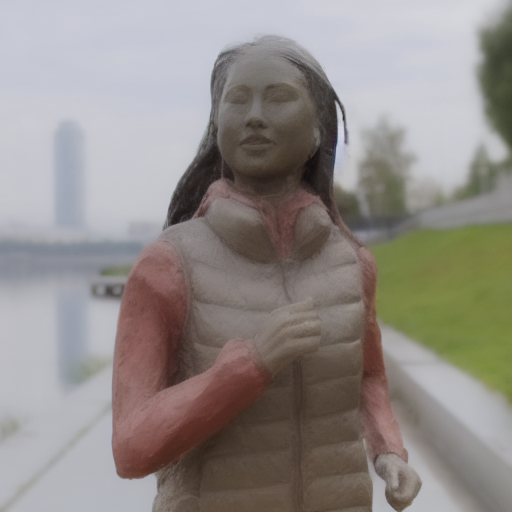}%
        \hspace{0.01mm}
        \includegraphics[width=0.24\linewidth]{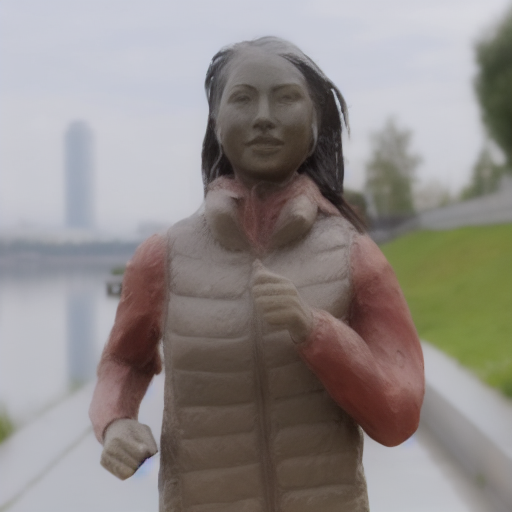}%
        \hspace{0.01mm}
        \includegraphics[width=0.24\linewidth]{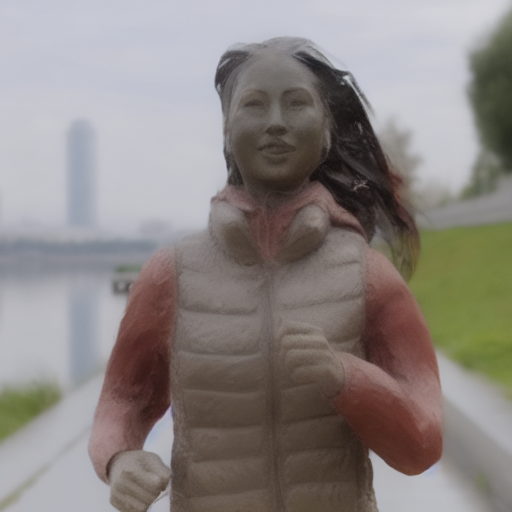}
        \hspace{0.01mm}
        \includegraphics[width=0.24\linewidth]{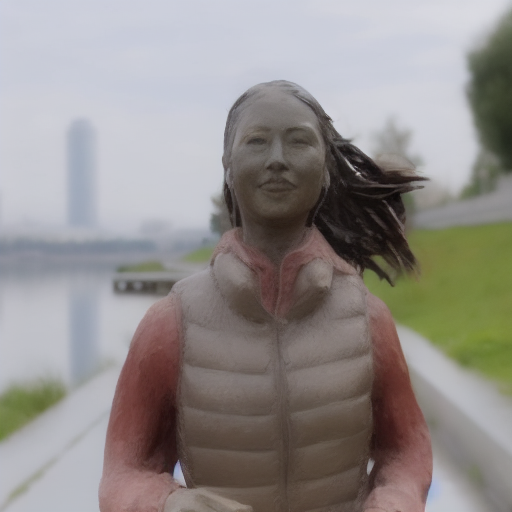}
    \end{minipage}\vspace{0.5mm}

    \begin{minipage}[c]{0.04\linewidth}
        \rotatebox{90}{\small FRESCO}
    \end{minipage}%
    \begin{minipage}[c]{0.96\linewidth}
        \includegraphics[width=0.24\linewidth]{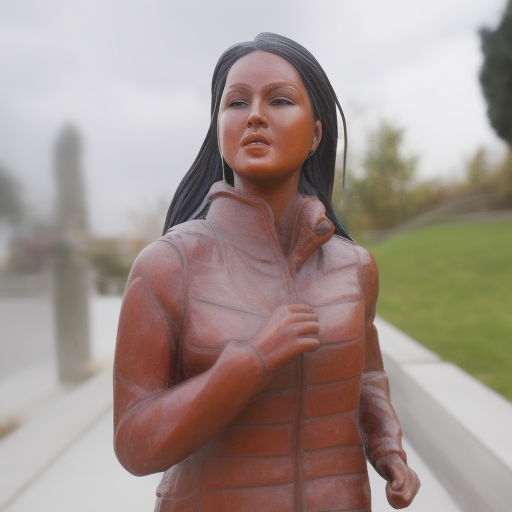}%
        \hspace{0.01mm}
        \includegraphics[width=0.24\linewidth]{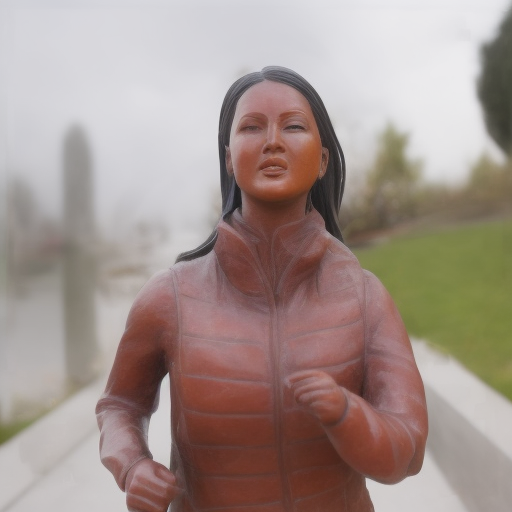}%
        \hspace{0.01mm}
        \includegraphics[width=0.24\linewidth]{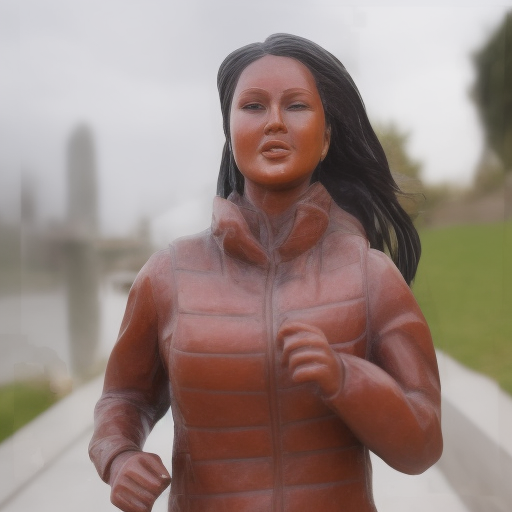}
        \hspace{0.01mm}
        \includegraphics[width=0.24\linewidth]{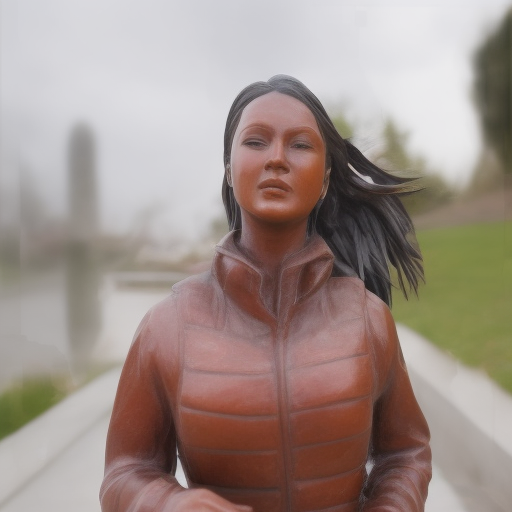}
    \end{minipage}\vspace{0.5mm}

    \begin{minipage}[c]{0.04\linewidth}
        \rotatebox{90}{\small Ours}
    \end{minipage}%
    \begin{minipage}[c]{0.96\linewidth}
        \includegraphics[width=0.24\linewidth]{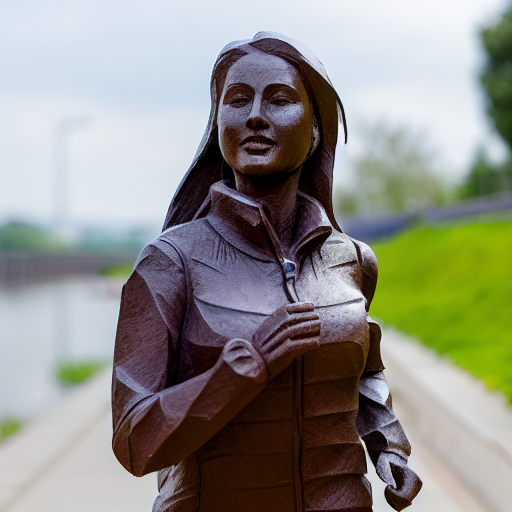}%
        \hspace{0.01mm}
        \includegraphics[width=0.24\linewidth]{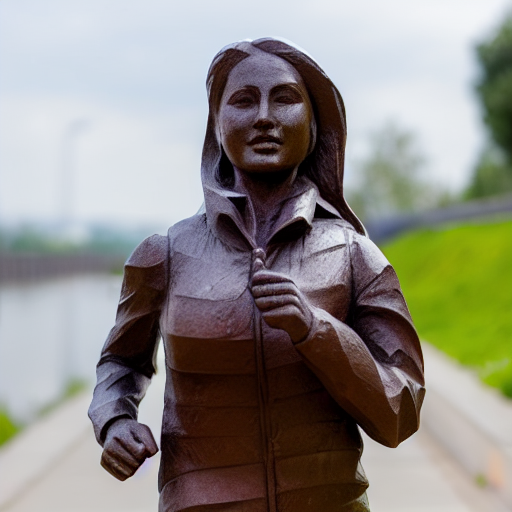}%
        \hspace{0.01mm}
        \includegraphics[width=0.24\linewidth]{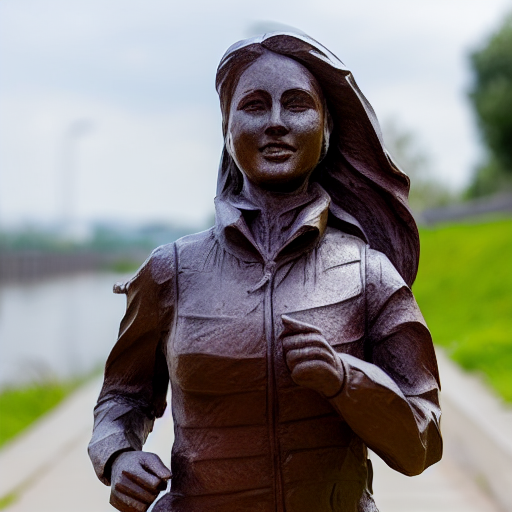}
        \hspace{0.01mm}
        \includegraphics[width=0.24\linewidth]{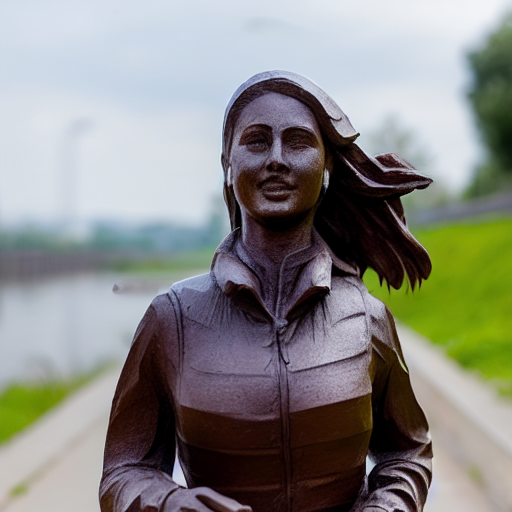}
    \end{minipage}
    \vspace{-1mm}
    \caption{Qualitative comparisons with flow-attention competitors. \textit{Prompt: A sculpture of a woman running.}}
    \vspace{-3mm}
    \label{fig:com_2}
\end{figure}

\revision{We also present a comparison between condition-constrained and inversion-based methods. The inversion-based methods include FLATTEN~\cite{cong2023flatten} and TokenFlow~\cite{geyer2023tokenflow}, which embed source information through DDIM inversion. In contrast, the condition-constrained methods include FRESCO~\cite{yang2024fresco} and our \emph{TokenWarping}, which leverages ControlNet’s structural constraints to preserve the source structure without requiring inversion.}
As shown in Fig.~\ref{fig:com_2}, FLATTEN overfits and retains the pink clothes from the source video, while both TokenFlow and FRESCO fail to accurately capture the subject's expression. In contrast, our method, \emph{TokenWarping}, warps the \emph{query}, \emph{key}, and \emph{value} tokens simultaneously. This approach not only maintains temporal consistency but also preserves the details of the source frame. For the video comparisons, please refer to supplementary.

\begin{table}[t]
    \caption{\revision{Comparison in runtime with 32 frames.}\label{tab:com_runtime}}
    \centering
    \small
    \setlength{\tabcolsep}{1mm}
    \begin{tabular}{lccc}
    \toprule
    Methods                         &DDIM Inv.         &Sampling            &Varm (MiB)           \\
    \midrule
    TokenFlow                       &6min14s           &1min31s             &12964        \\
    FLATTEN                         &5min35s           &3min00s             &32426       \\
    \cmidrule(lr){1-4} 
    ControlVideo                    &-                 &1min55s             &7280        \\
    FRESCO                          &-                 &4min14s             &15956       \\
    \cmidrule(lr){1-4} 
    Ours                            &-                 &\textbf{1min24s}     &20002     \\
    \bottomrule
    \end{tabular}\vspace{-3mm}
    \end{table}

\subsection{Quantitative Comparison}

The quantitative comparison with zero-shot methods is shown in Tab.~\ref{tab:com_zero_shot}, \emph{TokenWarping} achieves a balance between temporal consistency and accurate editing, showing results comparable to FRESCO~\cite{yang2024fresco}. 
Particularly, Text2Video-Zero~\cite{text2video-zero} performs significantly worse in terms of editing accuracy compared to our method.
ControlVideo~\cite{zhang2023controlvideo} tends to generate rough videos, resulting in low editing accuracy.
Rerender~\cite{yang2023rerender} exhibits lower temporal consistency compared to our method due to its less robust editing process.
FRESCO~\cite{yang2024fresco} exhibits better performance on the Warp-Err metric but struggles with shape-related translation. As demonstrated in Fig.~\ref{fig:com_1}, it fails to translate the background into a ``castle''. In contrast, our method shows an 8\% improvement in editing success and delivers better visual results in the attached video compared to FRESCO.

Inversion-based methods like TokenFlow~\cite{geyer2023tokenflow} and FLATTEN~\cite{cong2023flatten} excel in temporal consistency due to their inversion codes. However, as shown in our attached video, they often fail in video editing for specific prompts. TokenFlow achieves the lowest Warp-Err score because the edited video closely resembles the source. In contrast, our \emph{TokenWarping} method balances editing accuracy and temporal consistency.

We also conduct a user study with 50 participants. Participants are tasked with selecting the most preferable results among the seven methods. As shown in Tab.~\ref{tab:com_zero_shot}, our method received the most favored votes.

\subsection{{Runtime Comparison}}

Zero-shot video translation methods can be broadly categorized into two types: \revision{inversion-based (the latent code is obtained from by inversion) and noise-based (latent code is random initialized by Gaussian noise) methods.}
To evaluate the efficiency of our approach, we conducted a comparative runtime experiment with 32 frames at a resolution of 512x512, as detailed in Table~\ref{tab:com_runtime}. Inversion-based methods, such as TokenFlow~\cite{geyer2023tokenflow} and FLATTEN~\cite{cong2023flatten}, require significantly more memory (Varm) and time to process the inversion, whereas noise-code approaches like those described in~\cite{yang2024fresco,zhang2023controlvideo} generate outputs from gaussian noise and rely solely on SD-model reverse sampling.
Notably, flow-based methods such as FRESCO~\cite{yang2024fresco} and FLATTEN~\cite{cong2023flatten} necessitate additional memory to store flow trajectories. In contrast, our method requires only the storage of optical flow maps, which is more memory-efficient. For instance, FRESCO requires \revision{15,956} MiB to store trajectories for 8 frames, and FLATTEN needs \revision{32,426} MiB for 32 frames, while our method uses only \revision{20,002} MiB for 32 frames.
Overall, our method achieves faster speed (1min24s) with superior generation quality (0.9868 Tem-con and 0.9488 Edit-Acc).

\subsection{Ablation Study}
\label{subsec:ablation}

\begin{table}[t]
    \caption{Quantitative Ablation Study\label{tab:com_ab}}
    \centering
    \small
    \setlength{\tabcolsep}{1mm}
    \begin{tabular}{lcccc}
    \toprule
    Metric                    &Baseline     &w/ \emph{Q} Warping  &w/ \emph{KV} Warping   &Full   \\
    \midrule
    Edit-Acc $\uparrow$       &1.0000       &1.0000     &1.0000          &1.0000     \\
    Tem-Con $\uparrow$        &0.9775       &0.9731     &0.9833          &\textbf{0.9849}     \\
    Warp-Err $\downarrow$     &0.0154       &0.0105     &0.0096          &\textbf{0.0089}    \\
    \bottomrule
\end{tabular}
\vspace{-3mm}
\end{table}

\begin{figure}[t]
    \centering
        \captionsetup[subfloat]{labelformat=empty,justification=centering}
        \subfloat[ (a) Input]{
        \begin{minipage}{0.185\linewidth} 
        \includegraphics[width=\linewidth]{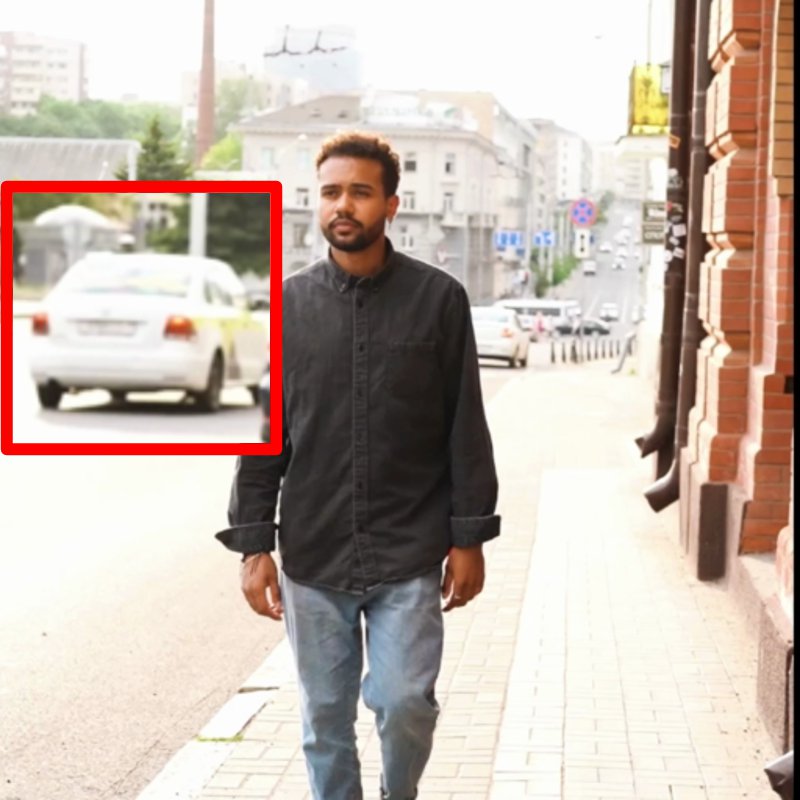}\vspace{0.5mm}
        \includegraphics[width=\linewidth]{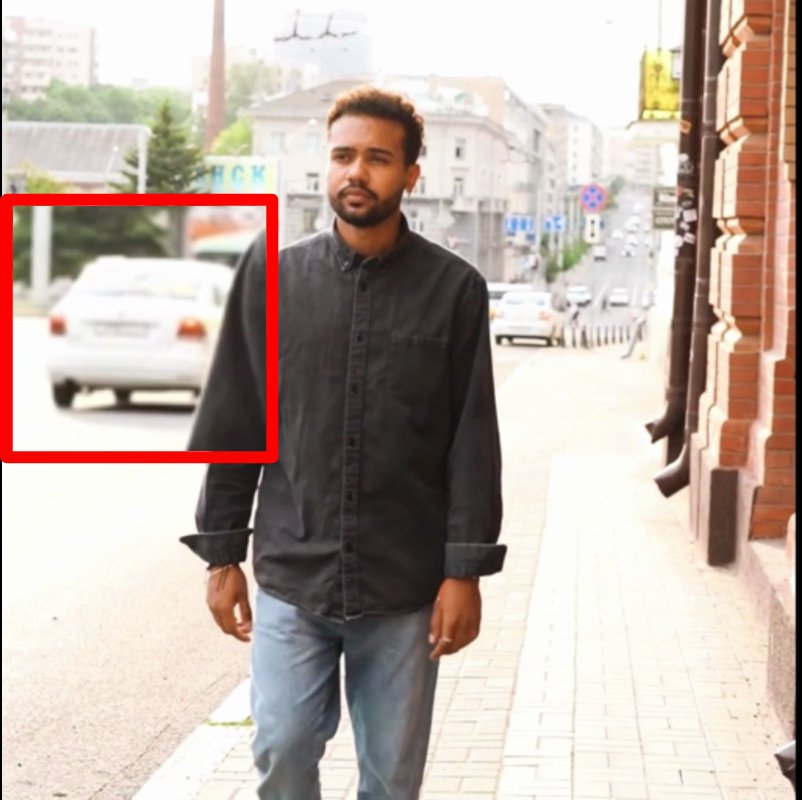}
        \end{minipage}
        }
        \hspace{-2.0mm} 
        \subfloat[ (b) Baseline]{
        \begin{minipage}{0.185\linewidth} 
        \includegraphics[width=\linewidth]{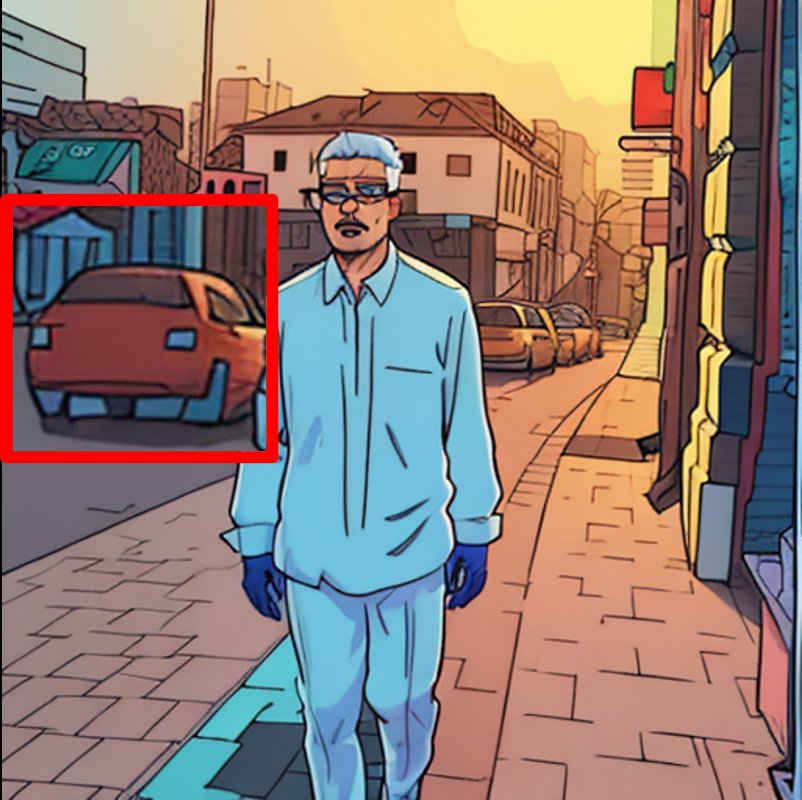}\vspace{0.5mm}
        \includegraphics[width=\linewidth]{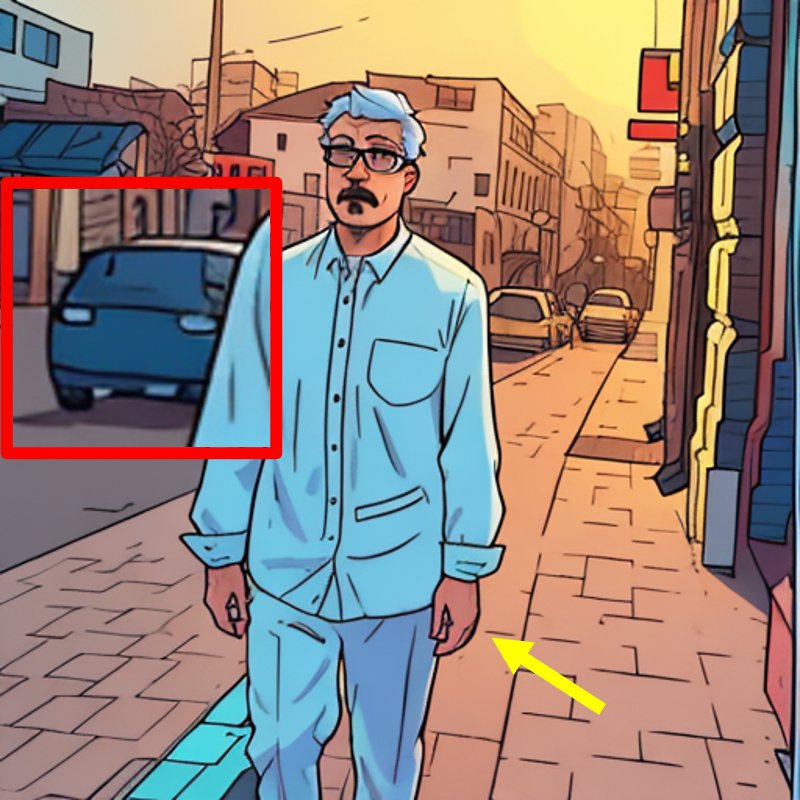}
        \end{minipage}
        }
        \hspace{-2.0mm} 
        \subfloat[ (c) w/ \emph{Q} Warping]{
        \begin{minipage}{0.185\linewidth} 
        \includegraphics[width=\linewidth]{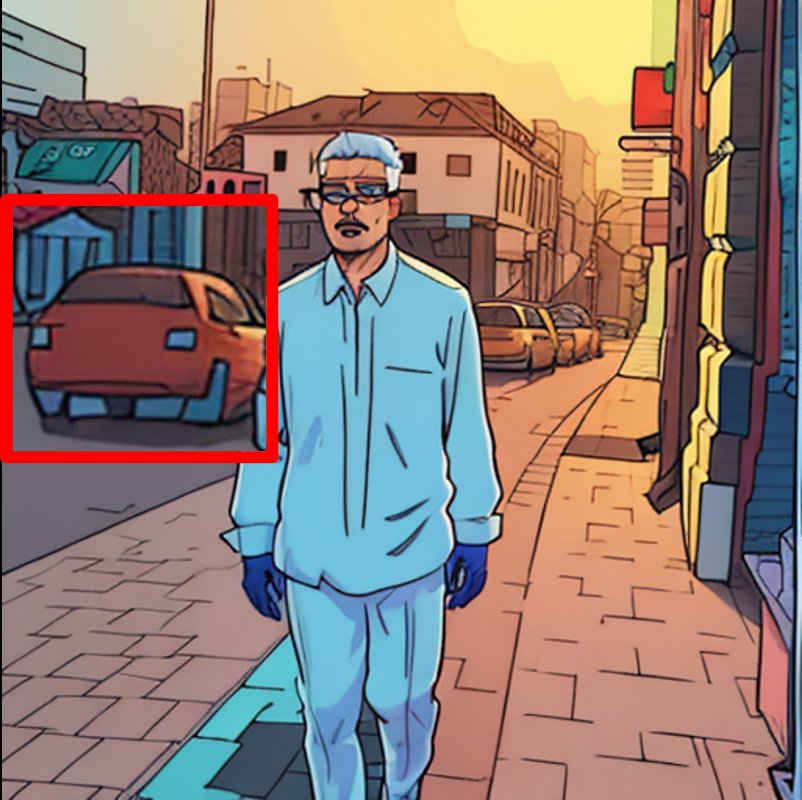}\vspace{0.5mm}
        \includegraphics[width=\linewidth]{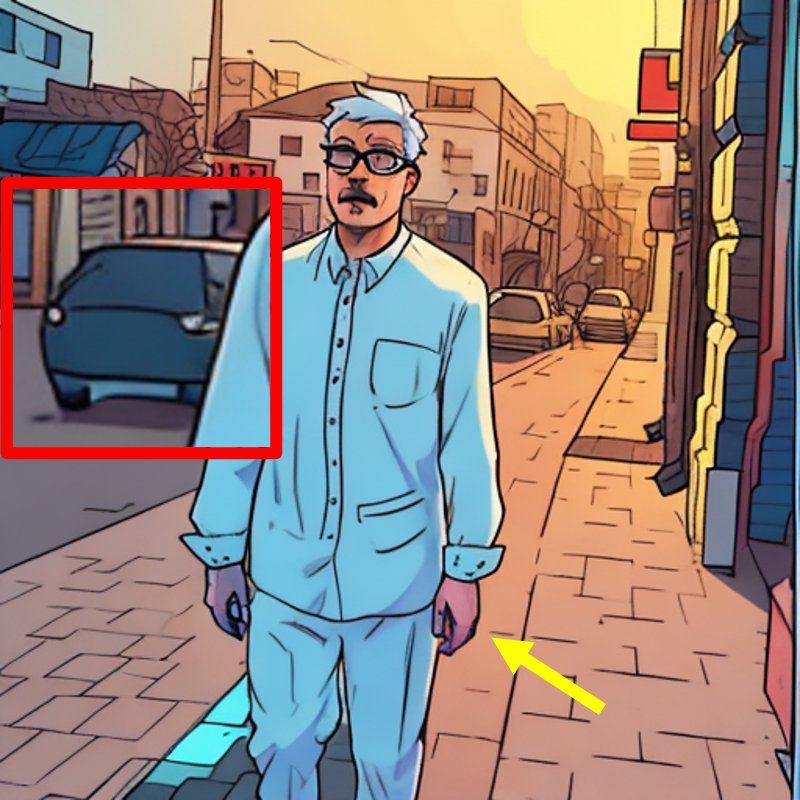}
        \end{minipage}
        }
        \hspace{-2.0mm} 
        \subfloat[ (d) w/ \emph{KV} Warping]{
        \begin{minipage}{0.185\linewidth} 
        \includegraphics[width=\linewidth]{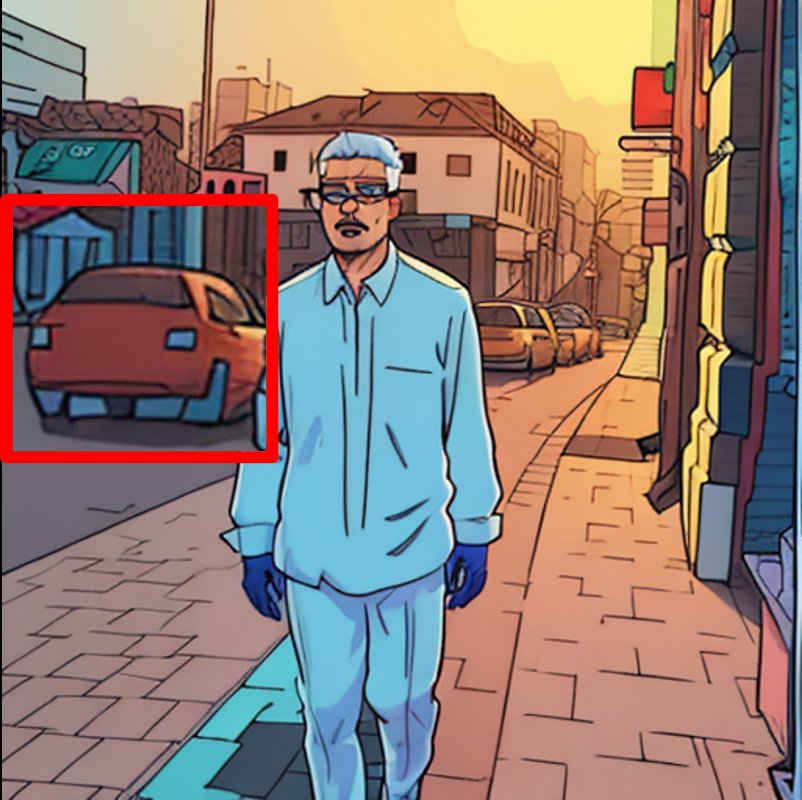}\vspace{0.5mm}
        \includegraphics[width=\linewidth]{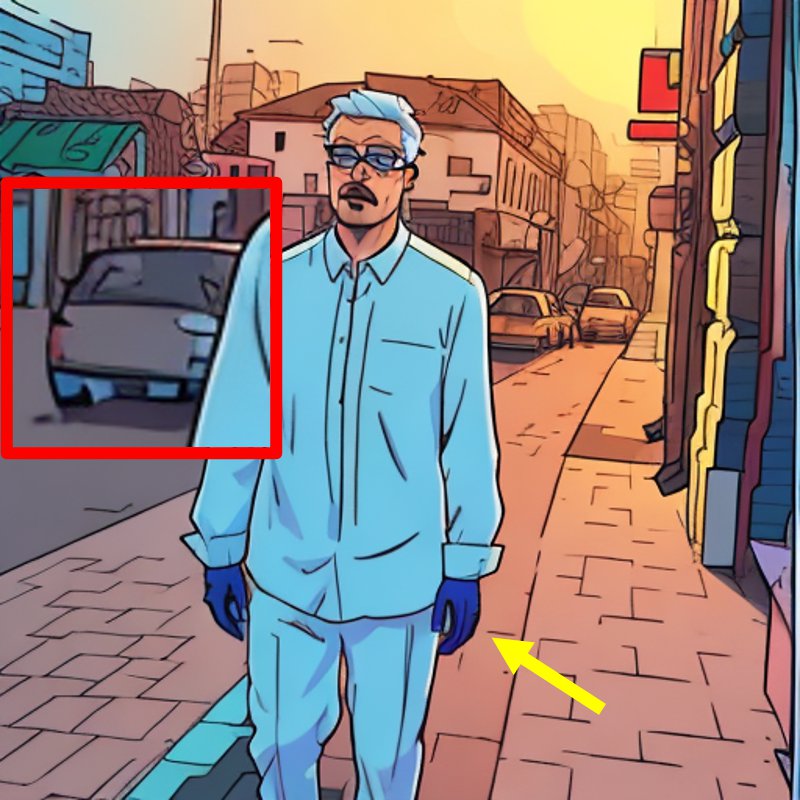}
        \end{minipage}
        }
        \hspace{-2.0mm} 
        \subfloat[ (e) Full]{
        \begin{minipage}{0.185\linewidth} 
        \includegraphics[width=\linewidth]{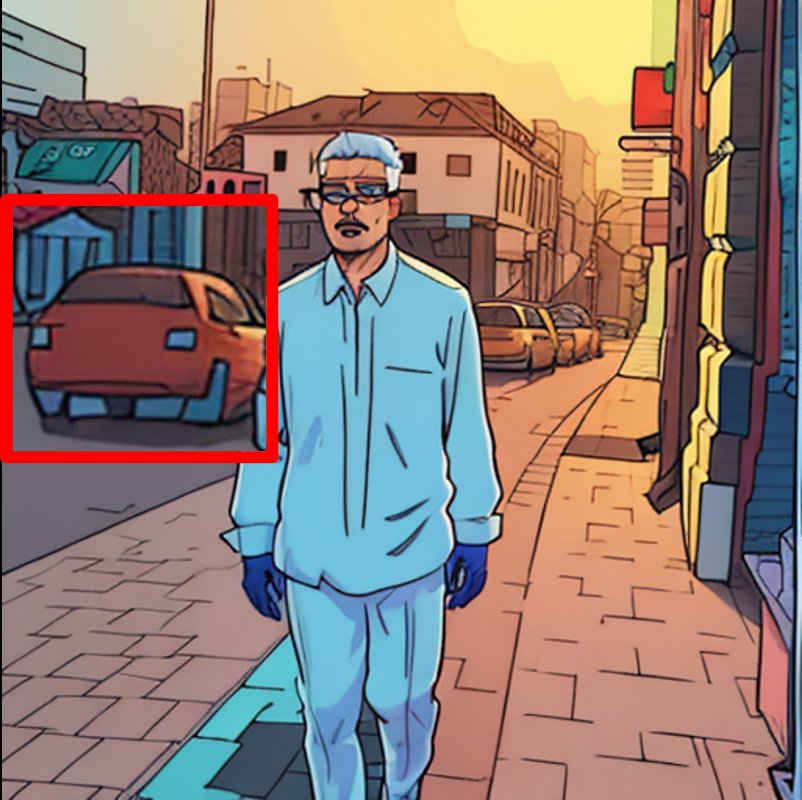}\vspace{0.5mm}
        \includegraphics[width=\linewidth]{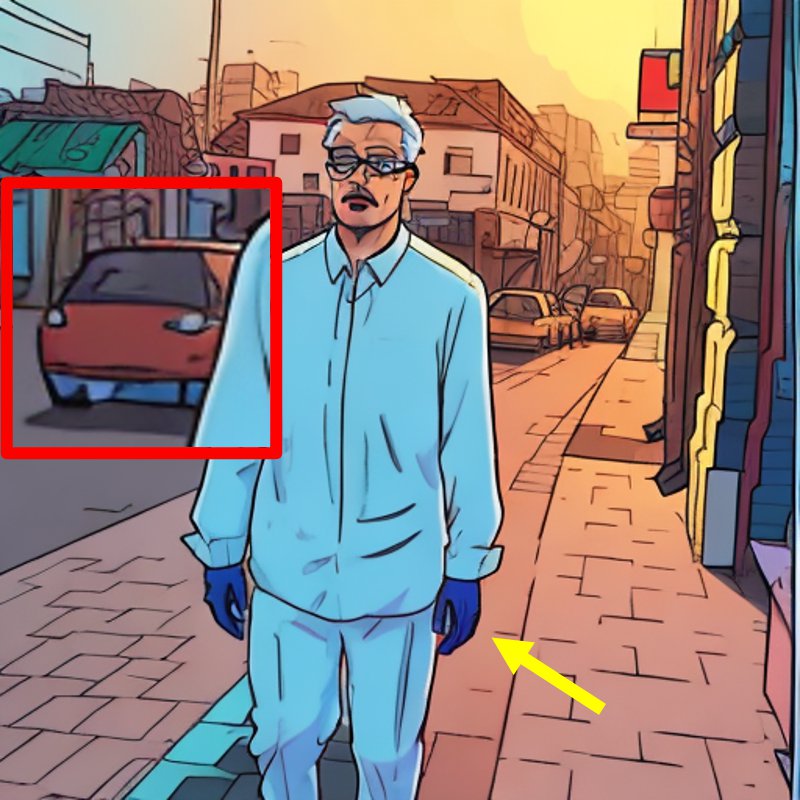}
        \end{minipage}
        }

    \vspace{-1mm}
    \caption{\revision{Qualitative comparison with different variants. Incorporating all components ensures consistency in the color of car (red box) and hand (yellow arrow).} \textit{Prompt: A man with glass walks in the street, cartoon style.}}
    \label{fig:ab_1}
    \vspace{-3mm}
    \end{figure}

\textbf{Effectiveness of Warping of Different Components.}
To evaluate the effectiveness of different components in our \emph{TokenWarping}, we conduct ablation studies in this section. {To avoid other factors such as flow errors, we used 10 videos that achieved a \(100\%\) editing success rate with our baseline to faithfully evaluate our components.} We first set a baseline by using first \emph{key} and \emph{value} as cross-frame attention in Eq.~\ref{eq:1}. Moreover, we use optical flow to warp the \emph{query}, \emph{key} and \emph{value} patches in the self-attention. As shown in Fig.~\ref{fig:ab_1}, compared with Baseline, variant \emph{Q} Warping gets small improvement. Variant \emph{KV} Warping improves the color consistency of ``hands'', but still failed on preserving the color on ``car''. Benefit from warping \emph{query} patches aligned with warping \emph{key} and \emph{value} patches, our full method improves the temporal consistency effectively.

The same conclusion also can be evidenced by quantitatively comparison in Tab.~\ref{tab:com_ab}, our full model receives the best performance on all three metrics. Compared to the baseline, warping the \emph{query} patches results in lower Warp-Err scores. Warping the \emph{key} and \emph{value} patches is more effective in improving the Tem-Con score.

\begin{figure}[t]
\centering
    \captionsetup[subfloat]{labelformat=empty,justification=centering}

    \subfloat[(a) Input]{
    \begin{minipage}{0.235\linewidth}
    \label{fig:ab_3a}
    \includegraphics[width=\linewidth]{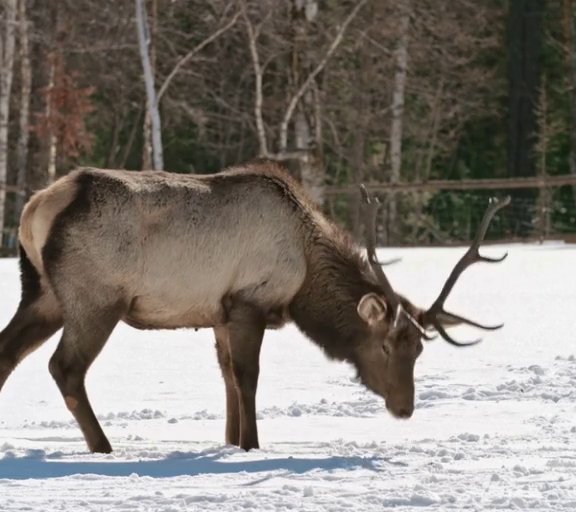}\vspace{0.5mm}
    \includegraphics[width=\linewidth]{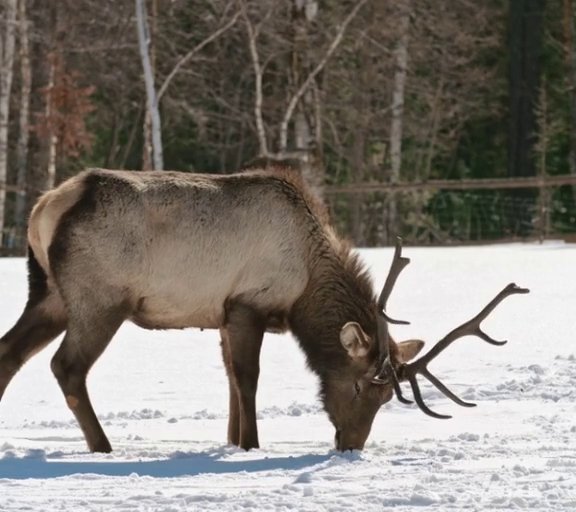}
    \end{minipage}
    }
    \hspace{-1.8mm}
    \subfloat[(b) w/o Anchor]{
    \begin{minipage}{0.235\linewidth}
    \label{fig:ab_3b}
    \includegraphics[width=\linewidth]{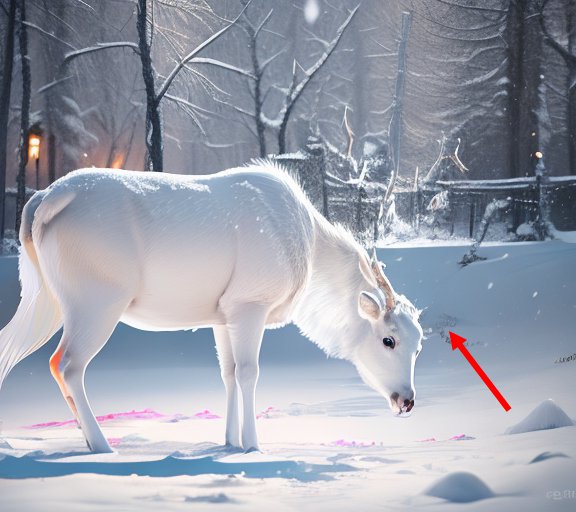}\vspace{0.5mm}
    \includegraphics[width=\linewidth]{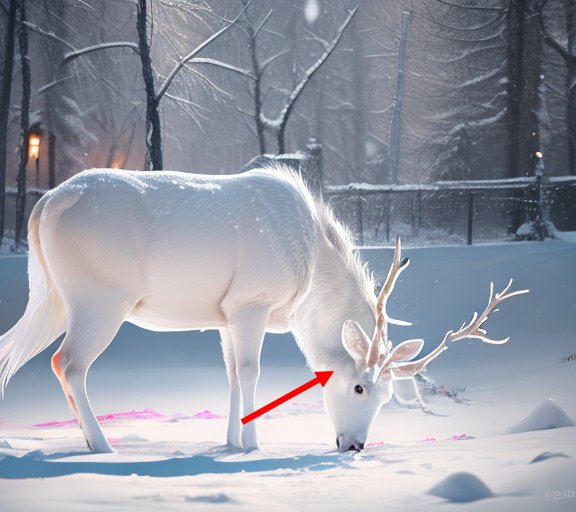}
    \end{minipage}
    }
    \hspace{-1.8mm}
    \subfloat[(c) w/o Warping]{
    \label{fig:ab_3c}
    \begin{minipage}{0.235\linewidth}
    \includegraphics[width=\linewidth]{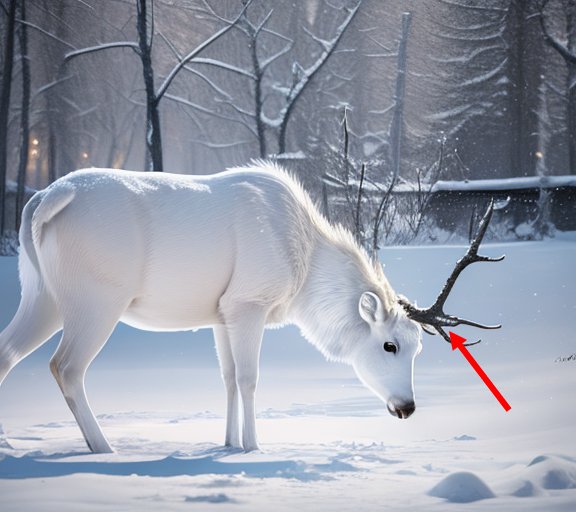}\vspace{0.5mm}
    \includegraphics[width=\linewidth]{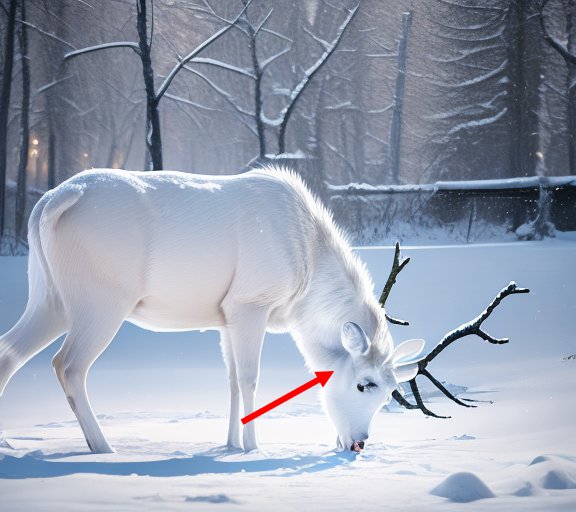}
    \end{minipage}
    }
    \hspace{-1.8mm}
    \subfloat[(d) Full]{
    \begin{minipage}{0.235\linewidth}
    \label{fig:ab_3d}
    \includegraphics[width=\linewidth]{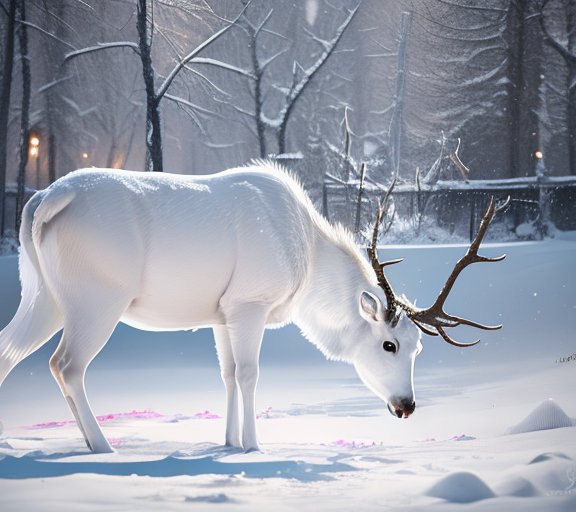}\vspace{0.5mm}
    \includegraphics[width=\linewidth]{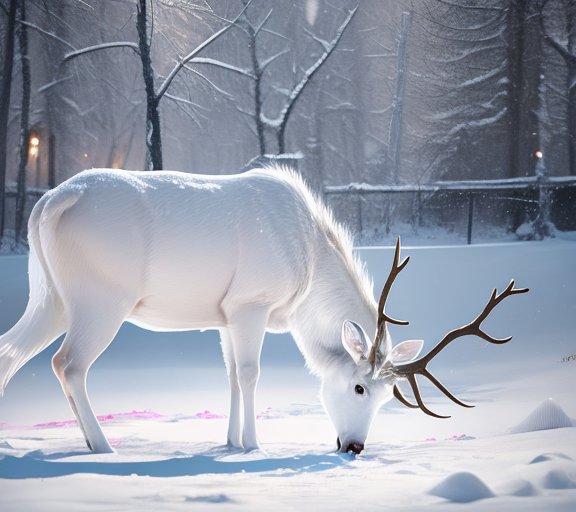}
    \end{minipage}
    }

    \vspace{-1mm}
    \caption{Ablation study of introducing anchor tokens and effectiveness of flow warping. \textit{Prompt: A white deer in the snow.}}
    \vspace{-5mm}
\label{fig:ab_3}
\end{figure}


\noindent \textbf{Effectiveness of Anchor Tokens.}
We also make a comparison in processing \emph{key} and \emph{value} patches to evaluate the effectiveness of anchor tokens. In Fig.~\ref{fig:ab_3a} the optical flow for the antlers is difficult to match, resulting in inaccurate optical flow estimation. Fig.~\ref{fig:ab_3b} shows the results only warping key and \emph{value} patches lead to incomplete antlers. Fig.~\ref{fig:ab_3c} the results demonstrate that using anchor tokens alone cannot achieve correct spatial correspondence, resulting in the antlers being rendered behind the ``deer's head''. Fig.~\ref{fig:ab_3d} shows the results of fully combining the operator. It can be observed that concatenating the anchor tokens and warping tokens effectively handles both spatial and temporal correspondences. The anchor tokens provides global correspondence, while the warping tokens provide local correspondence.

\begin{figure}[t]
\centering
    \captionsetup[subfloat]{labelformat=empty,justification=centering}
    \subfloat[ (a) Input]{
    \begin{minipage}{0.235\linewidth}
    \label{fig:ab_2a}
    \includegraphics[width=\linewidth]{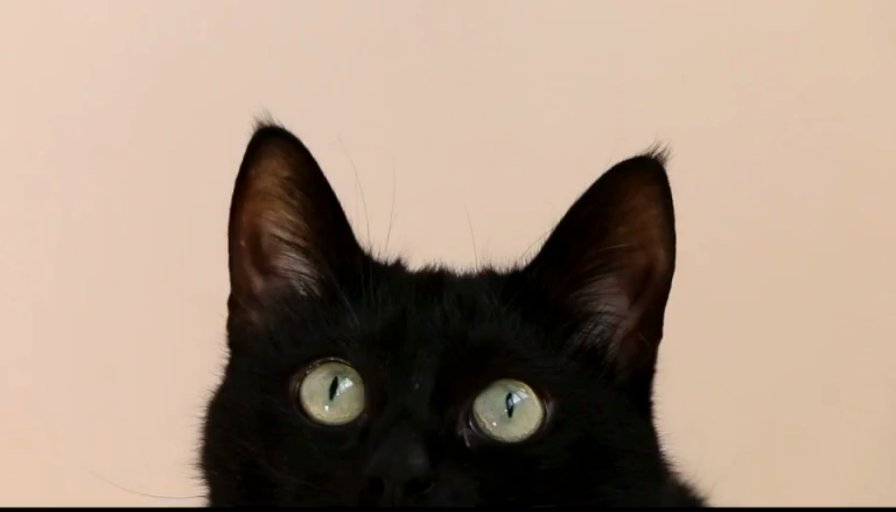}\vspace{0.5mm}
    \includegraphics[width=\linewidth]{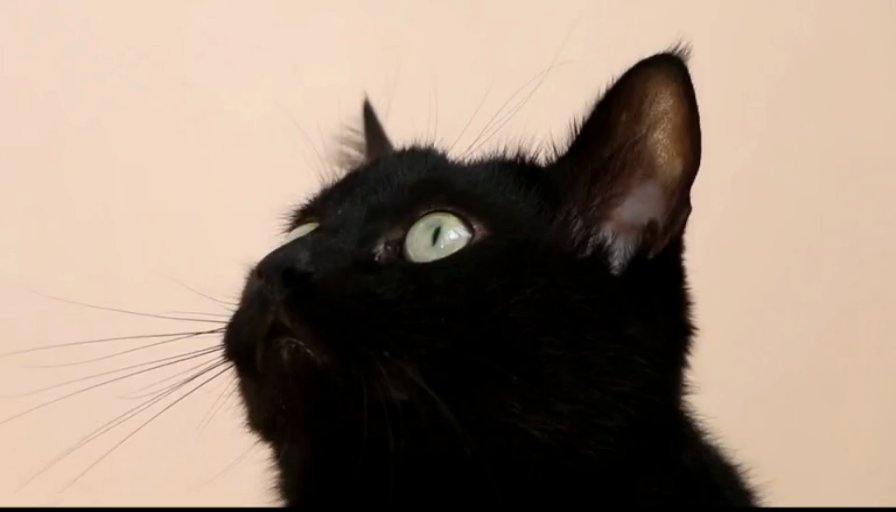}
    \end{minipage}
    }
    \hspace{-1.8mm}
    \subfloat[ (b) w/o \emph{Q} Fusion]{

    \begin{minipage}{0.235\linewidth}
    \label{fig:ab_2b}
    \includegraphics[width=\linewidth]{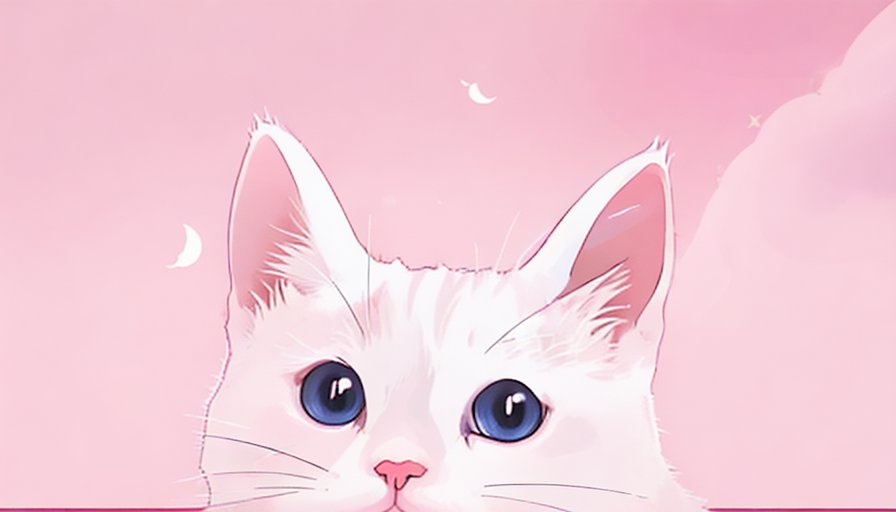}\vspace{0.5mm}
    \includegraphics[width=\linewidth]{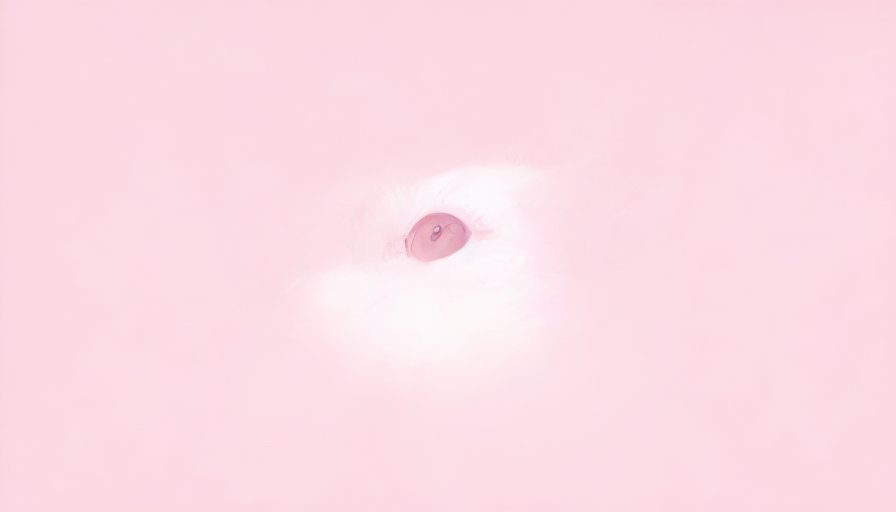}
    \end{minipage}
    }
    \hspace{-1.8mm}
    \subfloat[ (c) w/o \emph{KV} Fusion]{
    \begin{minipage}{0.235\linewidth}
    \label{fig:ab_2c}
    \includegraphics[width=\linewidth]{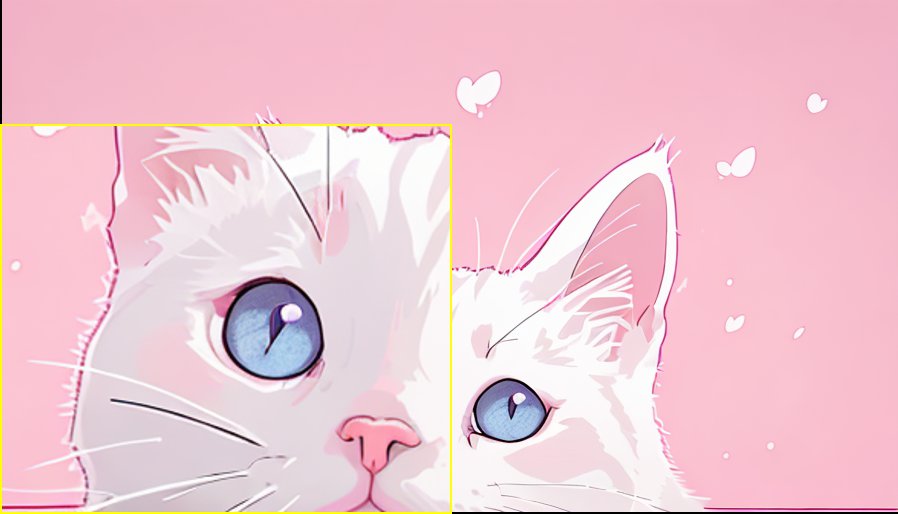}\vspace{0.5mm}
    \includegraphics[width=\linewidth]{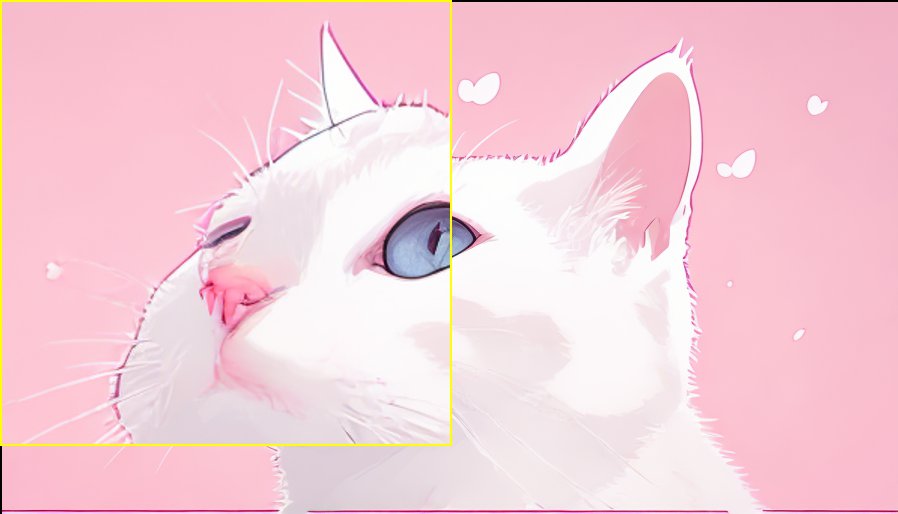}
    \end{minipage}
    }
    \hspace{-1.8mm}
    \subfloat[ (d) Full]{
    \begin{minipage}{0.235\linewidth}
    \label{fig:ab_2d}
    \includegraphics[width=\linewidth]{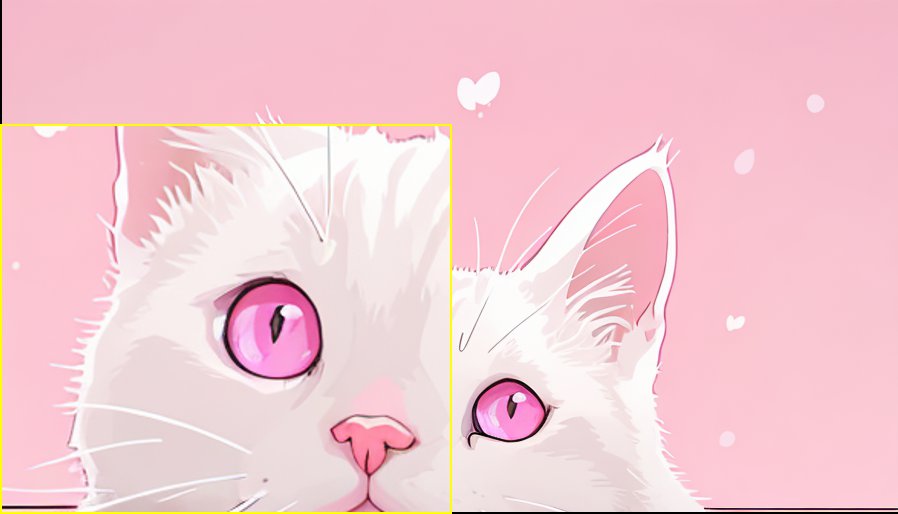}\vspace{0.5mm}
    \includegraphics[width=\linewidth]{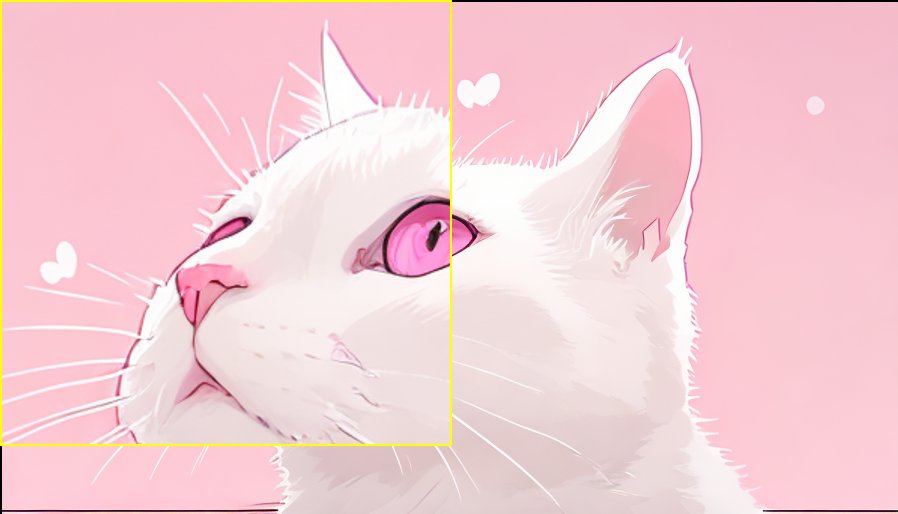}
    \end{minipage}
    }
    \vspace{-1mm}
    \caption{Effectiveness of occlusion-mask fusion in warping operation. \textit{Prompt: A white cat in pink background.}}
    \vspace{-5mm}
\label{fig:ab_2}
\end{figure}


\noindent \textbf{Effectiveness of Occlusion-mask Fusion in Warping Operation.} In Fig.~\ref{fig:ab_2}, we study the effectiveness of fusion in warping operation. Fig.~\ref{fig:ab_2b} illustrates the results without fusion during the warping of \emph{query} patches, where the subsequent frame fails to aggregate aligned features from the warped \emph{query} patches. Fig.~\ref{fig:ab_2c} demonstrates the results without fusion in warping the \emph{key} and \emph{value} patches. The ``cat's nose'' exhibits a distinct ghosting, indicating that the occlusion region has not been translated effectively. Fig.~\ref{fig:ab_2d} displays the results of the fully warped operator, showing that the fusion operation leads to smoother and more coherent results.



\noindent \textbf{Spatial Correspondence of Different Attention-block Combinations.}
Our \emph{TokenWarping} leverages optical flow to propagate correspondences across frames, which requires precise spatial alignment between token features and the source frames. Once aligned, optical flow can effectively enforce temporal consistency in the token features. 
We conducted ablation experiments on different Transformer blocks within the decoder to evaluate the spatial correspondence achieved by various block configurations. Our model is based on Stable Diffusion 1.5 with a U-Net architecture, whose decoder comprises stacked Transformer blocks, each containing a self-attention layer. \revision{In this context, “Block \&1” denotes applying the warping operator only to the first block, while “Block 2\&3” indicates applying it to the second and third blocks.} The first block captures the most abstract features, whereas the third block is closer to the pixel space.

\begin{figure}[t]
\centering
    \captionsetup[subfloat]{labelformat=empty,justification=centering}

    \subfloat[(a) Input]{
    \begin{minipage}{0.235\linewidth}
    \label{fig:ab_4a}
    \includegraphics[width=\linewidth]{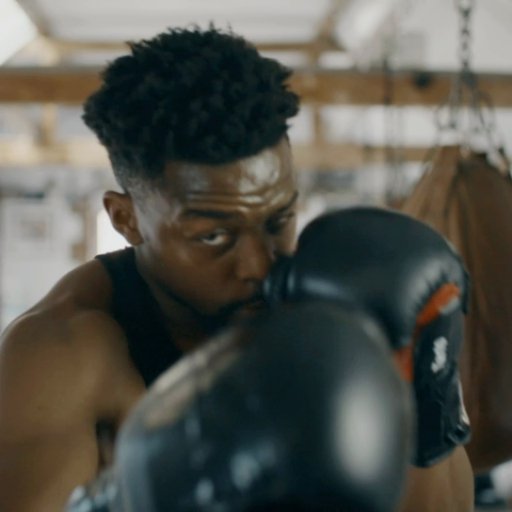}\vspace{0.5mm}
    \includegraphics[width=\linewidth]{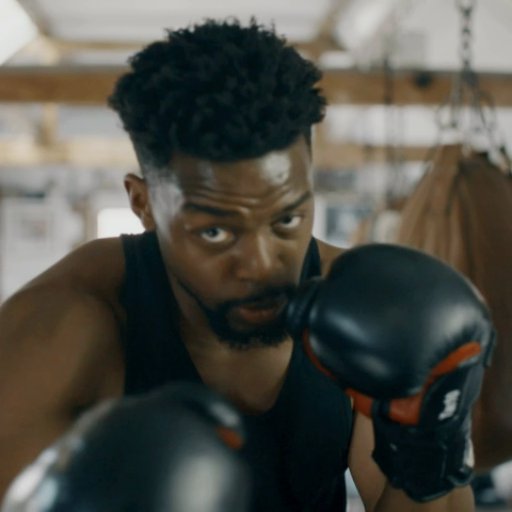}
    \end{minipage}
    }
    \hspace{-1.8mm}
    \subfloat[(b) Block 1\&2]{
    \begin{minipage}{0.235\linewidth}
    \label{fig:ab_4b}
    \includegraphics[width=\linewidth]{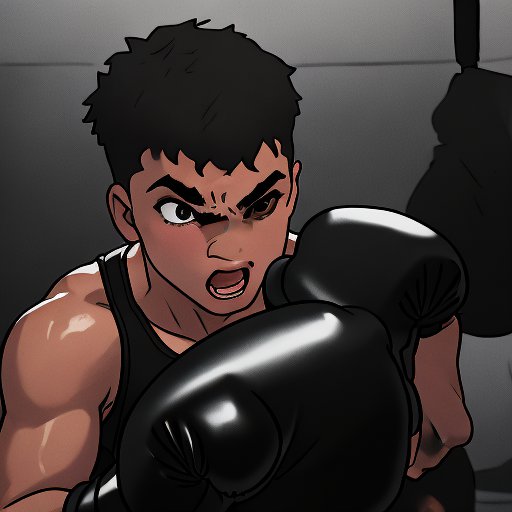}\vspace{0.5mm}
    \includegraphics[width=\linewidth]{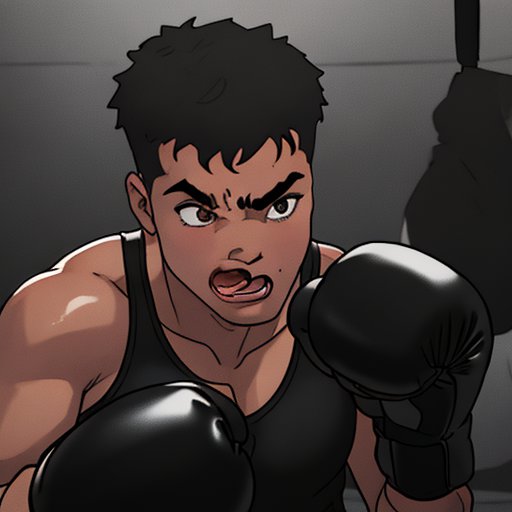}
    \end{minipage}
    }
    \hspace{-1.8mm}
    \subfloat[(c) Block 2\&3]{
    \begin{minipage}{0.235\linewidth}
    \label{fig:ab_4c}
    \includegraphics[width=\linewidth]{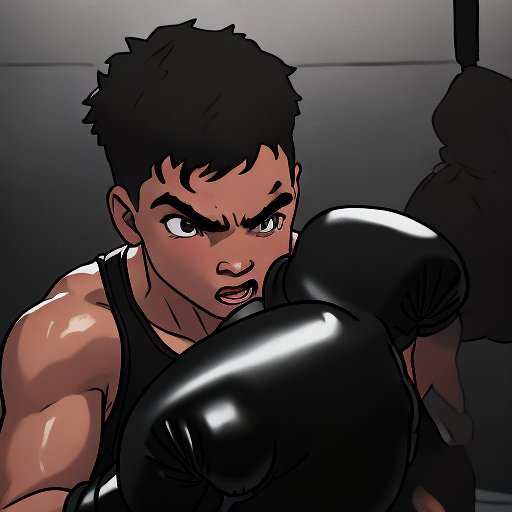}\vspace{0.5mm}
    \includegraphics[width=\linewidth]{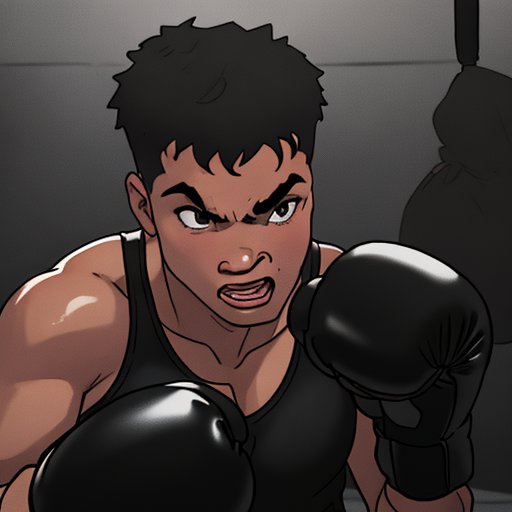}
    \end{minipage}
    }
    \hspace{-1.8mm}
    \subfloat[(d) Block 1\&2\&3]{
    \begin{minipage}{0.235\linewidth}
    \label{fig:ab_4d}
    \includegraphics[width=\linewidth]{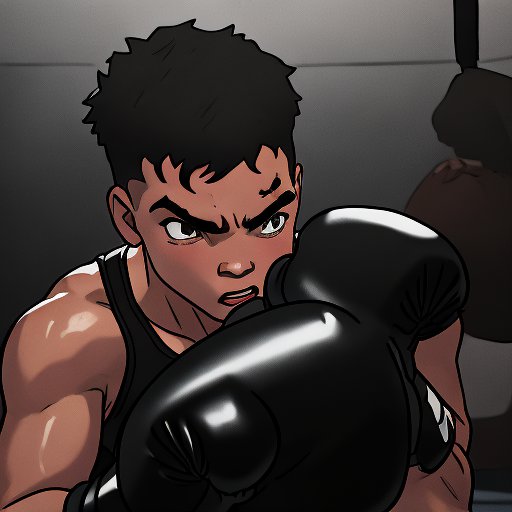}\vspace{0.5mm}
    \includegraphics[width=\linewidth]{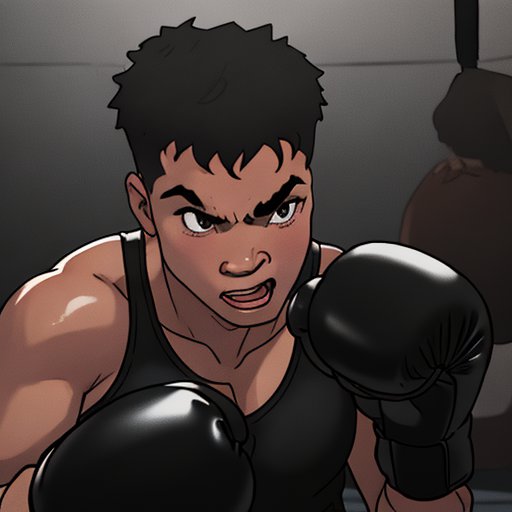}
    \end{minipage}
    }

    \vspace{-1mm}
    \caption{\revision{Effectiveness of different blocks in U-net decoder. ``Block 2\&3” denotes applying the warping operator in $1_{st}$ and $2_{nd}$ blocks.} \textit{Prompt: A black boxer wearing black boxing gloves punches towards the camera, cartoon style.}}
    \vspace{-3mm}
\label{fig:ab_4}
\end{figure}

As shown in Fig.~\ref{fig:ab_4b}, applying flow-based attention solely to Block 12 yields inaccurate results for fine details (e.g., the ``Boxer's eyes''), likely due to ineffective feature aggregation from Block 3. In Fig.~\ref{fig:ab_4c}, removing constraints from Block 1 introduces minor pseudo-shadows around the same region. By contrast, applying token warping to Blocks 1\&2\&3 simultaneously produces more visually coherent and appealing results.

\subsection{Further Analysis}
\label{sec:limitation}



\noindent \textbf{Long Video Translation}
\revision{To enable our method on devices with limited VRAM, we split the video into multiple clips for processing. By default, each clip contains 8 frames, which at a resolution of $512 \times 512$ requires approximately {6–8 GB} of GPU memory. For each clip, we store the tokens from the last frame of the previous clip as well as the anchor tokens, ensuring temporal consistency across clips. This design allows us to perform translation in a clip-by-clip manner, while leveraging the flow-guided attention mechanism to maintain temporal coherence.}

\revision{However, as the frame numbers increase, the error in optical flow estimation will be accumulated, resulting in the temporal inconsistency. According to our empirical observations, our method is capable of handling videos of approximately 120 frames with satisfactory performance.}

\begin{figure}[t]
\centering
    \captionsetup[subfloat]{labelformat=empty,justification=centering}
    \subfloat[\small frame \#05]{
    \begin{minipage}{0.235\linewidth}
    \includegraphics[width=\linewidth]{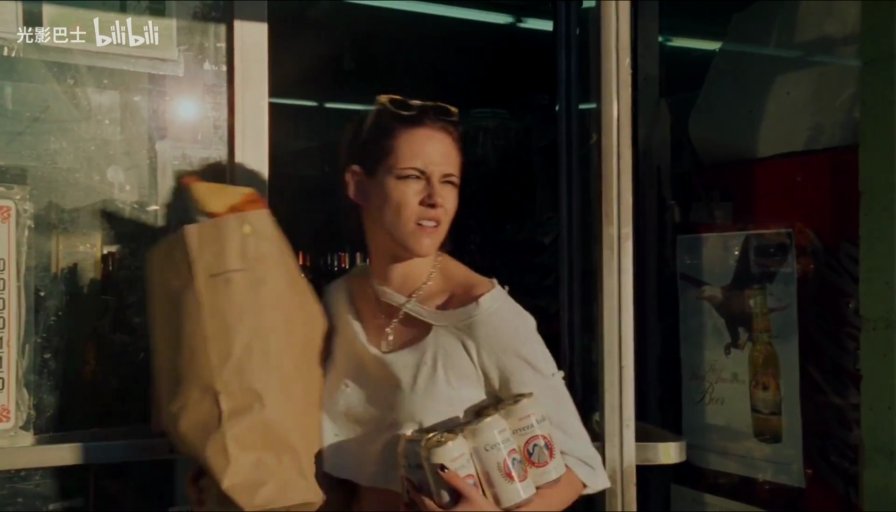}\vspace{0.5mm}
    \includegraphics[width=\linewidth]{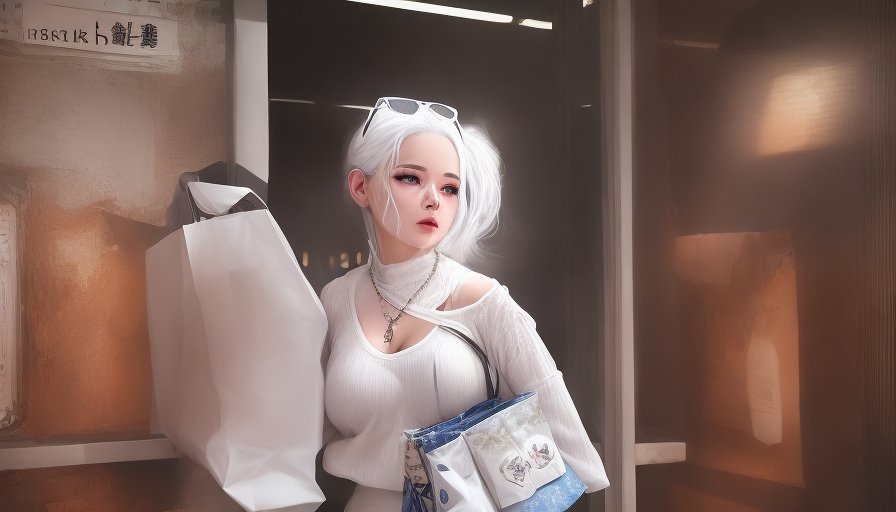}
    \end{minipage}
    }
    \hspace{-1.8mm}
    \subfloat[\small frame \#12]{
    \begin{minipage}{0.235\linewidth}
    \includegraphics[width=\linewidth]{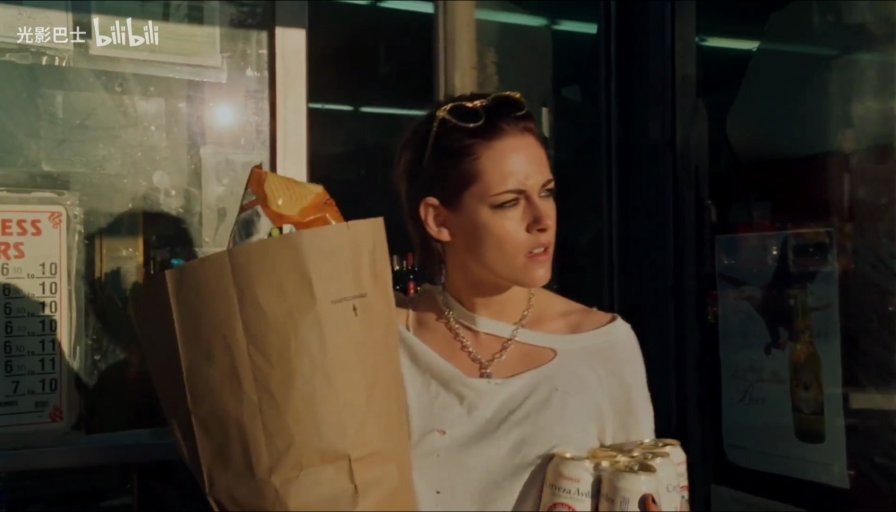}\vspace{0.5mm}
    \includegraphics[width=\linewidth]{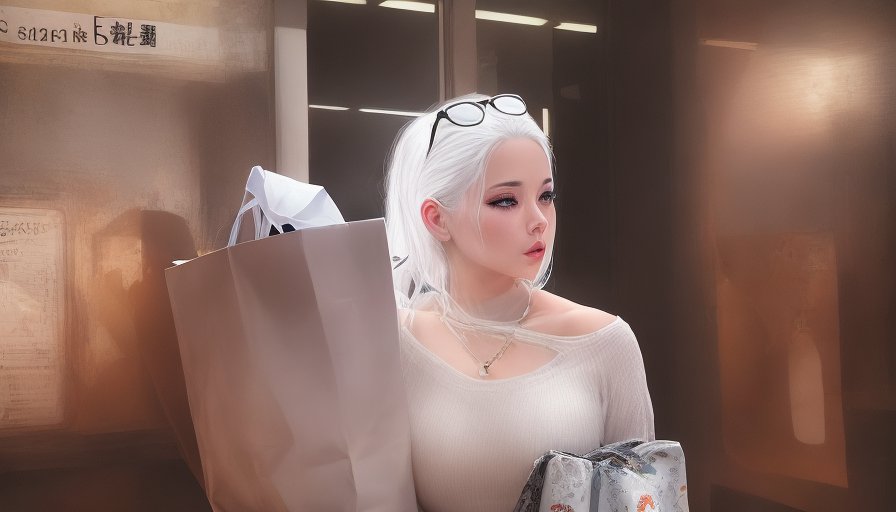}
    \end{minipage}
    }
    \hspace{-1.8mm}
    \subfloat[\small frame \#21]{
    \begin{minipage}{0.235\linewidth}
    \includegraphics[width=\linewidth]{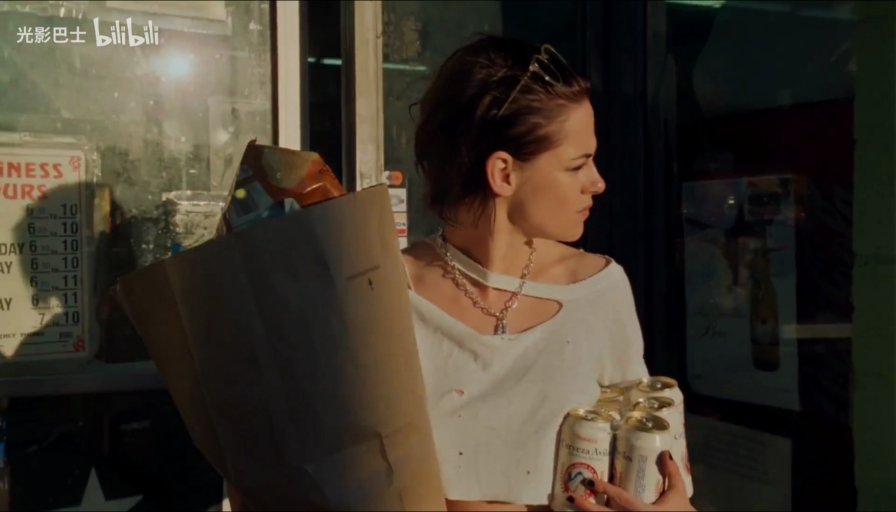}\vspace{0.5mm}
    \includegraphics[width=\linewidth]{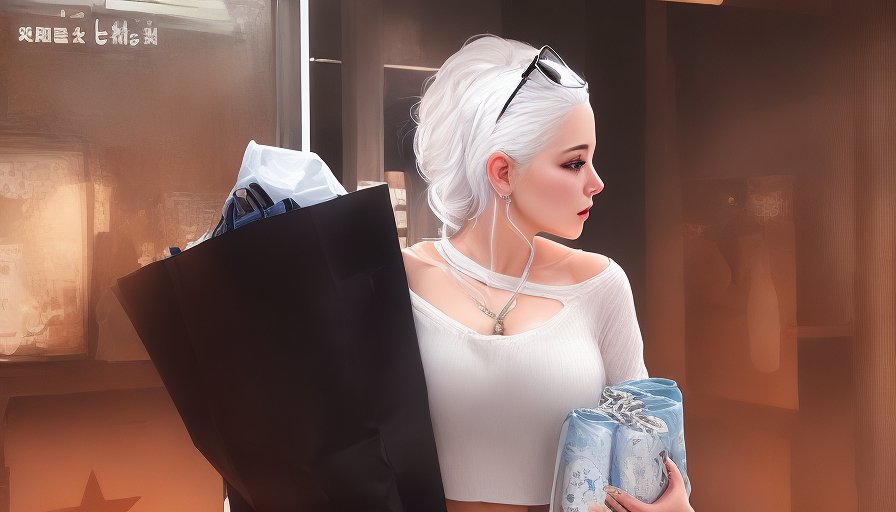}
    \end{minipage}
    }
    \hspace{-1.8mm}
    \subfloat[\small frame \#57]{
    \begin{minipage}{0.235\linewidth}
    \includegraphics[width=\linewidth]{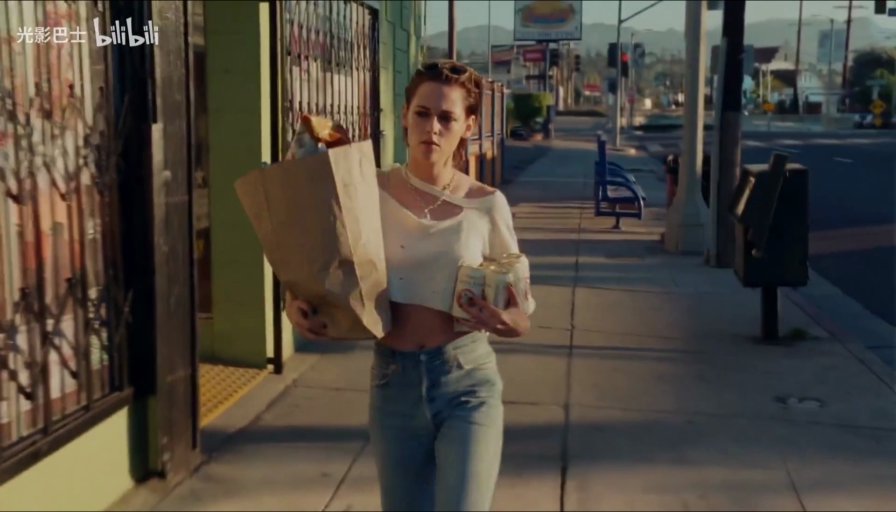}\vspace{0.5mm}
    \includegraphics[width=\linewidth]{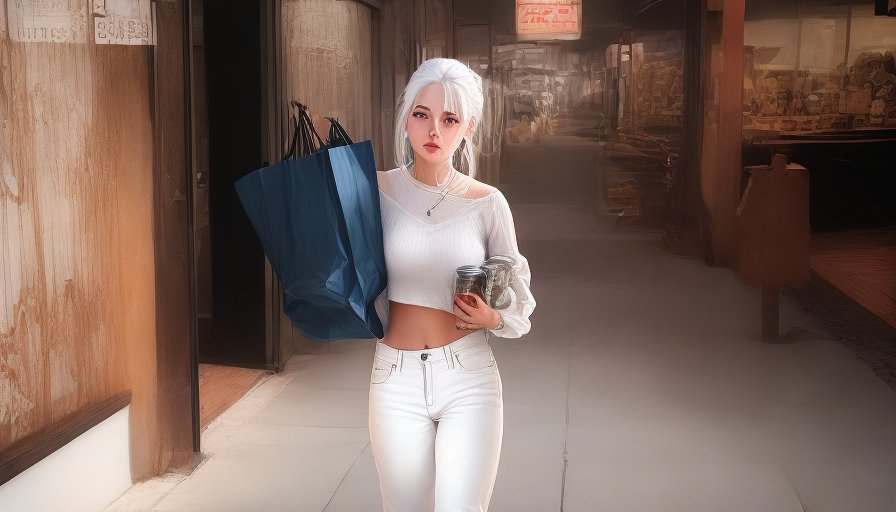}
    \end{minipage}
    }
    \vspace{-1mm}
    \caption{Failure case in complex scene. \textit{Prompt: A woman with white hair walking down a sidewalk, shopping bags and bear can, white top and white jeans.}}
    \vspace{-3mm}
\label{fig:failure}
\end{figure}


\noindent \textbf{Analysis of Flow Errors.} 
Complex non-rigid motion and severe occlusions present significant challenges for flow-based approaches. When the estimated flow becomes inaccurate and temporal correspondences weaken, the overall performance inevitably degrades.
Nevertheless, in scenarios where optical flow estimation remains relatively tractable, such as scenes with a single object and a simple background (see ``cat'' in Fig.~\ref{fig:ab_2} or the flow visualization in the Supplementary Materials), our flow-based attention demonstrates a certain tolerance. We believe that warped \emph{query} patches can more easily aggregate attention with the warped \emph{key} and \emph{value} patches. In non-occluded areas, tokens along the same trajectory can aggregate more effectively.

\begin{figure}[t]
\centering
    \captionsetup[subfloat]{labelformat=empty,justification=centering}

    \begin{minipage}[c]{0.04\linewidth}
        \rotatebox{90}{\small Source}
    \end{minipage}%
    \begin{minipage}[c]{0.96\linewidth}
        \includegraphics[width=0.24\linewidth]{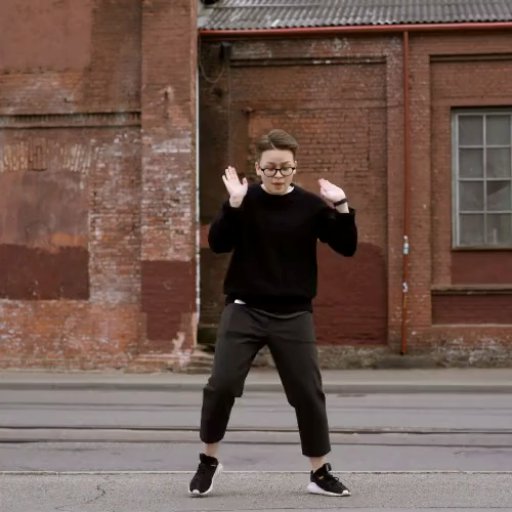}%
        \hspace{0.01mm}
        \includegraphics[width=0.24\linewidth]{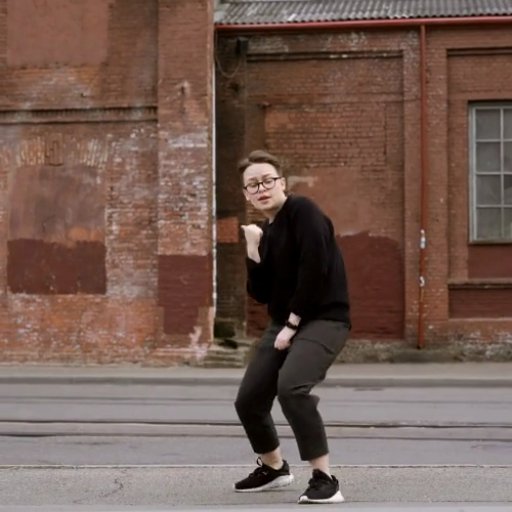}%
        \hspace{0.01mm}
        \includegraphics[width=0.24\linewidth]{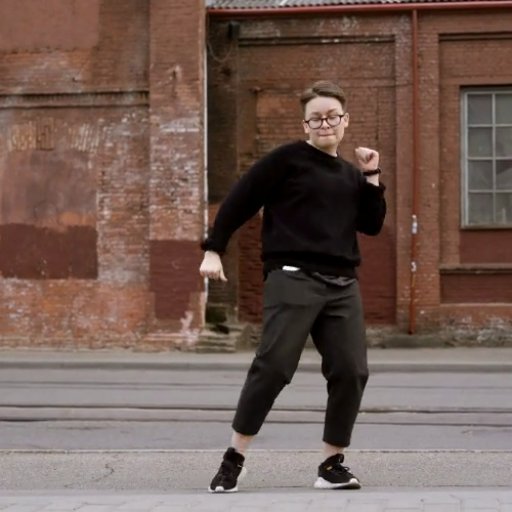}
        \hspace{0.01mm}
        \includegraphics[width=0.24\linewidth]{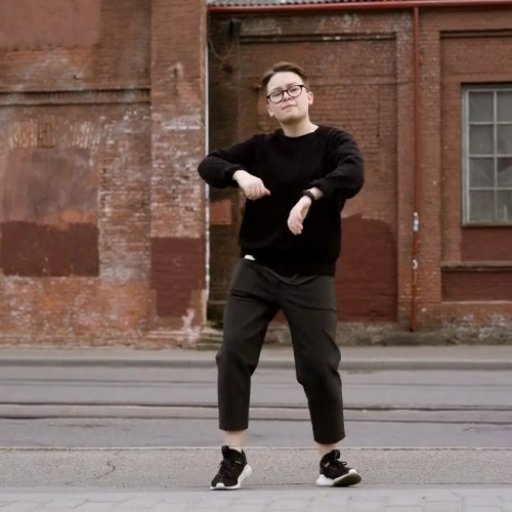}
    \end{minipage}\vspace{0.5mm}

    \begin{minipage}[c]{0.04\linewidth}
        \rotatebox{90}{\small Style}
    \end{minipage}%
    \begin{minipage}[c]{0.96\linewidth}
        \includegraphics[width=0.24\linewidth]{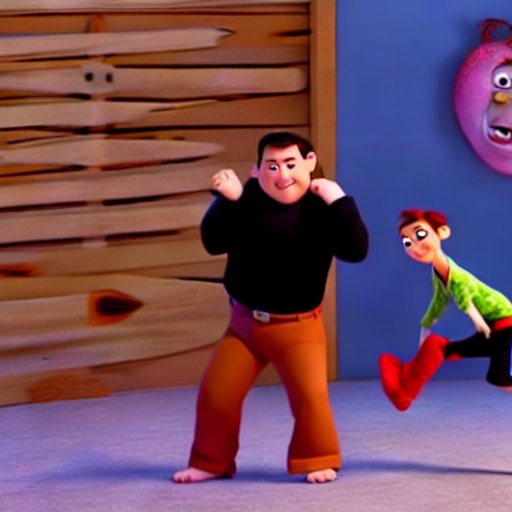}%
        \hspace{0.01mm}
        \includegraphics[width=0.24\linewidth]{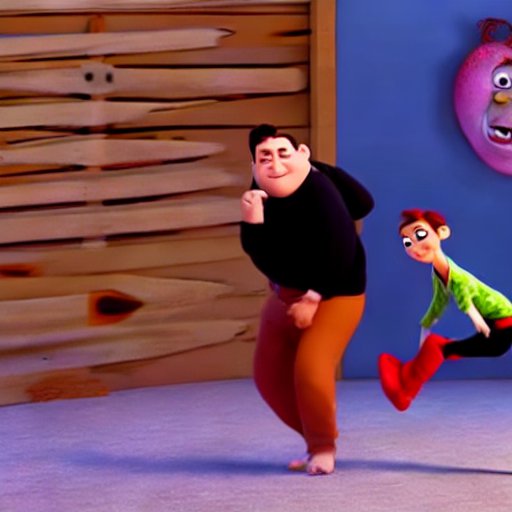}%
        \hspace{0.01mm}
        \includegraphics[width=0.24\linewidth]{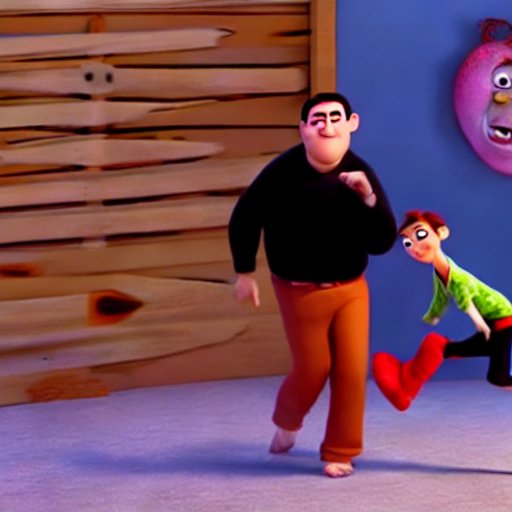}
        \hspace{0.01mm}
        \includegraphics[width=0.24\linewidth]{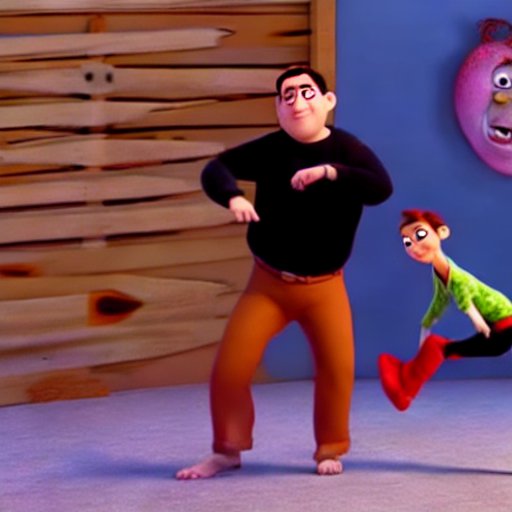}
    \end{minipage}\vspace{0.5mm}

    \begin{minipage}[c]{0.04\linewidth}
        \rotatebox{90}{\small Shape}
    \end{minipage}%
    \begin{minipage}[c]{0.96\linewidth}
        \includegraphics[width=0.24\linewidth]{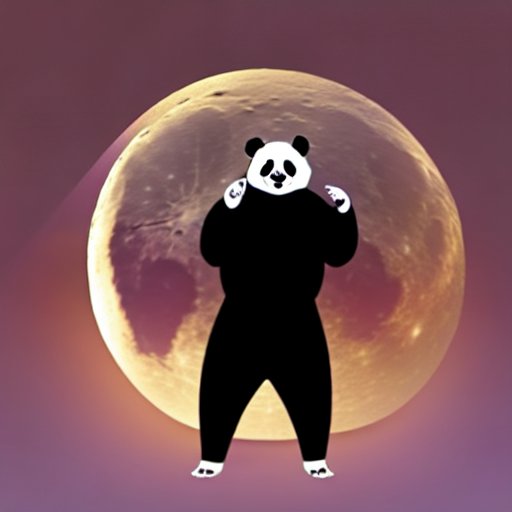}%
        \hspace{0.01mm}
        \includegraphics[width=0.24\linewidth]{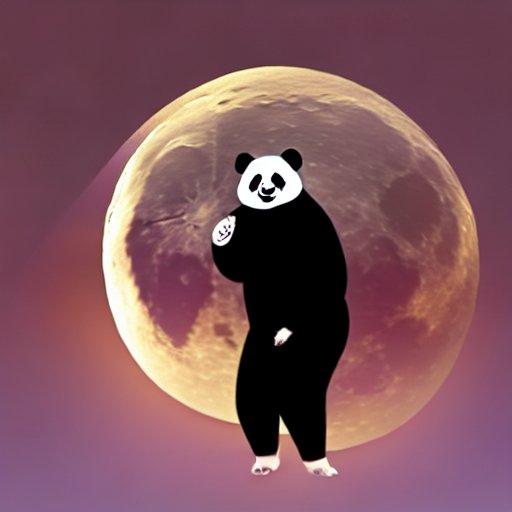}%
        \hspace{0.01mm}
        \includegraphics[width=0.24\linewidth]{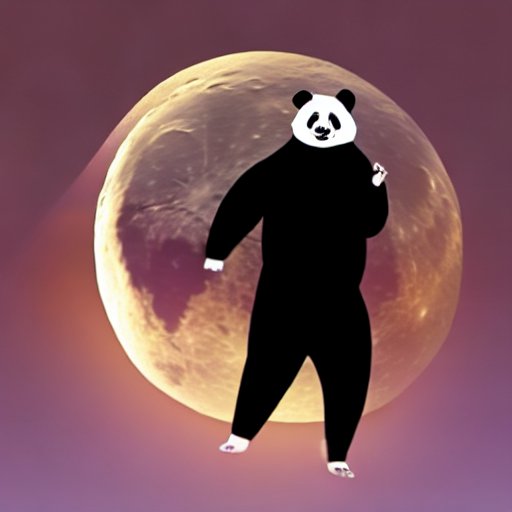}
        \hspace{0.01mm}
        \includegraphics[width=0.24\linewidth]{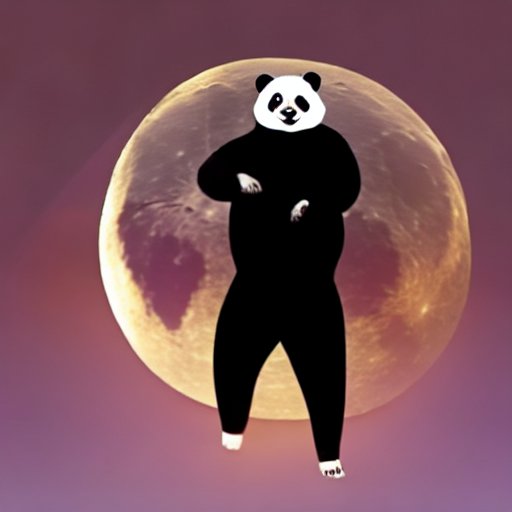}
    \end{minipage}
    \vspace{-1mm}
    \caption{Visualization of large gap editing. \textit{Prompt: A panda is dancing in moon.}}
    \vspace{-3mm} 
    \label{fig:gap_1}
\end{figure}

\noindent \textbf{Translation under Large Prompt Gaps.}
\label{sec:gap}
When there is a significant domain gap between the source and target prompts, the source correspondences cannot effectively guide the target videos. To address this, we extract appearance flows\cite{li2019dense} from the shared pose sequence in human action motion videos, which is agnostic to large prompt gaps. In Fig.~\ref{fig:gap_1}, we show the results of large gap editing. The appearance flow provide regional flow guidance, such as the face region lack of temporal constraints. In the future, we plan to explore more effective methods to manage large editing gaps.

\begin{figure}[t]
\centering
    \captionsetup[subfloat]{labelformat=empty,justification=centering}

    \begin{minipage}[c]{0.05\linewidth}
        \rotatebox{90}{\small Source}
    \end{minipage}%
    \begin{minipage}[c]{0.95\linewidth}
        \includegraphics[width=0.24\linewidth]{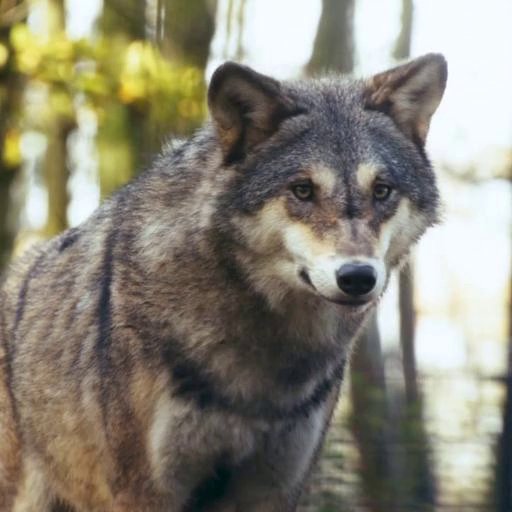}%
        \hspace{0.01mm}
        \includegraphics[width=0.24\linewidth]{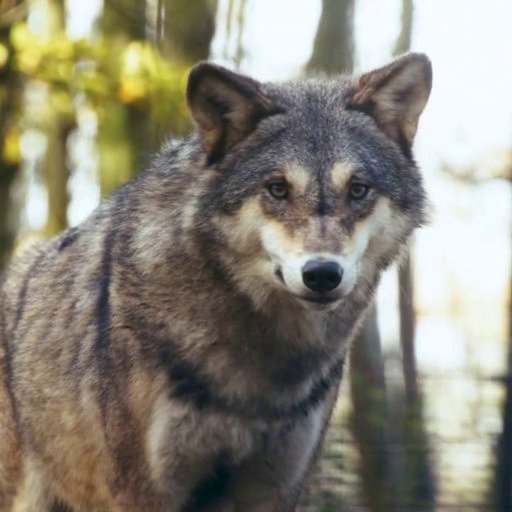}%
        \hspace{0.01mm}
        \includegraphics[width=0.24\linewidth]{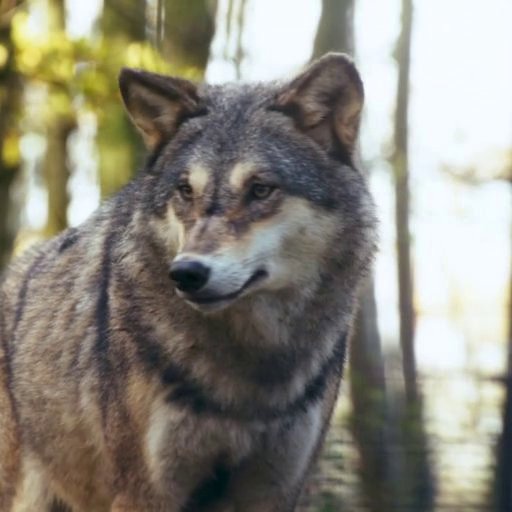}
        \hspace{0.01mm}
        \includegraphics[width=0.24\linewidth]{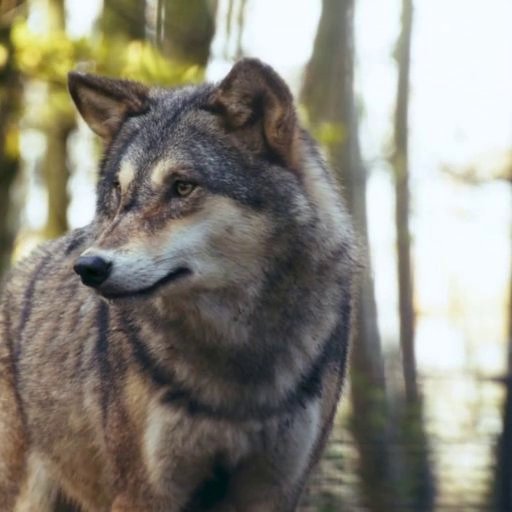}
    \end{minipage}\vspace{0.5mm}

    \begin{minipage}[c]{0.05\linewidth}
        \subfloat[ \rotatebox{90}{\small Target}]{}
    \end{minipage}%
    \begin{minipage}[c]{0.95\linewidth}
        \includegraphics[width=0.24\linewidth]{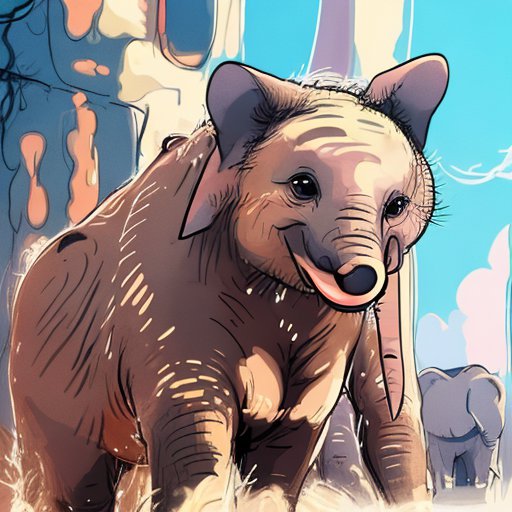}%
        \hspace{0.01mm}
        \includegraphics[width=0.24\linewidth]{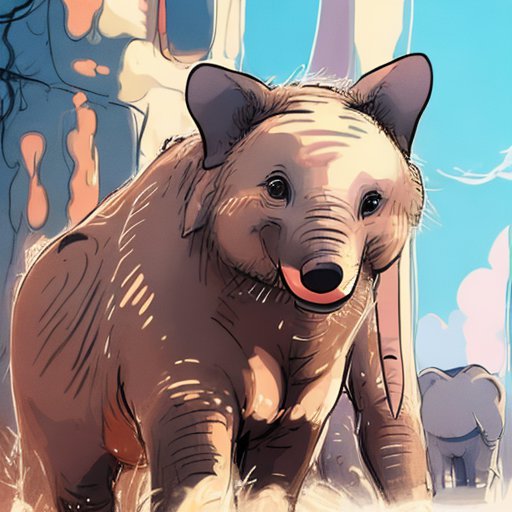}%
        \hspace{0.01mm}
        \includegraphics[width=0.24\linewidth]{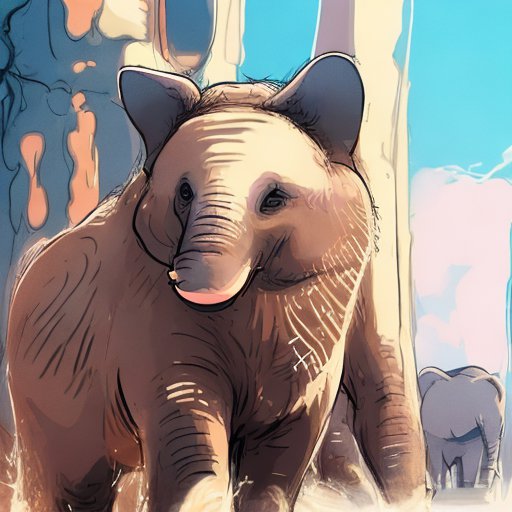}
        \hspace{0.01mm}
        \includegraphics[width=0.24\linewidth]{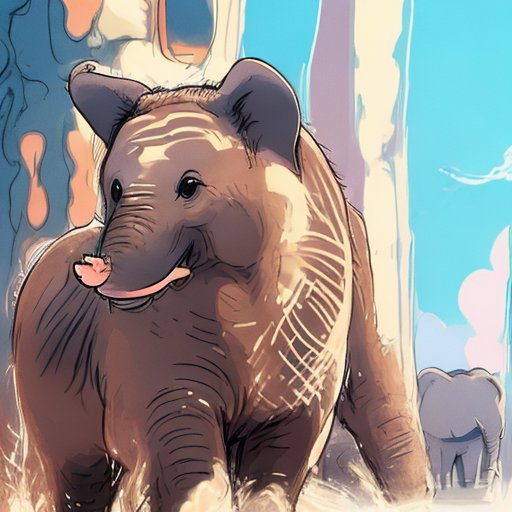}
    \end{minipage}\vspace{0.5mm}

    \vspace{-1mm}
    \caption{\revision{Our method fails the translation under large structural gap, since the optical flow preserve too much structural information of source videos.
    Prompt: \textit{A hand-drawn animation of an elephant in cartoon style.}}}     
    \vspace{-3mm} 
    \label{fig:shape}
\end{figure}

\noindent \revision{\textbf{Translation under Large Structural Gaps.}}
\revision{Our pipeline employs the optical flow to guide the translation of the source video. However, optical flow alone cannot propagate correspondences when there are significant structural gaps between the source and target prompts. As shown in Fig.\ref{fig:shape}, when editing from ``wolf'' to ``elephant'', the translated ``elephant'' presents more characters of ``wolf''. That shows the optical flow embeds the structural information from source videos, which is not suitable to guides the target video. For the translation under large structural deviation, we believe those motion transfer works~\cite{ling2024motionclone, pondaven2025video,ma2025follow} are more promising to handle those tasks.}

\section{Conclusion and Future Work}


\label{sec:conclusion}

In the paper, we present \emph{TokenWarping}, a novel framework for temporally zero-shot video translation. By identifying the inconsistency of tokens in SD's self-attention layer across different frames as a key challenge, we introduce optical flow extracted from the source video to warp the last frame's token. We then fuse the warped result with the current frame's token according to the occlusion mask.

Our warping process depends on the quality of the off-the-shelf optical flow detector and occlusion masks. While we have made efforts to ensure the accuracy of our approach, the performance of our framework can be influenced by the accuracy and robustness of these external components. 
Future work in this area could involve developing more advanced optical flow detection techniques or refining the occlusion mask generation process to improve the robustness of our framework. Additionally, exploring the integration of alternative sources of motion information, such as pose estimations or scene understanding, could further enhance the reliability and versatility of our approach. These considerations will be central to our ongoing efforts to address and mitigate these limitations and extend the applicability of our framework.

{
\bibliographystyle{ieee}
\bibliography{egbib}
}

\end{document}